\documentclass[journal=jctcce,manuscript=article]{achemso}

\usepackage[T1]{fontenc}
\usepackage[utf8]{inputenc}
\usepackage{color}
\usepackage{dcolumn} 
\usepackage{bm} 
\usepackage{graphicx}
\usepackage{multirow} 
\usepackage{pifont} 
\usepackage{epsfig}
\usepackage[version=4]{mhchem}
\usepackage{array}
\usepackage[pagewise]{lineno}
\usepackage{mathrsfs}
\usepackage{float}
\usepackage{amsfonts}
\usepackage{amsmath} 
\usepackage{amssymb}
\usepackage{subfigure}
\usepackage{mathtools}
\usepackage[mathcal]{euscript}
\usepackage{threeparttable}
\usepackage{enumitem}
\usepackage{caption}
\usepackage{url}
\usepackage{comment}
\usepackage[bf,raggedright]{titlesec}
\usepackage[titletoc,toc,title]{appendix}
\usepackage{xcolor,listings} 
\usepackage[nolist]{acronym}
\usepackage{titlesec}

\titleformat{\paragraph}
{\normalfont\normalsize\bfseries}{\theparagraph}{1em}{}
\titlespacing*{\paragraph}
{0pt}{3.25ex plus 1ex minus .2ex}{1.5ex plus .2ex}
\DeclareMathOperator*{\argmax}{argmax}

\newcommand{\etal}{\emph{et~al.}}
\newcommand{\etc}{\emph{etc.}}
\newcommand{\ie}{\emph{i.e.}}
\newcommand{\eg}{\emph{e.g.}}

\newcommand{\si}{\text{Supporting Information}}

\newcommand{\bv}{\mathbf{v}}
\newcommand{\bx}{\mathbf{x}}

\newcommand{\mr}{\multirow}

\author{Mohammad Mostafanejad}
\email{smostafanejad@vt.edu}
\affiliation{Department of Chemistry, Virginia Tech, Blacksburg, Virginia 24061, USA}
\alsoaffiliation{Molecular Sciences Software Institute, Blacksburg, Virginia 24060, USA}
\author{Paul Saxe}
\email{psaxe@vt.edu}
\affiliation{Department of Chemistry, Virginia Tech, Blacksburg, Virginia 24061, USA}
\alsoaffiliation{Molecular Sciences Software Institute, Blacksburg, Virginia 24060, USA}
\author{T. Daniel Crawford}
\email{crawdad@vt.edu}
\affiliation{Department of Chemistry, Virginia Tech, Blacksburg, Virginia 24061, USA}
\alsoaffiliation{Molecular Sciences Software Institute, Blacksburg, Virginia 24060, USA}
\title{BERTology of Molecular Property Prediction}
\begin{document}
\acresetall	
\begin{abstract}
    \noindent
    Chemical language models (CLMs)\acused{CLM} have emerged as promising
    competitors to popular classical \acl{ML} models for \ac{MPP} tasks.
    However, an increasing number of studies have reported inconsistent and
    contradictory results for the performance of \acp{CLM} across various
    \ac{MPP} benchmark tasks. In this study, we conduct and analyze hundreds of
    meticulously controlled experiments to systematically investigate the
    effects of various factors, such as dataset size, model size, and
    standardization, on the pre-training and fine-tuning performance of
    \acp{CLM} for \ac{MPP}. In the absence of well-established scaling laws for
    encoder-only masked language models, our aim is to provide comprehensive
    numerical evidence and a deeper understanding of the underlying mechanisms
    affecting the performance of \acp{CLM} for \ac{MPP} tasks, some of which
    appear to be entirely overlooked in the literature.
\end{abstract}
%
\maketitle
%
\acresetall	
%
Molecular property prediction (MPP)\acused{MPP} is a fundamental task in
materials design and drug discovery campaigns which involves using computational
models to predict the physicochemical properties of chemical compounds from
their molecular
features.\cite{David:2020:56,Deng:2023:6395,Wigh:2022:e1603,Musil:2021:9759} In
order to be able to significantly accelerate the computational simulations and
reduce the cost of experimental procedures in discovery workflows, \ac{MPP}
models should overcome two major challenges:\cite{Wouters:2020:844} the scarcity
of large-scale high-quality annotated datasets in
chemistry,\cite{Editorial:2023:438} and the complexity of finding an effective
molecular representation that can capture the underlying physicochemical
phenomena governing the target
properties.\cite{David:2020:56,Deng:2023:6395,Wigh:2022:e1603,Musil:2021:9759}

For decades, classical \ac{ML} models have been widely used for \ac{MPP} tasks,
where the molecular structures are represented using expert-engineered molecular
descriptors. However, the presence of label noise, the absence of
standardization and the lack of expertise in feature engineering can harm the
generalization capabilities of the models.\cite{Kolmar:2021:92,Deng:2023:6395}
Alternatively, feed-forward neural networks\cite{Pang:2023:395} can learn
complex molecular features directly from the data. Nonetheless, they often
require large amounts of labeled data for training and can be prone to
overfitting.

Recent advances in \ac{NLP}, especially the introduction of
Transformers,\cite{Vaswani:2017:5999} have contributed to the development of
\acp{CLM} for \ac{MPP} tasks. Transformers treat the textual representations of
molecules, such as \ac{SMILES}\cite{Weininger:1988:31}, as the language of
chemistry and learn its underlying structure, rules, syntax and semantics via
language modeling objectives. Although Transformers are extremely effective in
parallel processing of long-range dependencies in sequences, they are limited by
their uni-directional left-to-right
self-attention\cite{Bahdanau:2014:ARXIV.1409.0473v7} and auto-regressive
training objectives.

The aforementioned limitation of Transformer architecture sparked the
development of \ac{BERT}\cite{Devlin:2019:4171} along with a two-step \ac{SSL}
paradigm which promotes the training of deep bidirectional encoder-only models
for language understanding tasks. The \ac{SSL} begins with training a \ac{LLM}
on a vast corpus of unlabeled data to learn the underlying structure and
semantics of the language at a high level. The pre-trained foundation model is
then fine-tuned on a smaller annotated dataset to adapt its learned
representations towards specific requirements of the downstream
task.\cite{Devlin:2019:4171} The fine-tuned models can then be converted into
\acp{SLM} using compression and optimization techniques such as knowledge
distillation\cite{Hinton:2015:ARXIV.1503.02531v1}, pruning and quantization to
improve their computational efficiency and reduce their memory requirements for
deployment in resource-constrained
environments.\cite{Han:2016:ARXIV.1510.00149v5,Li:2020:5914}

The success of \ac{BERT} in \ac{NLP} has inspired the development of a slew of
\acp{CLM} for \ac{MPP}.\cite{Wang:2019:429,Honda:2019:ARXIV.1911.04738v1,
Fabian:2020:ARXIV.2011.13230v1,Maziarka:2020:ARXIV.2002.08264v1,
Karpov:2020:17,Li:2021:7181815,Xue:2022:899,Wu:2022:1,
Chithrananda:2020:ARXIV.2010.09885v2,Ahmad:2022:ARXIV.2209.01712v1,
Ross:2022:015022,Wen:2022:71,Ross:2022:1256,Soares:2023:ARXIV.2306.14919v1,
Zhang:2023:1216765,Wu:2021:5312} However, an increasing number of studies
involving \acp{CLM} have reported performance results that are inconsistent and
contradictory across various \ac{MPP} benchmark tasks. For instance, several
recent
studies\cite{Chen:2021:10793,Ross:2022:1256,Ahmad:2022:ARXIV.2209.01712v1}
reported that the performance of multiple \ac{BERT}-based \acp{CLM} for \ac{MPP}
tasks can deteriorate as the size of the pre-training dataset increases. Other
similar cases have been documented in a recent review.\cite{Sultan:2024:6259}

The scaling laws of \acp{LLM} establish an empirical power-law relation between
the testing performance of auto-regressive language models and factors such as
the model size, dataset size and the amount of compute used for
training.\cite{Kaplan:2020:ARXIV.2001.08361v1,Hoffmann:2022:ARXIV.2203.15556v1}
Nevertheless, to our knowledge, there is no rigorous framework that extends the
scaling laws to encoder-only models with \ac{MLM} objective. As such, in a quest
to find the source of reported inconsistencies in the literature, we resort to
conducting hundreds of carefully controlled experiments to systematically
explore the impact of elements such as dataset size, model size, tokenization,
model architecture and standardization on the pre-training and fine-tuning
performance of \acp{CLM} for \ac{MPP}. Through this study, we aim to provide a
comprehensive understanding of the underlying mechanisms, backed by numerical
evidence, to shed light on factors that appear to be entirely overlooked in the
literature.
\section*{Results}
\label{SEC:RESULTS}
In the following sections, we investigate the impact of a variety of factors on
the pre-training and fine-tuning performance of \ac{BERT}. Throughout our
analysis, two aspects will frequently come up: the model size and the dataset
size. This is intentional, as we intend to provide evidence pertinent to the
scaling laws of \acp{LLM} with \ac{MLM} objective and demonstrate that the
observed numerical trends remain consistent across all studied experimental
settings.

\subsection*{Model Initialization and Data Sampling Randomness in Pre-training}
\label{SUBSEC:RANDOMNESS} 
Randomness is an inherent part of the pre-training process but seldom receives
much attention in the \ac{MPP} literature. We assess the variability of the
pre-training performance of \ac{BERT} variants with respect to the model weight
initialization and data sampling randomness. For each model variant, we perform
five independent experiments over 10 epochs with different random seeds. All
pre-training runs use the entire 119,184,806 canonical stereoisomeric
\ac{SMILES} entries in the PubChem dataset which is randomly split into
95,347,844 data points ($\sim$ 80\% of the data) for training and 23,836,962
data points for validation ($\sim$ 20\% of the data) sets. The average
pre-training performance metrics, defined in Methods, are reported in
Table~\ref{TAB:PTMODELSEED}. 

\begin{table}
\centering
\setlength{\tabcolsep}{4pt} 
\resizebox{\textwidth}{!}{
\begin{tabular}{lccccc}
\hline\hline
 \textbf{BERT} &  \mr{2}{*}{\textbf{T-Loss} $\downarrow$$^c$} & \mr{2}{*}{\textbf{V-Loss} $\downarrow$$^c$} & \mr{2}{*}{\textbf{V-Acc} $\uparrow$$^c$}  & \mr{2}{*}{\textbf{V-wF1} $\uparrow$$^c$} & \mr{2}{*}{\textbf{V-PPPL} $\downarrow$$^c$}  \\
 \textbf{Variant}  & & & & \\
\hline
\textbf{Tiny}        & 0.6511 $\pm$ 0.0090 & 0.4696 $\pm$ 0.0053 & 0.8823 $\pm$ 0.0077 & 0.8686 $\pm$ 0.0085 & 1.5993 $\pm$ 0.0085 \\
\textbf{Small}       & 0.2353 $\pm$ 0.0009 & 0.1836 $\pm$ 0.0008 & 0.9453 $\pm$ 0.0109 & 0.9412 $\pm$ 0.0116 & 1.2016 $\pm$ 0.0010 \\
  \textbf{Base}$^b$  & 0.1779 $\pm$ 0.0006 & 0.1359 $\pm$ 0.0011 & 0.9595 $\pm$ 0.0046 & 0.9564 $\pm$ 0.0041 & 1.1455 $\pm$ 0.0013 \\
\hline\hline
\end{tabular}
} \caption{Variations of the averaged pre-training performance metrics with 95\%
confidence intervals$^a$ for masked language modeling with respect to the model
weight initialization and data sampling seeds}
\label{TAB:PTMODELSEED}
\begin{tablenotes}
    \footnotesize 
    \item $^a$ The statistical summaries ($\bar{x} \pm (s_N/\sqrt{N}) t_{95\%,
    N-1} $) for each performance metric is calculated over five runs with
    different data sampling and model initialization seeds. The critical value
    of the two-sided $t$-distribution with 95\% confidence interval is taken from
    Ref.~\citenum{NIST:2012:STATHANDBOOK}.
    \item $^b$ A diverged model training session was excluded from the mean and
    standard deviation calculations which persisted over several runs. For more
    details, see the Table 1 in the Extended Data.
    \item $^c$ $\uparrow$: higher is better; $\downarrow$: lower is better.
\end{tablenotes}
\end{table}

The results in Table \ref{TAB:PTMODELSEED} demonstrate that the variations in
the performance of the models, due to randomness in data sampling and model
initialization, are small ($\approx$1\%) compared to those caused by the model
size. Specifically, increasing the model size from Tiny-\ac{BERT} to
Base-\ac{BERT} decreases the average training (validation) loss values from
0.6511 $\pm$ 0.0090 (0.4696 $\pm$ 0.0053) to 0.1779 $\pm$ 0.0006 (0.1359 $\pm$
0.0011) by $>$70\%, decreases the average pseudo-perplexity (V-PPPL) values by
more than 28\% from 1.5993 $\pm$ 0.0085 to 1.1455 $\pm$ 0.0013, and increases
the average validation accuracy, V-Acc, (weighted-F1 score, V-wF1) by more
than 8\% from 0.8823 $\pm$ 0.0077 (0.8686 $\pm$ 0.0085) to 0.9595 $\pm$ 0.0046
(0.9564 $\pm$ 0.0041), respectively. 

\subsection*{Standardization Effects on Pre-training}
\label{SUBSEC:STDEFFECT} 
\begin{figure}[!htbp]
    \setlength{\abovecaptionskip}{0.15cm}
    \centering
    \subfigure[]{
        \includegraphics[width=0.36\textwidth]{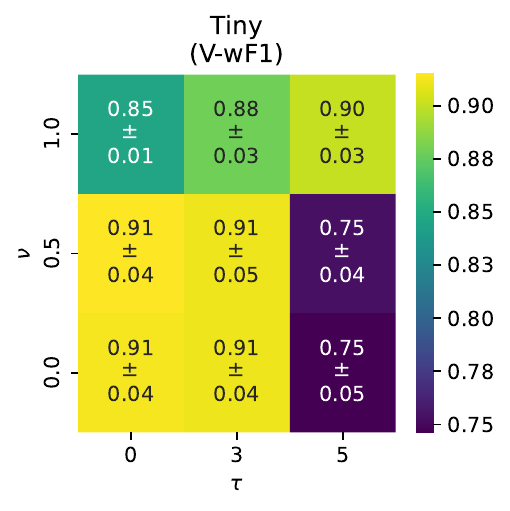}
    }
    \subfigure[]{
        \includegraphics[width=0.36\textwidth]{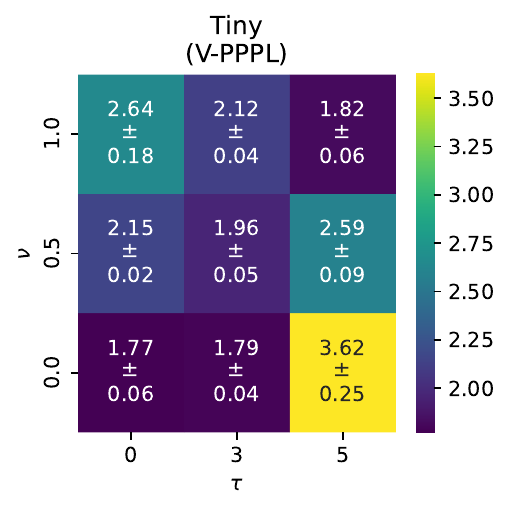}
    }
    \subfigure[]{
        \includegraphics[width=0.36\textwidth]{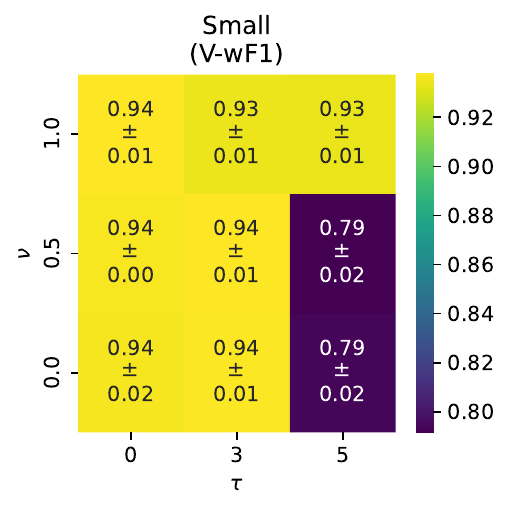}
    }    
    \subfigure[]{
        \includegraphics[width=0.36\textwidth]{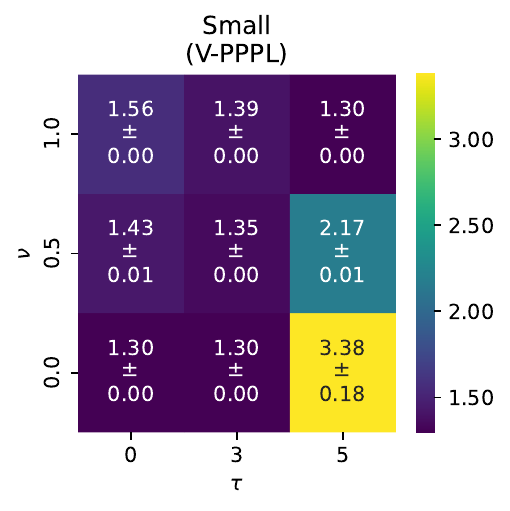}
    }
    \subfigure[]{
        \includegraphics[width=0.36\textwidth]{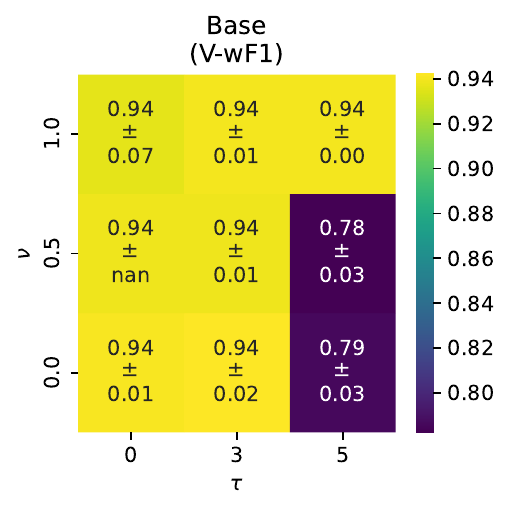}
    }
    \subfigure[]{
        \includegraphics[width=0.36\textwidth]{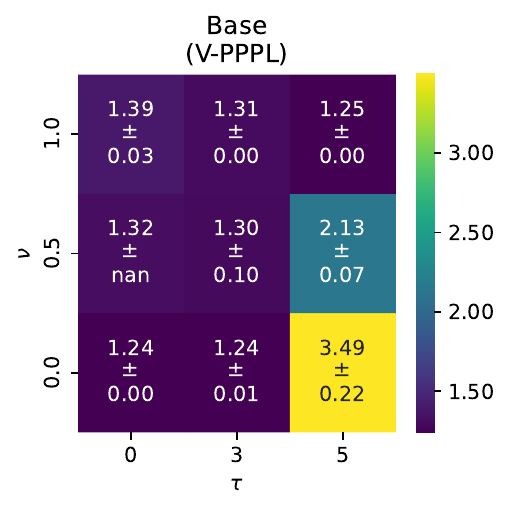}
    } \caption{The effect of standardization noise on the pre-training
    validation weighted-F1 score (a,c,e) and pseudo-perplexity (b,d,f) of BERT
    for masked language modeling. Both metrics are averaged over three
    independent runs with different data sampling and model initialization
    seeds. $\tau$ and $\nu$ control the standardization noise in the
    pre-training and validation splits, respectively. The error bars are based
    on 95\% confidence interval which are not defined (NaN) for samples of $N=1$
    runs, if the other two have diverged and removed from calculating the
    statistical summaries.}
    \label{FIG:PTSTDEFFECT}
\end{figure}

We hypothesize that combining \ac{SMILES} from different chemical databases may
amount to mixing different standardization protocols which can confuse the model
during pre-training and lead to degraded performance. In order to simulate the
impact of standardization noise on the pre-training performance of \ac{BERT}, we
gradually replace various percentages of the PubChem \ac{SMILES} in the training
and validation splits with their corresponding ChEMBL-standardized counterparts.
The \ac{SMILES} corruption percentages in the training and validation splits are
controlled by the parameters $\tau$ and $\nu$, respectively
(Fig.~\ref{FIG:PTSTDEFFECT}) according to the recipe, described in the \si.
Briefly, the standardization noise in the training split can change between pure
PubChem ($\tau$=0) and pure ChEMBL ($\tau$=5) and similarly, in the validation
split between ($\nu$=0.0) and ($\nu$=1.0), respectively.

Figure \ref{FIG:PTSTDEFFECT} illustrates the variations of average V-wF1 and
V-PPPL versus the standardization noise in the training and validation splits.
Here, increasing the percentages of ChemBL-standardized \ac{SMILES} in each
split increases the average value of V-PPPL for all variants of \ac{BERT}. For
instance, the V-PPPL for Tiny-\ac{BERT} increases from 1.7715 $\pm$ 0.0587 to
3.6218 $\pm$ 0.2487 and 2.6416 $\pm$ 0.1838 when \ac{SMILES} in the pure PubChem
($\tau$=$\nu$=0) dataset are completely replaced by their ChemBL-standardized
counterparts in the training ($\tau$=5, $\nu$=0) or validation splits ($\tau$=0,
$\nu$=1.0), respectively. In an extreme case where the entire training data is
replaced with ChEMBL-standardized \ac{SMILES} ($\tau$=5, $\nu$=0),
Tiny-\ac{BERT} shows signs of divergence which triggers the early stopping
mechanism after a few epochs, in all three independent runs (see \si\ for more
details). Therefore, the model's ability to predict the masked tokens in the
input sequences can be severely hampered by the standardization noise when the
model is trained and validated on \ac{SMILES} with mixed standardization
protocols. This observation is consistent with the observed degradation in V-wF1
score as the \ac{SMILES} standardization noise in the training and validation
splits increases. Other performance metrics such as V-Loss and V-Acc show
similar trends and are presented in Fig.~1 of the Extended Data.

Figure \ref{FIG:PTSTDEFFECT} also demonstrates that larger models become more
resilient to the standardization noise, as evidenced by the V-wF1 and V-PPPL
values, going from the top to bottom row. The impact of the standardization
noise on the model performance can be minimized via taking the ``path of least
desruction'' (the diagonal parts of the heatmaps in Fig.~\ref{FIG:PTSTDEFFECT})
where the standardization noise is gradually added to both training and
validation splits, simultaneously. Here, we assume the \ac{SMILES}, added to both
splits, are generated by the same or similar data distributions.

\subsection*{The Effect of Tokenization on Pre-training}
\label{SUBSEC:TOKENIZATIONEFFECT} 
The tokenization process is a crucial step in the pre-training of \acp{LLM} for
\ac{MPP} tasks as it determines how the input sequences are processed by the
models and what their vocabulary composition will be. In this study, we
investigate the impact of WordPiece\cite{Wu:2016:ARXIV.1609.08144v2} and
\ac{BPE}\cite{Sennrich:2016:1715} tokenization algorithms on the pre-training
performance of \ac{BERT}. Both algorithms start with a base vocabulary of
individual characters and iteratively apply the learned merging rules to form
new tokens until a pre-defined vocabulary size is reached. The main difference
between the two algorithms is that the WordPiece algorithm uses a
likelihood-based criterion to select the subword units for merging while
\ac{BPE} relies on a frequency-based
criterion.\cite{Wu:2016:ARXIV.1609.08144v2,Sennrich:2016:1715} Regardless of the
selected tokenization method, the average metric values improve as the model
size increases from Tiny-\ac{BERT}, to Base-\ac{BERT}. For instance, the
magnitude of V-PPPL decreases from 1.5978 $\pm$ 0.0138 (1.5435 $\pm$ 0.0332) to
1.1450 $\pm$ 0.0052 (1.1233 $\pm$ 0.0125) using WordPiece (\ac{BPE}) tokenizer--
an improvement of 28\% for WordPiece and 27\% for \ac{BPE}. As the sample size
(\ie, the number of experiments) for each model variant is small ($N \leq 3$),
we refrain from making any judgments on the statistical significance based on
the estimated \ac{CI} and choose to proceed with WordPiece as our tokenizer,
consistent with the original \ac{BERT} model.\cite{Devlin:2019:4171} For further
details on the tokenization experiments, see Table 2 in the Extended Data.

\subsection*{The Effect of Dataset and Model Sizes on Pre-training}
\label{SUBSEC:DATAMODELSIZEEFFECT}
In order to study the effect of dataset size on the pre-training performance, we
create six dataset bins with the number of training samples in each bin
following an exponential expression of the form
\begin{equation}\label{EQ:DATASIZE}
    N_\text{train} = a \times 2^{k} + b,
\end{equation}
where the bin indices, $k=$ 0, 1, 2, 3, 4 and 5, correspond to 2.5\%, 5\%, 10\%,
20\%, 40\% and 80\% of the data, respectively. Here, $a =$ 2,979,620 fixes the
size of the first bin and $b$ is the correction factor, which ensures that the
addition of the sixth bin ($k=5$) will exactly cover $80\%$ of the PubChem
dataset. As such, $b = 4$ when $k = 5$ and zero otherwise.

\begin{figure}[!tbph]
    \centering
    \includegraphics{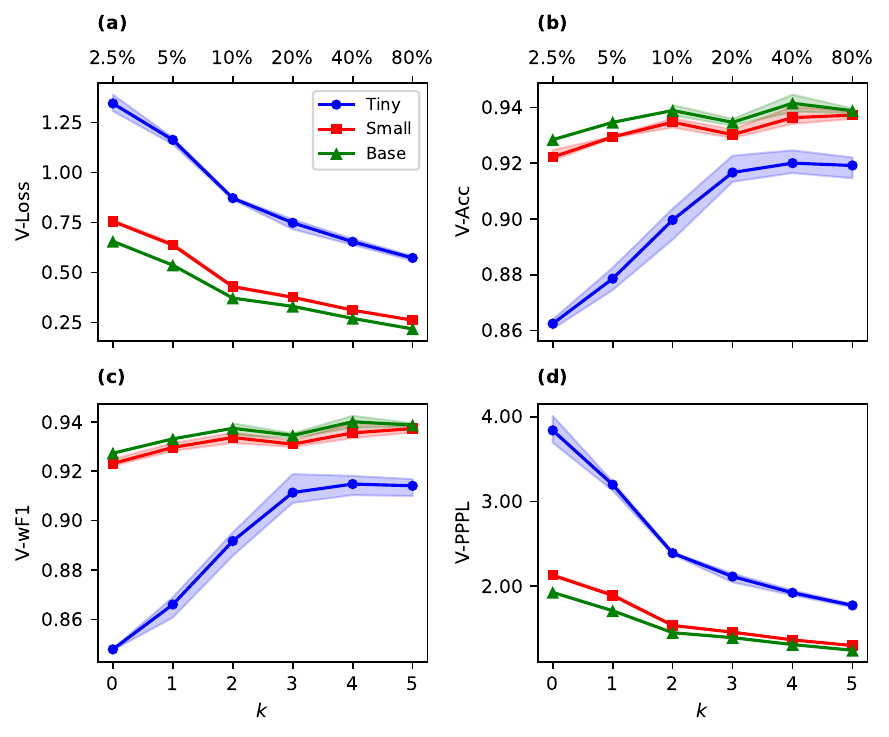}
    \caption{Variations of BERT's pre-training performance metrics: (a)
    validation loss (V-Loss), (b) accuracy (V-Acc), (c) weighted-F1 score
    (V-wF1), and (d) pseudo-perplexity (V-PPPL) versus the dataset size bin
    index, $k$ (the corresponding data percentages are shown on the top axes).
    Results pertinent to the Tiny-, Small- and Base-BERT are shown in blue
    circles, red squares and green triangles, respectively. Shaded bands are
    used to indicate the 95\% confidence intervals across three independent runs
    with different model initialization and data sampling random seeds.}
    \label{FIG:DATASIZEPERF}
\end{figure}

Having a coherent standardization protocol in place, we expect the pre-training
performance to improve as the dataset and model sizes
increase.\cite{Kaplan:2020:ARXIV.2001.08361v1,Hoffmann:2022:ARXIV.2203.15556v1}
This is indeed the case as shown in Fig.~\ref{FIG:DATASIZEPERF}. For all three
variants of \ac{BERT}, the average pre-training V-Loss decreases as the dataset
size increases (Fig.~\ref{FIG:DATASIZEPERF}a). For each bin index $k$, the
magnitude of the average V-Loss also decreases as the model size increases from
Tiny-\ac{BERT} to Base-\ac{BERT}. Similar trends are observed for V-PPPL
(Fig.~\ref{FIG:DATASIZEPERF}d) which suggest that larger models are more
effective at learning the syntax and semantics of the language of chemistry via
\ac{MLM} pre-training.

It is important to note that both V-Loss and V-PPPL show a sudden change in the
slope of their diagrams at around $k=2$ which corresponds to training on 10\% of
the data ($\approx$12 million samples). This change can be an indication of a
critical threshold in dataset size, beyond which the performance improvements
start to plateau. Furthermore, the performance gap is more pronounced at smaller
dataset sizes, especially between Tiny- and Small-\ac{BERT} variants, but it
tends to diminish as the dataset size increases. This suggests that larger
models are more sample efficient and can achieve better performance with smaller
samples of data compared to their smaller size counterparts, which is consistent
with previous studies.\cite{Kaplan:2020:ARXIV.2001.08361v1,
Hoffmann:2022:ARXIV.2203.15556v1} Figure \ref{FIG:DATASIZEPERF}b and c
illustrate the variations of V-Acc and V-wF1 with respect to the dataset size.
The magnitude of both metrics increases as the dataset size increases, with the
performance gap between the different model variants being more noticeable at
smaller dataset sizes, especially between Tiny- and Small-\ac{BERT} variants.
However, the aforementioned performance gap diminishes as the dataset size
increases. At around $k=3$, corresponding to training on 20\% of the data
($\sim$24 million samples), both V-Acc and V-wF1 for Tiny-\ac{BERT} show a
sudden change in the slope of their corresponding diagrams, after which the
performance improvements starts to plateau. Notably, the magnitude of V-Acc and
V-wF1 for both Small- and Base-\ac{BERT} variants show a small dip at around
$k=3$ but continue to increase on average, as the dataset size increases.

\subsection*{Model and Dataset Size Effects on Fine-tuning}
\label{SUBSEC:DATASIZEONFINETUNING}
In this section, we use Biogen's \ac{ADME} public dataset to investigate the
impact of pre-training dataset and model sizes on the fine-tuning performance of
\ac{BERT} for downstream \ac{MPP} regression tasks.\cite{Fang:2023:3263} The
fine-tuning is performed using 3-fold cross-validation on the training split
where 20\% of data was set aside for testing and the remaining 80\% was split
into three folds, with two folds used for training and one fold used for
validation within three iterations. A Bayesian hyperparameter search over 50
model architectures is performed on each fold using foundation models, which
were trained on 2.5\%, 20\% and 80\% (or $k$=0, 3, and 5) of the PubChem data.
The best model from each search is selected based on the validation $R^2$
metric, following Ref.~\citenum{Fang:2023:3263}. For each $k$, the top
performing cross-validated model is then refitted on the entire (training and
validation) fine-tuning data and evaulated on a test set. The cross-validation
outcomes for the \ac{HLM}, \ac{HPPB} and solubility endpoints are shown in
Figs.~2--4 of Extended Data.

The average 3-fold cross-validation performance metrics for the \ac{HLM} and
\ac{HPPB} endpoints indicate that the Pearson $R$ and $R^2$ increase for all
variants of \ac{BERT} as the pre-training dataset size increases. The opposite
trends are observed for the regerssion error metrics where the average \ac{MAE}
and \ac{RMSE} values decrease as the pre-training dataset size increases. These
trends highlight the importance of pre-training dataset size for improving the
fine-tuning performance of \ac{BERT} on the downstream tasks. Furthermore, for
each pre-training dataset size (bin index $k$), the performance improves as the
model size increases from Tiny-\ac{BERT} to Base-\ac{BERT}. Note that the
cross-validation results for the solubility endpoint are significantly impacted
by the large standard deviations ($\approx$0.68) and the strongly skewed nature
of the distribution of the experimental data.\cite{Fang:2023:3263} As such,
providing a fair assessment of the cross-validation results for the solubility
endpoint is challenging.

\begin{figure}[!tbph]
    \centering
    \includegraphics[]{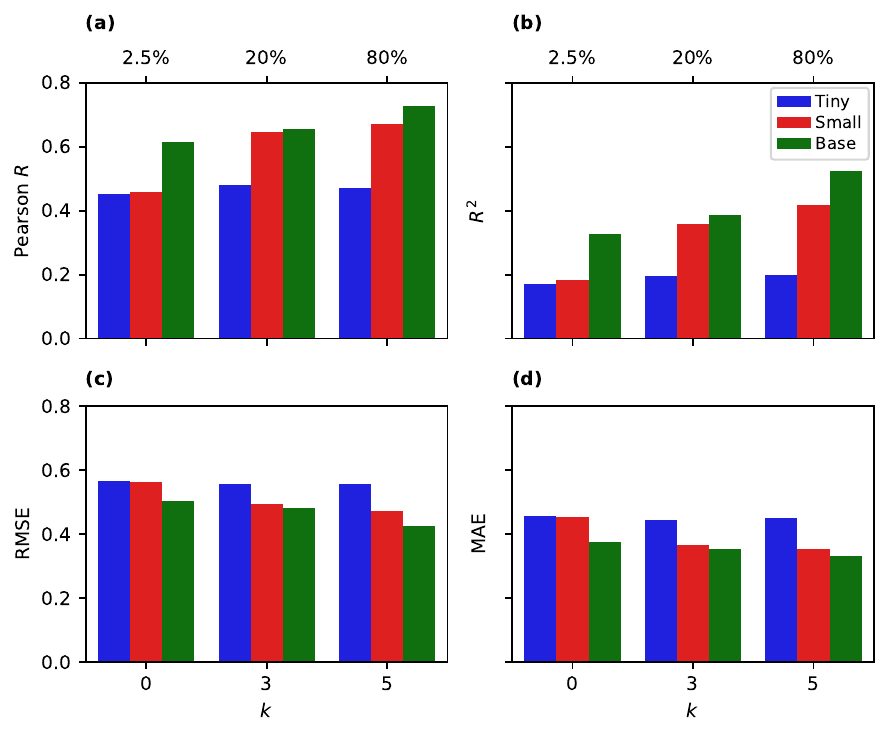}
    \caption{Variations of the fine-tuned BERT's testing performance
    metrics, (a) Pearson $R$, (b) $R^2$, (c) RMSE, (d) MAE, versus the
    pre-training dataset size bin index, $k$, for the HLM endpoint (the
    corresponding data percentages are shown on the top axes)}
    \label{FIG:SFTHLMTEST}
\end{figure}

\begin{figure}[!tbph]
    \centering
    \includegraphics[]{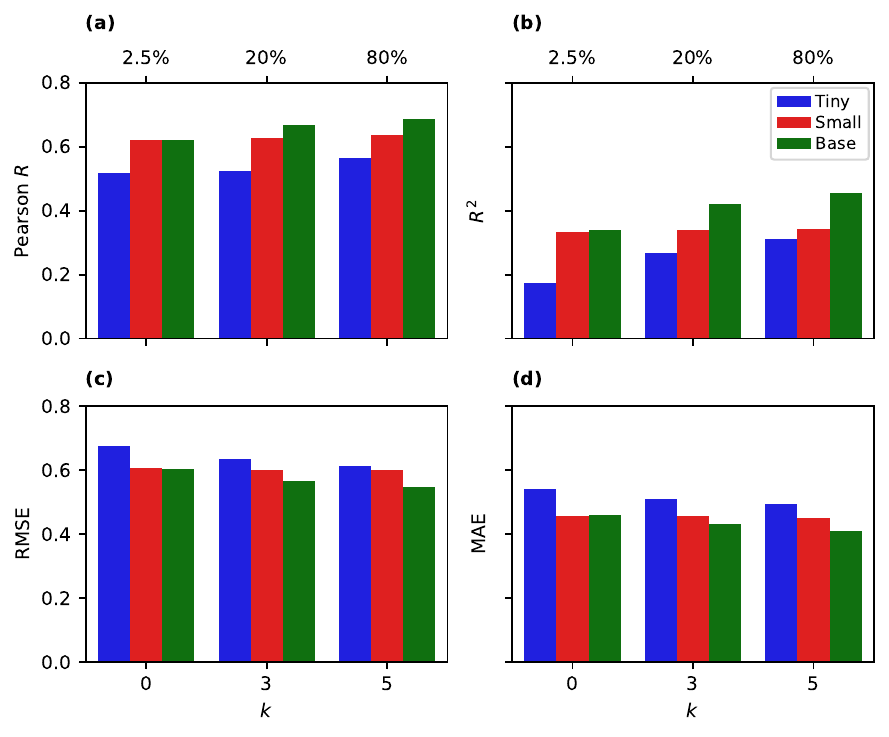}
    \caption{Variations of the fine-tuned BERT's testing performance metrics,
    (a) Pearson $R$, (b) $R^2$, (c) RMSE, (d) MAE, versus the pre-training
    dataset size bin index, $k$, for the hPPB endpoint (the corresponding data
    percentages are shown on the top axes)}
    \label{FIG:SFTHPPBTEST}
\end{figure}

\begin{figure}[!tbph]
    \centering
    \includegraphics[]{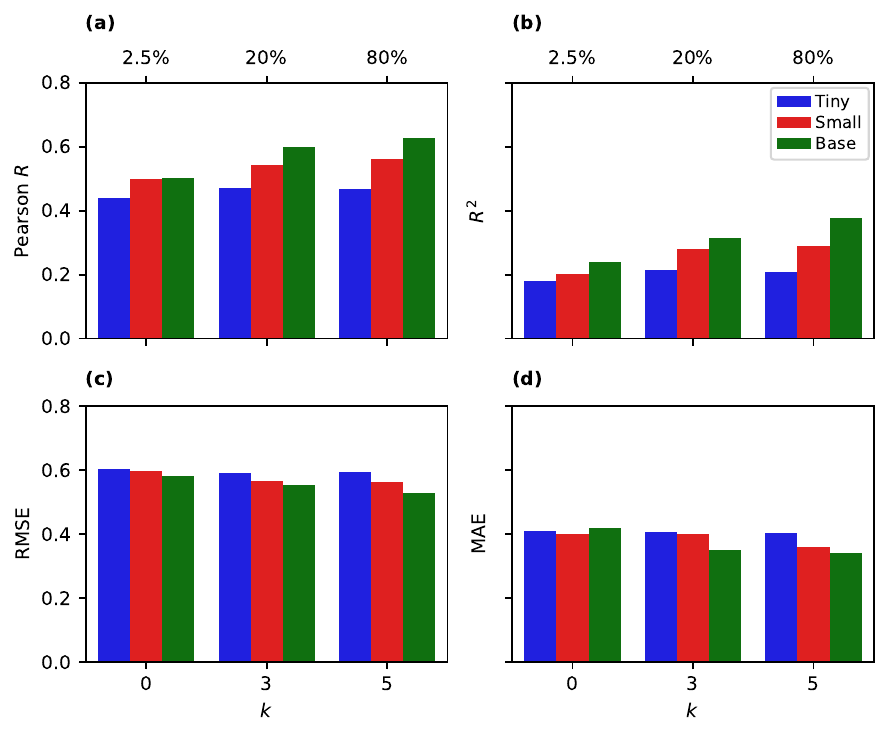}
    \caption{Variations of the fine-tuned BERT's testing performance metrics,
    (a) Pearson $R$, (b) $R^2$, (c) RMSE, (d) MAE, versus the pre-training
    dataset size bin index, $k$, for the solubility endpoint (the corresponding
    data percentages are shown on the top axes)}
    \label{FIG:SFTHLMSOLTEST}
\end{figure}

The testing performance results for \ac{BERT} on the \ac{HLM}, \ac{HPPB} and
solubility endpoints are presented in
Figs.~\ref{FIG:SFTHLMTEST}--\ref{FIG:SFTHLMSOLTEST}. The test Pearson $R$ and
$R^2$ values for all variants of \ac{BERT} increase as the pre-training dataset
size increases. The opposite trends are observed for the regression error
metrics where the average \ac{MAE} and \ac{RMSE} values decrease as the
pre-training dataset size increases. These trends further confirm the importance
of pre-training dataset size for improving the fine-tuning performance of
\ac{BERT} on the downstream tasks and are consistent with the empirical scaling
laws of \acp{LLM}.\cite{Kaplan:2020:ARXIV.2001.08361v1,
Hoffmann:2022:ARXIV.2203.15556v1} Here, caution should be exercised in making
direct comparisons due to the differences in the model architecture and the
training objectives.

We extend the fine-tuning results to all endpoints using \ac{BERT} foundation
models that are trained on 80\% ($k$=5) of PubChem data. For comparison, we have
also fine-tuned a set of classical \ac{ML} models such as \ac{LASSO}, \ac{RF},
\ac{SVM}, \ac{XGB}, and \ac{LGBM} on the same training and validation splits
using the fine-tuning procedure mentioned above. The only caveat is that we used
grid search instead of Bayesian search for the hyperparameter optimization of
the classical models to be consistent with the training process adopted in
Ref.~\citenum{Fang:2023:3263}.

The fine-tuning results (Tables 4--9 of the \si) demonstrate that the testing
performance of Base-\ac{BERT} is superior or similar to those of the classical
\ac{ML} models for all six endpoints. The \ac{HPPB} and \ac{RPPB} endpoints show
testing performance results for Base-\ac{BERT} that are competitive to those of
classical \ac{ML} models. Unfortunately, both endpoints have the smallest sample
sizes (1808 and 885 datapoints before preprocessing, respectively) of mixed
sources (ChEMBL and Biogen) and have two of the largest standard deviations in
their experimental values among all six endpoints ($>$0.6), which makes it
challenging for the models to generalize well on even smaller test sets (20\% of
the total dataset size).\cite{Fang:2023:3263}

\section*{Discussion}
\label{SEC:DISCUSSION}
Recent results,\cite{Chen:2021:10793,Ross:2022:1256,
Ahmad:2022:ARXIV.2209.01712v1,Sultan:2024:6259} focusing on the pre-training and
fine-tuning of \acp{LLM} on \ac{MPP} downstream tasks, show trends that are
inconsistent and contradictory. For example, Chen \etal\cite{Chen:2021:10793}
pre-trained a variant of \ac{BERT} on three combinations of \ac{SMILES} from
ChEMBL (\url{https://www.ebi.ac.uk/chembl}), PubChem
(\url{https://pubchem.ncbi.nlm.nih.gov}) and ZINC
(\url{https://zinc.docking.org}) databases. The three sets involve 1,941,410
compounds from ChEMBL, 103,395,400 compounds from ChEMBL and PubChem, and
775,007,514 compounds from ChEMBL, PubChem and ZINC. The resulting foundation
models were subsequently fine-tuned on 10 downstream regression and
classification \ac{MPP} tasks from MoleculeNet.\cite{Wu:2018:515} Surprisingly,
the model pre-trained on the smallest dataset outperformed the other models in 7
out of 10 tasks. Similar observations were documented in a recent
review.\cite{Sultan:2024:6259}

In the absence of well-established scaling laws for encoder-only models with a
\ac{MLM} objective, we resort to conducting hundreds of carefully controlled
experiments to systematically investigate how factors such as dataset size,
model size, tokenization, model architecture and standardization can influence
the performance of \ac{CLM} for \ac{MPP}.

Our results suggest that choosing different random seeds for model
initialization and data sampling (Table \ref{TAB:PTMODELSEED}) or choosing
between WordPiece and \ac{BPE} tokenization algorithms (Table 2 of Extended
Data) have minor effects on the pre-training performance of \ac{BERT} compared
to the choice of model size and dataset size. Both observations are consistent
with the scaling laws of auto-regressive
\acp{LLM}.\cite{Kaplan:2020:ARXIV.2001.08361v1,
Hoffmann:2022:ARXIV.2203.15556v1}

Our experiments highlight the importance and overarching role of datasets,
standardization protocols, and experimental settings in pre-training and
fine-tuning performance of \ac{BERT} for \ac{MPP} tasks. We choose PubChem for
pre-training our models as it is one of the largest publicly available
general-purpose chemical databases with over 119 million unique compounds with
their associated properties. PubChem offers a user-friendly interface for
accessing, searching and downloading the data. Furthermore, PubChem provides a
rigorous standardization pipeline\cite{Hahnke:2018:36} which can mitigate
potential data quality and consistency issues observed in non-standardized
databases. 

As a crucial step in data preprocessing, standardization transforms the
\ac{SMILES} representation of chemical structures into a consistent and unique
form that is aligned with the intended domain-specific applications of the
underlying databases. We hypothesize that studies\cite{Chen:2021:10793,
Ross:2022:1256,Ahmad:2022:ARXIV.2209.01712v1,Sultan:2024:6259} which combine
\ac{SMILES} from various chemical databases to create bigger pre-training sets
to improve the performance of their models on the downstream tasks, may have
inadvertently introduced standardization noise by mixing different
standardization protocols. We test this hypothesis by replacing various
percentages of the PubChem \ac{SMILES} in the training and validation splits
with their corresponding ChEMBL-standardized counterparts and study its impact
on the pre-training performance of \ac{BERT} via V-loss, V-Acc, V-wF1 and V-PPPL
metrics (see Fig.~\ref{FIG:PTSTDEFFECT}, Extended Data and \si). Figure
\ref{FIG:PTSTDEFFECT} demonstrates that the standardization noise can severly
degrade the pre-training performance of \ac{BERT}. Furthermore, larger models
tend to be more resilient to the standardization noise, especially if the
noise, added to both training and validation splits, follows the ``path of least
destruction'' (the diagonal parts of the heatmaps in Fig.~\ref{FIG:PTSTDEFFECT})
while its distribution in both splits does not drastically change.

The choice of benchmark dataset is another important factor. Unfortunately, the
majority of the studies involving \acp{CLM} have adopted the MoleculeNet dataset
for fine-tuning and evaluation purposes. MoleculeNet is a collection of 17
datasets across four domains of quantum chemistry, physical chemistry,
biophysics, and physiology. This dataset includes both regression and
classification objectives and is designed for benchmarking \ac{ML} models for
\ac{MPP}.\cite{Wu:2018:515} Despite its popularity, MoleculeNet suffers from
major issues such as invalid structures, inconsistent chemical representations,
ambiguous stereochemistry, duplicate entries with conflicting property values
\etc\cite{Walters:2023:WEBSITE} Such issues are not specific to MoleculeNet and
can be found in other popular datasets such as \ac{TDC}, accessible from
\url{https://tdcommons.ai}. 

Here, we use the curated Biogen \ac{ADME} dataset for fine-tuning and
evaluating our models on six \ac{ADME} regression endpoints: \ac{HLM}, \ac{RLM},
\ac{HPPB}, \ac{RPPB}, \ac{MDR1} and solubility.\cite{Fang:2023:3263} We
standardize all \ac{SMILES} using PubChem's standardization protocol and
tokenize them using the WordPiece tokenizer. Our 3-fold cross-validation
hyperparameter search results indicate that the testing performance of \ac{BERT}
on the studied \ac{ADME} tasks
(Figs.~\ref{FIG:SFTHLMTEST}--\ref{FIG:SFTHLMSOLTEST}) consistently improves as
the pre-training dataset and model sizes increase. These trends are consistent
with the scaling laws of auto-regressive
\acp{LLM},\cite{Kaplan:2020:ARXIV.2001.08361v1,
Hoffmann:2022:ARXIV.2203.15556v1} although not directly transferrable to
encoder-only models with \ac{MLM} objectives. 

Additional cross-validation and testing results with \ac{BERT} foundation
models, pre-trained on all 80\% ($k$=5) of the PubChem data, demonstrate that
the testing performance of Base-\ac{BERT} is superior or similar to those of the
classical \ac{ML} models for all \ac{ADME} endpoints. Judging by the differences
between the average error metrics of the best performing models, the choice of
pre-training and fine-tuning a \ac{CLM} for downstream \ac{MPP} tasks becomes an
economical one. That is, depending on the task at hand and the available
resources, one should weigh the benefits of using a \ac{CLM} against the costs
and complexities associated with pre-training and fine-tuning it. While
classical \ac{ML} models can be trained and fine-tuned much faster and with less
computational resources, \acp{CLM} often require significant computational
resources and time for pre-training. Meanwhile, foundation \acp{CLM} possess a
general understanding of the chemical language and can be fine-tuned on a wide
range of downstream tasks with relatively small labeled datasets and affordable
training times, often with a performance boost compared to their classical
counterparts.

In summary, we have provided numerical evidence from hunderds of experiments
which favor the potential existence of a scaling law for encoder-only \acp{CLM}
with \ac{MLM} objectives. Our results highlight the importance of (pre-training,
fine-tuning and benchmarking) datasets, standardization protocols and
experimental settings. We demonstrate that larger high-quality pre-training
datasets processed by well-defined and coherent standardization protocols lead
to better pre-training and fine-tuning performance on downstream \ac{MPP} tasks.
We have also delineated that larger models are more sample efficient, noise
resilient and performant in both pre-training and fine-tuning settings. While
our analyses are based on qualitative trends observed within small statistical
samples of experiments, the extraction of exact quantitative relations requires
a more comprehensive set of experimetns which will be the subject of our ongoing
and future studies.

\section*{Methods}
\label{SEC:METHODS}

\subsection*{Statistical Analysis of the Results}
\label{SUBSEC:STATS}
Throughout this work, in order to account for the data variability and to
provide an estimate for a range of plausible values for the true population
metrics, we often accompany our measurements of the property $x$ of a sample of
size $N$, with a 95\% \ac{CI}, which is defined as
\begin{equation}\label{SIEQ:CI}
    \text{CI}(x) = \bar{x} \pm \left( \frac{s_N}{\sqrt{N}} \right) t_{95\%, N-1}.
\end{equation}
Here, $t_{95\%, N-1}$ denotes the critical value from the Student's
$t$-distribution with $N-1$ degrees of freedom at 95\% confidence level. Also,
$\bar{x}$ and $s_N$ are the sample mean and the standard deviation of the
quantity $x$, respectively, which can be expressed as
\begin{equation}
    \bar{x} = \frac{1}{N} \sum_{i=1}^{N} x_i,
    \qquad  \text{and} \qquad
    s_N = \sqrt{\frac{1}{N-1} \sum_{i=1}^{N} (x_i - \bar{x})^2},
\end{equation}
where $x_i$ is the $i$-th input data or measurement outcome. 

Since pre-training \acp{LLM} is very expensive (see \si\ for more details) our
sample sizes are often small where the number of experiments $N=3$ or $5$. As
such, in interpreting the error bars and comparing the results of various
experiments, it is imperative to highlight the adoption of a conventional value
for $t_{95\%, N-1}$ which is often set to 1.96. While this value maybe
appropriate for large sample sizes, care must be taken in calculating the error
bars for very small sample sizes. In such cases, the critical value $t_{95\%,
N-1}$ can be much larger than 1.96, which can lead to a significant
overestimation of the error bars and misinterpretation of the results. In order
to mitigate the impact of small sample sizes on the error bars, we use NIST's
``Critical Values of the Student's t Distribution'' table from
Ref.~\citenum{NIST:2012:STATHANDBOOK} to determine the appropriate $t_{95\%,
N-1}$ for each sample size. Here, we do not provide the Cohen's $d$ or Hedges'
$g$ for calculating the effect size and comparing the results of different
experiments.

\subsection*{Performance and Error Metrics}
\label{SUBSEC:METRICS}
This section summarizes all metrics that are adopted for evaluating the
performance of \ac{BERT} models during pre-training and fine-tuning. Let $\bx^i
= (x^i_1 , \dots , x^i_T)$ be the $i$th sequence of tokens such as word,
wordpieces \etc\ Each token, $x^i_t$, can categorically assume one of the $W$
entries from a vocabulary $\bv = \lbrace v_1, \dots, v_W \rbrace$ which also
includes special entries such as \texttt{[UNK]}, reserved to represent
out-of-vocabulary tokens. We also create a masked variant of the original
sequence, $\bx^i_{\backslash t}$, by randomly replacing a portion (\eg, 15\%) of
the tokens in the sequence with a special token \texttt{[MASK]}. For example,
masking a token at position index $t$, belonging to $\mathcal{M}_i \subseteq
\lbrace 1, \dots, T_i \rbrace$, yields $\bx^i_{\backslash t} = (x^i_1, \dots,
x^i_{t-1}, \texttt{[MASK]}, x^i_{t+1}, \dots, x^i_{T_i})$. The objective of the
\ac{MLM} is to reconstruct the original sequence from its corrupted version by
minimizing the negative pseudo log-likelihood (cross-entropy) loss function
defined as
\begin{equation}\label{EQ:MLM}
    \min_\theta \mathcal{L}(\theta; \mathcal{D}) = - \frac{\sum_{i=1}^{
|\mathcal{D}|} \sum_{t=1}^{T_i} \mathbb{I}_{\mathcal{M}_i}(t) 
    \log p_\theta(x^i_t | \bx^i_{\backslash t})}{\sum_{i=1}^{
|\mathcal{D}|} |\mathcal{M}_i| }.
\end{equation}
Here, the denominator is the total number of masked tokens in the dataset and
the conditional probability $p_\theta(x^i_t | \bx^i_{\backslash t})$ is the
probability of observing the original token $x^i_t$ in the $i$th sequence given
its corrupted version $\bx^i_{\backslash t}$ and model parameters $\theta$. The
indicator function, $\mathbb{I}_{\mathcal{M}_i}(t)$, is equal to 1 if $t \in
\mathcal{M}_i$ and zero otherwise which means, only masked tokens will directly
contribute to the \ac{MLM} loss. The semi-colon in $\mathcal{L}(\theta;
\mathcal{D})$ highlights the difference between model parameters, $\theta$,
treated as variables, and the training data, $\mathcal{D}$ considered as
constants. Note that the pre-training process is an example of self-supervised
learning because the original tokens in each sequence, taken as labels, are the
generators of their masked counterparts in the dataset; \ie, $\mathcal{D} =
\lbrace (\bx^i_{\backslash t}, \bx^i) \rbrace_{i=1}^{|\mathcal{D}|}$.

Using Eq.~\ref{EQ:MLM}, we can express the model's \ac{PPPL} on the
pre-training dataset as\cite{SALAZAR:2020:2699}
\begin{equation}\label{EQ:PPPL}
    \text{PPPL}(\theta; \mathcal{D}) = \exp \left( - 
    \frac{1}{\sum_{i=1}^{|\mathcal{D}|} |\mathcal{M}_i|} \sum_{i=1}^{|\mathcal{D}|} \sum_{t=1}^{T_i} 
    \mathbb{I}_{\mathcal{M}_i}(t) \log p(x^i_t | \bx^i_{\backslash t}) \right),
\end{equation}
Similar to the BART paper,\cite{Lewis:2019:7871} we will also consider the
accuracy and weighted-F1 score in the context of \ac{MLM}. Assuming each entry
in the vocabulary is a class, the accuracy can then be converted into a series
of binary classification tasks in a one-versus-rest fashion and be expressed as
\begin{equation}\label{EQ:ACCURACY}
    \text{Acc}(\theta; \mathcal{D}) = 
    \frac{1}{\sum_{i=1}^{|\mathcal{D}|} |\mathcal{M}_i|} \sum_{i=1}^{|\mathcal{D}|} \sum_{t=1}^{T_i} \mathbb{I}_{\mathcal{M}_i}(t)
    \left\lbrack \mathbb{I}\left(\hat{x}^i_t = v_w, x^i_t = v_w\right) +
    \mathbb{I}\left(\hat{x}^i_t \neq v_w, x^i_t \neq v_w\right)\right\rbrack,
\end{equation}
where the unmasked reference token, $v_w \in \bv$, denotes the positive class
and $\hat{x}^i_t = \underset{x^i_t \in \bv}{\argmax}\ p_\theta( x^i_t|
\bx^i_{\backslash t})$, is the action of choosing the most probable token,
predicted by the model, from the vocabulary. In Eq.~\ref{EQ:ACCURACY}, we have
also used the abbreviation
\begin{equation}
    \mathbb{I}\left(\hat{x}^i_t = v_w, x^i_t = v_w\right)
    = \mathbb{I}\left(\hat{x}^i_t = v_w\right) \mathbb{I}\left(x^i_t = v_w \right).
\end{equation}

In order to account for token frequency imbalance in the vocabulary, we report
weighted-F1 score, which is defined as
\begin{equation}
    \text{wF1}(\theta; \mathcal{D}) = \frac{\sum_{w=1}^{W} N_{v_w} \times \text{F1}_{v_w}}{\sum_{w=1}^{W} N_{v_w}}.
\end{equation}
where, 
\begin{equation}
    N_{v_w} = \sum_{i=1}^{|\mathcal{D}|} \sum_{t=1}^{T_i} 
    \mathbb{I}_{\mathcal{M}_i}(t) \mathbb{I}\left(x^i_t = v_w\right),
\end{equation}
is the support or frequency of the token $v_w$ at the masked positions within all
sequences in the entire dataset and $\text{F1}_{v_w}$ is the F1 score calculated for
the token $v_w$ as the positive class. The F1 score itself is expressed as the
harmonic mean of precision and recall for each token (class) as
\begin{equation}\label{EQF1}
    \text{F1}(\theta; \mathcal{D}) = \frac{2 \times \text{Precision} \times \text{Recall}}{\text{Precision} + \text{Recall}},
\end{equation}
where
\begin{align}
    \text{Precision} &= \frac{TP}{TP + FP}, \\
    \text{Recall} &= \frac{TP}{TP + FN}.
\end{align}
In the context of \ac{MLM}, the true positive ($TP$) and true negative ($TN$)
are the number of masked tokens that are correctly predicted, and the false
positive ($FP$) and false negative ($FN$) are the number of masked tokens that
are incorrectly predicted. More specifically,
\begin{align}
    TP &= \sum_{i=1}^{|\mathcal{D}|} \sum_{t=1}^{T_i} \mathbb{I}_{\mathcal{M}_i}(t) \mathbb{I}\left(\hat{x}^i_t = v_w, x^i_t = v_w\right), \\
    FP &= \sum_{i=1}^{|\mathcal{D}|} \sum_{t=1}^{T_i} \mathbb{I}_{\mathcal{M}_i}(t) \mathbb{I}\left(\hat{x}^i_t = v_w, x^i_t \neq v_w\right), \\
    TN &= \sum_{i=1}^{|\mathcal{D}|} \sum_{t=1}^{T_i} \mathbb{I}_{\mathcal{M}_i}(t) \mathbb{I}\left(\hat{x}^i_t \neq v_w, x^i_t \neq v_w\right),   \\
    FN &= \sum_{i=1}^{|\mathcal{D}|} \sum_{t=1}^{T_i} \mathbb{I}_{\mathcal{M}_i}(t) \mathbb{I}\left(\hat{x}^i_t \neq v_w, x^i_t = v_w\right).
\end{align}

\subsection*{Datasets}
\label{SUBSEC:DATAMETHODS}
All variants of \ac{BERT} in this study are pre-trained the PubChem dataset
(version 4/18/2025), which is comprised of 119,184,806 unique chemical compounds
represented by standardized canonical isomeric \ac{SMILES}. Before pre-training,
the data was randomly split into training and validation sets with 95,347,844
data points ($\sim$ 80\% of the data) and 23,836,962 data points ($\sim$ 20\% of
the data), respectively.

We use Biogen's \ac{ADME} database\cite{Fang:2023:3263} for fine-tuning
and evaluation of our models for regression tasks. The dataset contains groups
of 885 to 3087 compounds with their corresponding experimental measurement
values for six \textit{in vitro} \ac{ADME} endpoints: \ac{HLM} stability
reported as \ac{CLINT} in \text{mL min$^{-1}$ kg$^{-1}$}, MDR1-MDCK efflux ratio
(MDR1-MDCK ER), solubility at pH 6.8 ($\mu$g mL$^{-1}$), \ac{RLM} stability
reported as \ac{CLINT} in \text{mL min$^{-1}$ kg$^{-1}$}, \ac{HPPB} in percent
unbound and \ac{RPPB} in percent unbound. The procedures for all experimental
measurements are described in Ref.~\citenum{Fang:2023:3263}.

\subsection*{Standardization}
\label{SUBSEC:STDMETHODS}
We use two different standardization pipelines for data-preprocessing: (i)
PubChem's standardization pipeline which is based on OpenEye's \texttt{OEChem}
toolkit (\url{https://docs.eyesopen.com/toolkits/python/oechemtk}) and (ii)
ChEMBL's standardization pipeline which is based on \texttt{RDKit}
(\url{https://www.rdkit.org}). The details of both standardization workflows are
summarized in the \si\ and further technical details can be found in
Refs.~\citenum{Hahnke:2018:36} and \citenum{Bento:2020:51}.

\subsection*{Tokenization}
\label{SUBSEC:TOKENIZATIONMETHODS}
The tokenization step happens before the pre-processed and standardized
\ac{SMILES} strings are fed into the model. Tokenization refers to the process
of breaking down the input text into smaller units called tokens, which can be
words, subwords, or characters. Some of the most popular tokenization methods in
\ac{NLP} are WordPiece,\cite{Wu:2016:ARXIV.1609.08144v2}
\ac{BPE},\cite{Sennrich:2016:1715} and Unigram
(SentencePiece).\cite{Kudo:2018:66} In chemistry, some groups have considered
using regular expressions to tokenize the \ac{SMILES} strings into meaningful
subunits\cite{Schwaller:2019:1572} or use them within \ac{BPE} tokenization
algorithm.\cite{Leon:2024:25016}

In this study, we use Hugging Face's \texttt{tokenizers} and
\texttt{transformers} libraries to implement and train the WordPiece and
\ac{BPE} tokenizers on the PubChem standardized canonical isomeric \ac{SMILES}
from scratch. We then compare the impact of the two tokenization methods on the
pre-training. In order to ensure a fair comparison between the two tokenizers,
we restrict the vocabulary size to 30522 tokens and limit the maximum sequence
length to 512 tokens for both tokenizers. A minimum frequency threshold of 2 is
used during training to filter out rare tokens. All experiments are conducted
for a maximum of 5 epochs. Each experiment is repeated three times with
different random seeds for model initialization and data sampling. The
performance metrics for the tokenizers are reported in Table 2 of the Extended
Data section.

\subsection*{Pre-training}
\label{SUBSEC:PTMETHODS}
In order to allow foundation \acp{LLM} learn the underlying structure, patterns
and semantics of the chemical language, we pre-train \ac{BERT} on the PubChem's
standardized canonical isomeric \ac{SMILES} using the \ac{MLM} objective. In
\ac{MLM}, a certain percentage of the input tokens (in this work, 15\%) are
randomly masked and the model is trained to predict the original tokens based on
the context provided by the unmasked tokens. Following RoBERTa's objective
design,\cite{Liu:2019:ARXIV.1907.11692v1} we use dynamic masking where the
masking pattern is changed for every batch on-the-fly during training. This
means that the model is exposed to different masked tokens in each epoch, which
can help improve the model's ability to learn from the entire input data and
prevent overfitting. We do not include the \ac{NSP} objective in our
pre-training because it is not pertinent to the downstream tasks considered in
this work. Futhermore, several works have questioned its necessity for natural
language
understanding\cite{Liu:2019:ARXIV.1907.11692v1,Yang:2019:ARXIV.1906.08237v2}.
The pre-training is performed using the \texttt{AdamW} optimizer with
$\beta_1=0.9$, $\beta_2=0.999$, and $\epsilon=10^{-8}$, a learning rate of
$10^{-4}$, a weight decay of 0.01 and an effective global batch size of 1024
(see the \si\ for more details).

\subsection*{Finetuning with Hyperparameter Search and Cross-validation}
\label{SUBSEC:FTCVSEARCHMETHODS}
All variants of pre-trained \ac{BERT} models in this study are fine-tuned on the
Biogen \ac{ADME} dataset for regression tasks, which involves predicting
the experimental values for six \textit{in vitro} \ac{ADME} endpoints: \ac{HLM},
\ac{RLM}, \ac{HPPB}, \ac{RPPB}, \ac{MDR1} and solubility.\cite{Fang:2023:3263}
All \ac{SMILES} are standardized using PubChem's standardization protocol and
tokenized using the WordPiece tokenizer. Fine-tuning is performed using 3-fold
cross-validation where the dataset is split into three folds, with two folds
used for training and one fold used for validation in each of the three
iterations. A Bayesian hyperparameter search over 50 model architectures is
performed on each fold using various foundation models, pre-trained on different
dataset sizes. Each model was trained for 100 epochs with early stopping based
on the validation loss metric with 5 epochs of patience. The best model from
each search is selected based on the validation $R^2$ metric, following
Ref.~\citenum{Fang:2023:3263}. The top performing cross-validated model is then
refitted on the entire (training and validation) data for a maximum of 20 epochs
and evaluated on a held-out test set which was not used during the
cross-validated hyperparameter search. We compare the performance of our
chemical \acp{LLM} with some of the best performing models from
Ref.~\citenum{Fang:2023:3263} which include a variety of \ac{ML} models such as
\ac{LASSO}, \ac{RF}, \ac{SVM}, \ac{XGB}, and \ac{LGBM}. All steps of the
cross-validated hyperparameter search and fine-tuning are performed for these
models on the same training data folds as those used for our \acp{LLM}.
Nonetheless, the hyperparameter search for these models is performed using the
grid search method on a pre-defined set of hyperparameters taken from
Ref.~\citenum{Fang:2023:3263}.

\section*{Data Availability}
\label{SEC:DATAAVAILABILITY}
The machine-learning ready version of the PubChem (version 04/18/2025) dataset
used for pre-training the BERT models in this study is available at
\url{https://huggingface.co/datasets/molssiai-hub/pubchem-04-18-2025}. The
original SDF files for PubChem Compounds are accessible from
\url{https://ftp.ncbi.nlm.nih.gov/pubchem/Compound}. The Biogen ADME dataset,
used for fine-tuning and evaluation of the BERT models, is provided in a public
repository at \url{https://github.com/molecularinformatics/Computational-ADME}
or \url{https://polarishub.io/datasets/biogen/adme-fang-v1}.

\section*{Code Availability}
\label{SEC:CODEAVAILABILITY}
All scripts necessary for reproducing the results of this work including
figures, tables, data split indices, data preprocessing, standardization,
tokenization, pre-training and fine-tuning the BERT models are provided in
\url{https://github.com/molssi-ai/bertology}. The ChEMBL standardization module
can be accessed from \url{https://github.com/chembl/ChEMBL_Structure_Pipeline}.
All training and fine-tuning artifacts, including the pre-trained BERT models
and the fine-tuned models for the six ADME endpoints, trainer, optimizer,
random-number generators as well as all monitoring logs for W\&B, Tensorboard
and MLFlow can be downloaded from the MolSSI's Zenodo Community page
(\url{https://zenodo.org/communities/molssi}).

%
\begin{acronym}
    \acro{ADME}{absorption, distribution, metabolism and excretion}
    \acro{APE}{atomic-pair encoding}
    \acro{BERT}{Bidirectional Encoder Representations from Transformers}
    \acro{BPE}{Byte Pair Encoding}
    \acro{CI}{confidence interval}
    \acro{CLINT}[CLint]{intrinsic clearance}
    \acro{CLM}{chemical language model}
    \acro{CNN}{convolutional neural network}
    \acro{DDP}{distributed data parallel}
    \acro{DFT}{Density Functional Theory}
    \acro{FPBERT}[FP-BERT]{Fingerprints-BERT}
    \acro{GPU}{graphics processing unit}
    \acro{HLM}{human liver microsomal}
    \acro{HPC}{high performance computing}
    \acro{HPPB}[hPPB]{human plasma protein binding}
    \acro{INCHI}[InChI]{International Chemical Identifier}
    \acro{KBERT}[K-BERT]{Knowledge-based BERT}
    \acro{LASSO}{least absolute shrinkage and selection operator}
    \acro{LGBM}{light gradient boosting machine}
    \acro{LLM}{large language model}
    \acro{MAE}{mean absolute error}
    \acro{MAP}{maximum a posteriori}
    \acro{MAT}{Molecule Attention Transformer}
    \acro{MDR1}[MDR1-MDCK ER]{MDR1-MDCK efflux ratio}
    \acro{ML}{machine learning}
    \acro{MLM}{masked language modeling}
    \acro{MPP}{molecular property prediction}
    \acro{NLP}{natural language processing}
    \acro{NN}{neural network}
    \acro{NSP}{next sentence prediction}
    \acro{PPPL}{pseudo-perplexity}
    \acro{QSAR}{quantitative structure-activity relationship}
    \acro{QSPR}{quantitative structure-property relationship}
    \acro{RF}{random forest}
    \acro{RLM}{rat liver microsomal}
    \acro{RNN}{recurrent neural network}
    \acro{RPPB}[rPPB]{rat plasma protein binding}
    \acro{RMSE}{root mean square error}
    \acro{SLM}{small language model}
    \acro{SMGBERT}[SMG-BERT]{Stereo Molecular Graph BERT}
    \acro{SMILES}{Simplified Molecular Input Line Entry System}
    \acro{SSL}{self-supervised learning}
    \acro{SVM}{support vector machine}
    \acro{TDC}{Therapeutics Data Commons}
    \acro{XGB}{extreme gradient boosting}
\end{acronym}
\nolinenumbers
\begin{acknowledgement}
    The present work is funded by the National Science Foundation grant
    CHE-2136142. PS and MM acknowledge the Advanced Research Computing
    (\url{https://arc.vt.edu}) at Virginia Tech for providing computational
    resources and technical support that have contributed to the results
    reported in this manuscript. MM thanks OpenEye Scientific (C\={a}dence
    Molecular Sciences) for providing a free academic license for their
    cheminformatics and modelling toolkits. MM is grateful to NVIDIA for
    providing two NVIDIA A100 GPUs via the NVIDIA Academic Hardware Grant
    Program which made the tasks of prototype design and testing a breeze
    throughout the course of this work.
\end{acknowledgement}
%
\bibliography{ms}

\end{document}


\newpage
\tableofcontents
\newpage



%
\acresetall	
\section{Computational Details}
\label{SISEC:COMPUTATIONAL}

\subsection{High-performance Compute Nodes and Accelerator Hardware}
\label{SISUBSEC:BATCHSIZES}
All prototyping and initial testings are performed on an on-premise
high-performance desktop machine with 2 $\times$ NVIDIA A100 80GB PCIe4
\acp{GPU}, awarded by NVIDIA Academic Hardware Grant Program. The
production-ready scripts, pre-processed standardized data and all model
artifacts are transferred to one of Virginia Tech's \ac{ARC} flagship
TinkerCliffs or Falcon compute cluster nodes for pre-training and fine-tuning.
We pre-train the Tiny-\ac{BERT} models using 16 $\times$ NVIDIA A30 24GB
\acp{GPU} on 4 Falcon nodes (each with 4 \acp{GPU}), the Small-\ac{BERT} on 8
$\times$ NVIDIA L40S 48GB \acp{GPU} on 2 Falcon nodes (each with 4 \acp{GPU}),
and the Base-\ac{BERT} models on 8 $\times$ NVIDIA H200 141GB or NVIDIA A100
SMX4 80GB PCIe4 \acp{GPU} on 1 TinkerCliffs node (each with 8 \acp{GPU}) in
\ac{DDP} mode. The effective global batch sizes of 1024 and 512 are adopted for
pre-training and evaluation of all models, respectively (see
Sec.~\ref{SISUBSEC:BATCHSIZES} for more details). Subsequent fine-tunings and
hyperparameter searches are performed on the same nodes but using a single
\ac{GPU} for each experiment.

\subsection{Global and Micro-Batch Sizes}
\label{SISUBSEC:BATCHSIZES}
Since we use \ac{DDP} alongside a wide range of nodes with different number of
\acp{GPU} and \ac{GPU} memory capacities in our experiments, we ensured that the
effective global batch size ($B_\text{eff}$) remains the same across all
experiments as it will be different from the adopted micro-batch size on each
\ac{GPU}, ($B_\text{GPU}$), through the following expression 
\begin{equation}\label{EQ:BSEFF}
    B_\text{eff} = N_\text{nodes} \times N_\text{GPU} \times B_\text{GPU} \times N_\text{GA}.
\end{equation}
Here, $N_\text{nodes}$ is the number of nodes, $N_\text{GPU}$ is the number of
\acp{GPU} per node and $N_\text{GA}$ is the number of gradient accumulation
steps. For all our calculations, $N_\text{GA}$ is set to 1.

\section{The Eﬀect of Standardization on Pre-training}
\label{SISEC:STANDARDIZATION}

\subsection{Dataset Pre-processing}
\label{SISUBSEC:STDPREPROCESS}
The \ac{SMILES} string,\\
\texttt{COC1=CC=C(C=C1)C2C3=C(C=CC4=CC=CC=C43)OC5=CC6=C(C=C5)C7}\\
\texttt{=NC8=C9C=CC1=CC9=C(N8)N=C3C4=C5C=CC(=C4)OC4=C(C(C8=C(C=CC9=CC=CC=C98)OC8}\\
\texttt{=CC9=C(C=C8)C8=NC9=NC9=C\%10C=C(C=CC\%10=C(N9)N=C9C\%10=C(C=C(C=C\%10)OC\%10}\\
\texttt{=C2C2=CC=CC=C2C=C\%10)C(=N9)NC2=NC(=N8)C8=C2C=C(C=C8)OC2=C(C(C8=C(C=CC9=CC}\\
\texttt{=CC=C98)OC8=CC9=C(C=C8)C(=NC5=N3)N=C9NC6=N7)C3=CC=C(C=C3)OC)C3=CC=CC=C3C}\\
\texttt{=C2)OC2=C(C(C3=C(O1)C=CC1=CC=CC=C13)C1=CC=C(C=C1)OC)C1=CC=CC=C1C=C2)C1=CC}\\
\texttt{=C(C=C1)OC)C1=CC=CC=C1C=C4}

\noindent
from the PubChem dataset (ver. 04/18/2025) cannot be processed by
\texttt{RDKit}'s \texttt{MolFromSmiles()} function raising the error:

\textit{Maximum BFS search size exceeded. This is likely due to a highly symmetric fused ring system.}

\noindent
As such, this \ac{SMILES} is excluded from the dataset before proceeding to the
standardization and tokenization steps.

\subsection{Molecular Sanitization in RDKit}
\label{SISUBSEC:SANITIZATION}
Since the ChEMBL standardization pipeline is based on \texttt{RDKit}, it is
imperative to understand the concept of molecular sanitization. The content of
this section closely follows the \texttt{RDKit} documentation on this
topic.\cite{RDKit:2025.03.03} 

All molecule parsing functions in \texttt{RDKit} perform a sanitization
operation by default to ensure that the generated molecules are accompanied by
useful computed properties, such as hybridization, bond types \etc\ and can be
represented by Lewis structures. The sanitization process in \texttt{RDKit}
involves the following operations, performed in the order listed below:
\begin{enumerate}[label=\arabic*-]
    \item \texttt{clearComputedProps}: Clears all existing computed properties
    on the molecule, its atoms and bonds. This operation is always performed in
    sanitization.
    \item \texttt{cleanUp}: standardizes the valence states in the following
    substructures:
    \begin{itemize}
        \item Neutral 5-valent nitrogen atoms connected to neighboring oxygens
        with double bonds or nitrogens with triple bonds are converted to their
        zwitterionic form. For example, \texttt{N(=O)(=O)} and \texttt{C-N=N\#N}
        are converted to \texttt{{[N+](=O)[O-]}} and \texttt{C-N=[N+]=[N-]},
        respectively,
        \item neutral 5-valent phosphorous atom connected to a neighboring
        oxygen with a double bond and another carbon or phosphorous atom is
        converted to its zwitterionic form. For example,
        \texttt{C=P(=O)O} is converted to \texttt{C=[P+]([O-])O},
        \item neutral 3-, 5-, or 7-valent chlorine, bromine, or iodine atoms
        connected to neighboring oxygens are converted to their zwitterionic
        form. For example, \texttt{O=Cl(=O)=O} is converted to
        \texttt{[O-][Cl+2]([O-])O}.
    \end{itemize}
    The aforementioned operations will not raise any exceptions.
    \item \texttt{cleanUpOrganometallics}: converts single bonds between a metal
    and hypervalent atom in organometallics to a dative bond,
    \item \texttt{updatePropertyCache}: computes the explicit and implicit
    valencies on all atoms in the structure. This operation is always performed
    by default but if skipped, non-standard valencies will not be tested,
    \item \texttt{symmetrizeSSSR}: performs the symmetrized smallest set of
    smallest rings (SSSR) algorithm,
    \item \texttt{Kekulize}: converts aromatic rings to their Kekul\'{e} form.
    This operation will raise an exception if the algorithm fails to Kekulize a
    ring or if aromatic bonds are found outside of rings.
    \item \texttt{assignRadicals}: determines the number of radical electrons,
    if any, on each atom,
    \item \texttt{setAromaticity}: identifies the aromatic rings and ring
    systems, sets the aromaticity flag on atoms and bonds, and converts the
    pertinent bond orders to aromatic,
    \item \texttt{setConjugation}: identifies the conjugated bonds,
    \item \texttt{setHybridization}: assigns the hybridization state to each atom,
    \item \texttt{cleanupChirality}: removes the chirality flag from non-$sp3$
    atoms,
    \item \texttt{adjustHs}: adds explicit hydrogens to preserve the chemistry,
    typically for heteroatoms in aromatic rings. The nitrogen atom in pyrrole
    offers an example for this operation.
    \item \texttt{updatePropertyCache}: is used to recompute the explicit and
    implicit valencies on all atoms to capture non-physical valencies
    on atoms with aromatic flags that are accepted in the previous steps.
\end{enumerate}
Finer control over the individual steps is offered by
\texttt{rdkit.Chem.rdmolops.SanitizeMol()} and
\texttt{rdkit.Chem.SanitizeFlags}.

\subsection{A Primer on Standardization Protocols}
\label{SISUBSEC:STANDARDIZATION}
Standardization protocols are workflows that process the chemical representation
of compounds, such as \ac{SMILES}, for an intended use. PubChem and ChEMBL
databases have their own standardization and curation pipelines which ensure the
uniqueness of the compounds and their usefulness for the potential downstream
applications. Standardization protocols must clearly delineate their strategy
about processing some of the key aspects of the chemical representations, such
as explicit or implicit specification of hydrogen atoms or handling of the
stereo-centers. 

Tautomerism, mesomerism and alternate ionization states can also contribute to
the number of possible valid, non-identical representations of the same
structure, which often exist in equilibrium. As such, tautomerism and the choice
of the predominant variants of a structure can heavily impact the corresponding
computed properties such as structural similarity, biological activity, as well
as topological and physicochemical features. 

Aromaticity is another aspect that can be defined in a multitude of ways, based
on various criteria such as chemical behavior, energetic properties, magnetic
effects, and structural traits. Different tools adopt different aromaticity
models and in the absence of a standard definition of aromaticity, it is
typically perceived from a Kekul\'{e} structure. However, the best
standardization approach may need to handle aromaticity and tautomerism
together, as the choice of a canonical tautomeric form may directly affect
aromaticity, depending on the aromaticity model employed.

\begin{figure}[!tbph]
    \centering
    \includegraphics[scale=0.9]{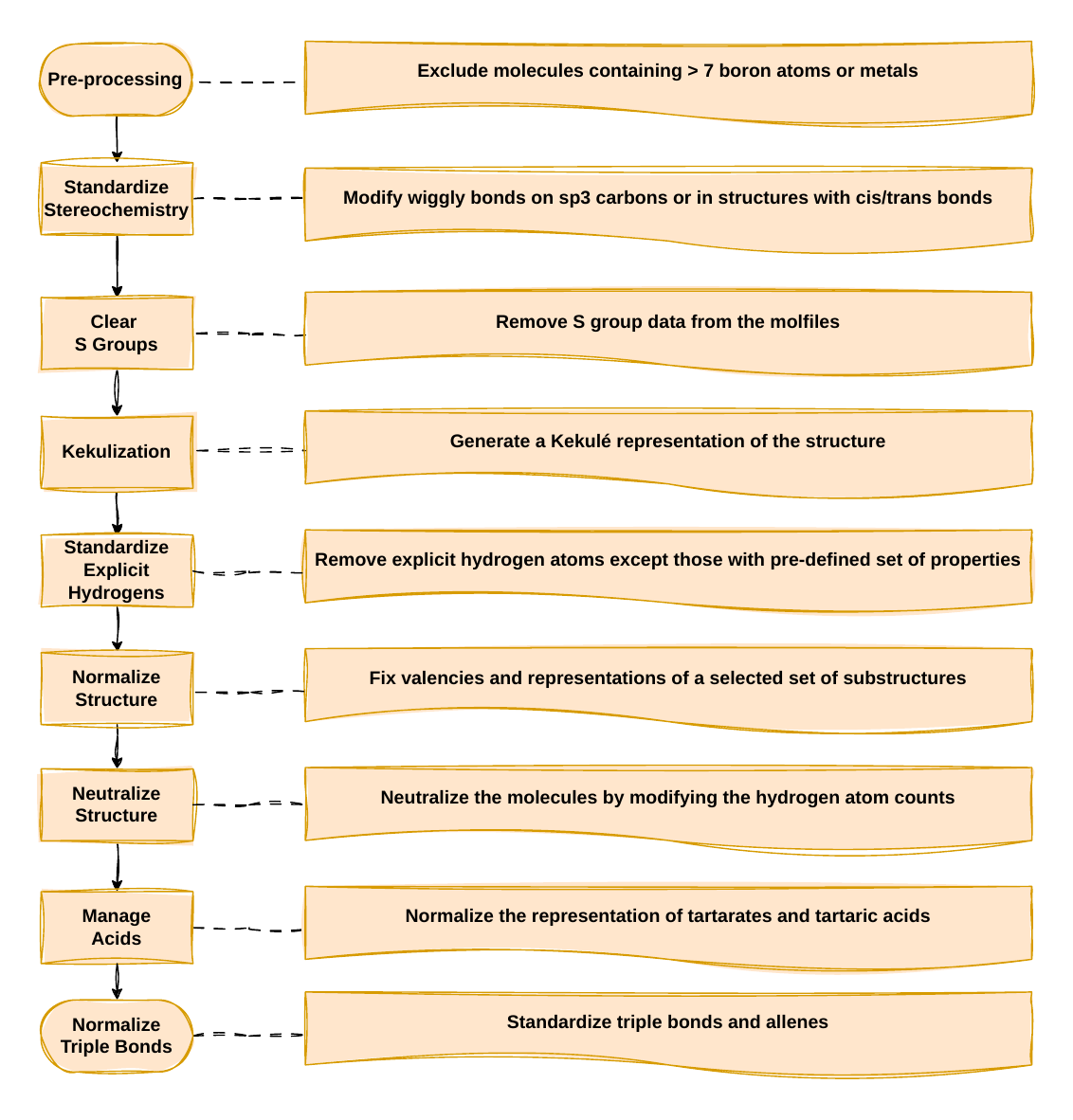}
    \caption{ChEMBL standardization protocol}
    \label{SIFIG:CHEMBLSTD}
\end{figure}

The ChEMBL curation and standardization pipeline\cite{Bento:2020:51}
(Fig.~\ref{SIFIG:CHEMBLSTD}) is based on \texttt{RDKit} and has three
components: (i) \texttt{Checker} which tests the validity of the chemical
structures and flags any serious errors, (ii) \texttt{Standardizer} which
formats compounds according to a predefined set of rules and conventions, and
(iii) \texttt{GetParent} which removes any salts and solvents from the compound
to create its parent structure, with the exception of salts in organometallic
compounds. The list of salts used in ChEMBL is an extension of USAN Council's
list of pharmacological salts. The \texttt{GetParent} module also removes all
isotopic information from the representation.\cite{Bento:2020:51}

\begin{figure}[!tbph]
\centering
    \includegraphics[scale=0.9]{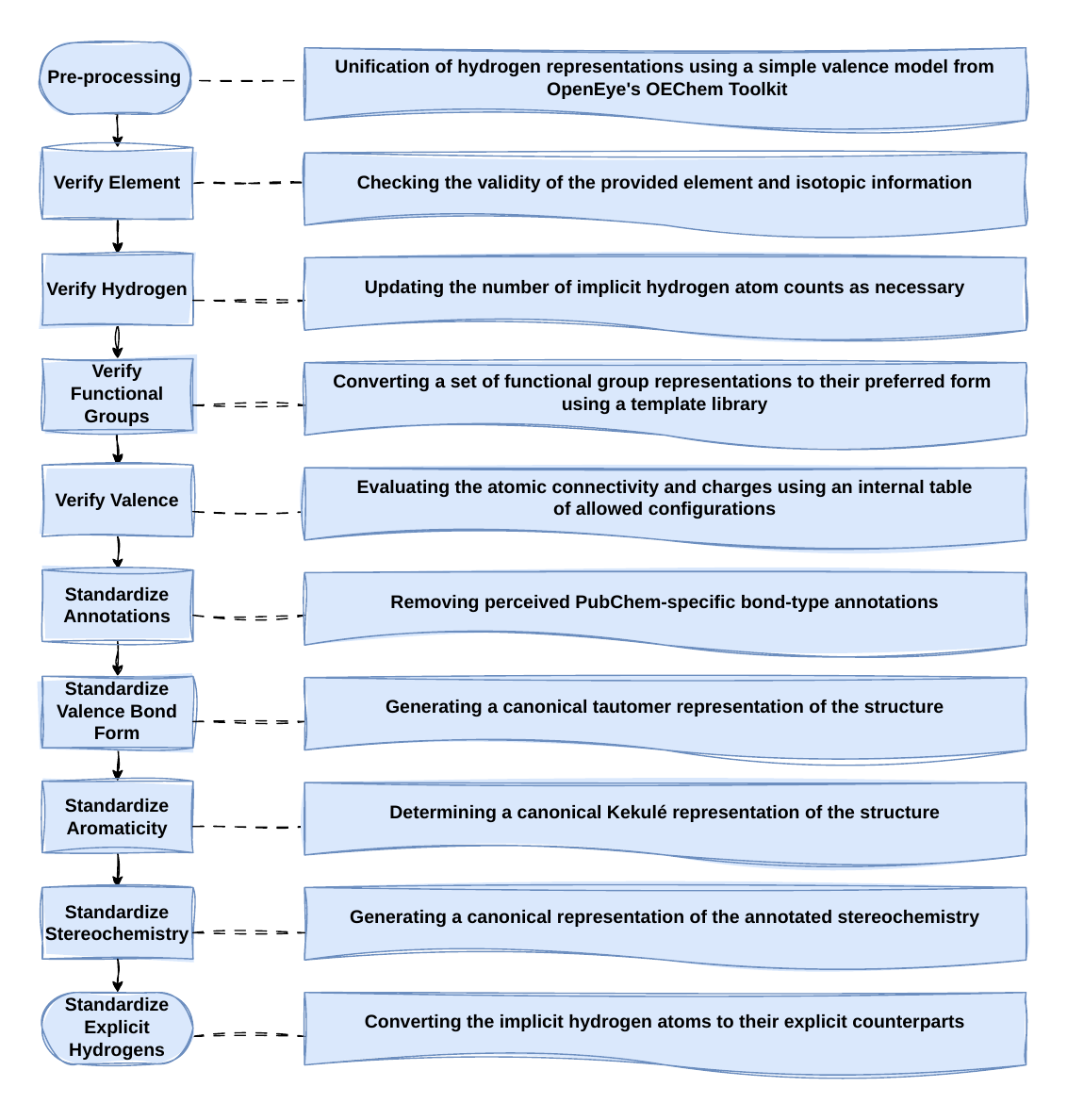}
    \caption{PubChem standardization protocol}
    \label{SIFIG:PUBCHEMSTD}
\end{figure}

ChEMBL standardization pipeline uses \acp{INCHI} and the corresponding hashed
\ac{INCHI} keys as a measure of uniqueness for the chemical structures
calculated from the \texttt{molfiles} using the \ac{INCHI} \texttt{Trust} 1.05
software. Canonical \ac{SMILES} are used as a secondary representation of the
chemical structures. Note that V2000 \texttt{molfiles} cannot distinguish
between relative and absolute stereochemistry but V3000 \texttt{molfiles} can.
Also, unlike \texttt{molfiles} and \ac{SMILES}, \acp{INCHI} are designed as an
identifier not a structural format, suitable for cheminformatics applications.
For example, the standard \ac{INCHI} representation is independent of the
tautomeric form. Thus, different tautomers of a compound will have identical
standard \acp{INCHI}. Also, \acp{INCHI} do not distinguish between cis/trans form
of organometallic compounds such as cis-/trans-platinum. 

PubChem standardization protocol (Fig.~\ref{SIFIG:PUBCHEMSTD}), on the other
hand, uses canonical isomeric \ac{SMILES} to identify the unique structures as
this method is proven to generate more unique structures than just using
standard \acp{INCHI}.\cite{Hahnke:2018:36} While the PubChem protocol provides
canonical tautomeric forms for their compounds, the ChEMBL protocol cites three
reasons not to. The developers of ChEMBL protocol rely on medicinal chemists to
assign the most appropriate tautomeric form to their compounds which is
consistent with the target application. Furthermore, tautomers can inter-convert
under experimental conditions. Moreover, changing the tautomeric form can change
the stereochemistry by modifying the chiral centers. For example, thalidomide
can inter-convert between the therapeutic $R$-enantiomer and teratogenic
$S$-enantiomer through enol tautomerism.

Regardless of the aforementioned technicalities, the standardization protocols
can have important practical implications for the performance downstream models
for \ac{MPP} tasks. For example, the ChEMBL standardization pipeline involves
the removal of a pre-defined set of salts and solvents. This means that a model
trained on ChEMBL-standardized data may not perform well for property prediction
tasks such as predicting toxicity values if the non-standardized \ac{SMILES}
forms are not available in the training data. For example,
2,4,5-trimethylaniline and its hydrochloride form, which are not differentiated
after ChEMBL standardization, have carcinogenic potency tumor dose (TD$_{50}$)
values of 33.6 and 98.5 mg per kg body weight per day in rats,
respectively.\cite{CPDB:2025:WEBSITE} The TD$_{50}$ value is the chronic
dose-rate that would induce tumors in half of the test population at the end of
their standard lifespan.

In order to further illustrate the importance of standardization in the context
of training \acp{LLM} for \ac{MPP} tasks, we provide a brief example to
demonstrate its potential impact on various steps of the \ac{NLP} workflows. Let
us consider five chemical compounds with their corresponding standardized
\ac{SMILES} representations, shown in Table \ref{SITAB:STDEXAMPLE}.

\begin{table}
\centering
\setlength{\tabcolsep}{10pt} 
\begin{tabular}{ll}
\hline\hline
    \textbf{PubChem}         & \textbf{ChEMBL}      \\
\hline
    CC(=O)[O--]              & \ce{CC(=O)O}         \\
    O=[As]([O--])([O--])O    & \ce{O=[As](O)(O)O}   \\
    \lbrack NH4+\rbrack      & \ce{N}               \\
    \lbrack CH3--\rbrack     & \ce{C}               \\
    \ce{CS(C)=O}             & C[S+](C)[O--]        \\
\hline\hline
\end{tabular}
\caption{The \ac{SMILES} representation of five different compounds 
standardized by PubChem and ChEMBL protocols}
\label{SITAB:STDEXAMPLE}
\end{table}

Looking at the first pair of rows in Table \ref{SITAB:STDEXAMPLE} and
considering the standardization workflows in Figs.~\ref{SIFIG:CHEMBLSTD} and
\ref{SIFIG:PUBCHEMSTD}, one can show that the PubChem pipeline preserves the
anionic forms of the functional groups while the ChEMBL pipeline neutralizes
them. In the fifth row, PubChem standardization preserves the neutral form of
\ce{S=O} with a double-bond while ChEMBL standardization generates a
charge-separated zwitterionic form with \ce{S+} and \ce{O-}. The decision making
behind the design of both PubChem and ChEMBL standardization pipelines can
become more clear if one considers the fact that the former uses OpenEye's
\texttt{OEChem} toolkit (commercial and closed-source) for processing the
\ac{SMILES} while the latter employs \texttt{RDKit} (free and open-source). Both
software packages have their own unique features and capabilities for handling
the \ac{SMILES} representation of chemical structures.

From the aforementioned examples, it becomes evident that the choice of
standardization pipeline can yield different vocabularies with unique token
frequencies and distributions which can potentially lead to distinct
tokenization, embedding and ultimately pre-training and fine-tuning outcomes.

\subsection{Simulating the Standardization Noise in the PubChem Dataset}
\label{SISUBSEC:STDDATAPREP}
\begin{figure}[!tbph]
    \centering
    \includegraphics{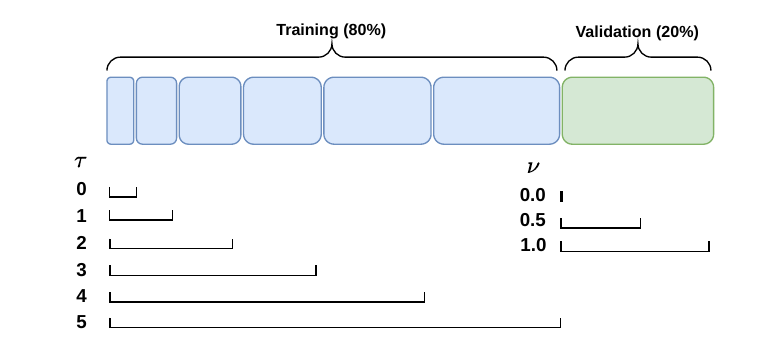}
    \caption{Corrupting the PubChem dataset with ChEMBL standardized SMILES}
    \label{SIFIG:STDDATAPREP}
\end{figure}

In order to study the impact of standardization noise on the pre-training
performance of \ac{BERT}, we create six data bins within the PubChem dataset,
each with different sizes (Fig.~\ref{SIFIG:STDDATAPREP}), and gradually replace
the PubChem canonical isomeric \ac{SMILES} with their ChEMBL standardized
counterparts.

First, the PubChem dataset is shuffled and randomly split into training and
validation splits with sizes $N_\text{train} = 95,347,844$ and $N_\text{valid} =
23,836,962$, respectively. The data corruption within each split is subsequently
controlled by two parameters, $\tau$ and $\nu$, according to the following
expressions
\begin{subequations}\label{SIEQ:STDDATA}
\begin{align}
    N^{\text{C}}_\text{train} & =  a \times 2^{\tau} + b, \\
    N^{\text{P}}_\text{train} & =  N_\text{train} - N^{\text{C}}_\text{train}, \\
    N^{\text{C}}_\text{valid} & =  \nu \times N_\text{valid}, \\
    N^{\text{P}}_\text{valid} & =  N_\text{valid} - N^{\text{C}}_\text{valid}.
\end{align}    
\end{subequations}
Here, $\tau=$ 0, 1, 2, ..., 5 and $\nu$ = 0.0, 0.5 and 1.0 correspond the
portions of the training and validation data splits, in which the PubChem
standardized canonical isomeric \ac{SMILES} are replaced with those standardized
by the ChEMBL standardization pipeline. Also, $a =$ 2,979,620 is the size of the
first bin and $b$ is the correction factor which ensures that the addition of
the sixth bin ($\tau=5$) will cover $80\%$ of the PubChem dataset. As such, $b =
4$ when $\tau = 5$ and zero otherwise. The superscripts C and P in
$N^{\text{C}}$ and $N^{\text{P}}$ correspond to the ChEMBL and PubChem
standardized data, respectively. We run each pre-training experiment three times
with different model initialization and data sampling seeds and present the
results in Table \ref{SITAB:PTSTANDARDIZATION}.

Some of the entries in Table \ref{SITAB:PTSTANDARDIZATION} are shown in red to
denote runs which resulted in model divergences or stopped by the early-stopping
conditions. For a convenient interpretation of the data, we illustrate the
statistical summaries in Table \ref{SITAB:PTSTANDARDIZATION} as heatmaps and
present them in the main text and the Extended Data section. 

\begin{ThreePartTable}
    \begin{TableNotes}[flushleft]
        \footnotesize
        \setlength{\itemsep}{-8pt}
        \item $^a$ $\tau$ and $\nu$ control the amount of standardization noise
        in the training and validation splits according to Eq.~\ref{SIEQ:STDDATA}a--d.
        \item $^b$ The model diverges during training and/or stopped by the early stopping
        criterion.
        \item $^c$ The model is excluded from calculating the mean ($\bar{x}$) and
        standard deviation ($s_N$).
        \item $^d$ $\uparrow$: higher is better; $\downarrow$: lower is better.
    \end{TableNotes}
    \centering
    \setlength{\tabcolsep}{1.pt} 
        {\renewcommand{\arraystretch}{0.64}
        \begin{xltabular}{\textwidth}{cccccccccc}
            \caption{Variations of the pre-training performance of BERT versus the
                SMILES standardization noise} \addtocounter{table}{-1} \\
                \hline\hline
                \textbf{BERT} & \textbf{(Model,Data)} & \mr{2}{*}{\textbf{Epochs}} & \mr{2}{*}{$\tau$$^a$} & \mr{2}{*}{$\nu$$^a$} & \mr{2}{*}{\textbf{T-Loss} $\downarrow$$^d$} & \mr{2}{*}{\textbf{V-Loss} $\downarrow$$^d$} & \mr{2}{*}{\textbf{V-Acc} $\uparrow$$^d$} & \mr{2}{*}{\textbf{V-wF1} $\uparrow$$^d$} & \mr{2}{*}{\textbf{V-PPPL} $\downarrow$$^d$} \\
                \textbf{Variant} & \textbf{Seeds} & & & & & & & & \\
                \hline
                \endfirsthead
                
                \caption{Variations of the pre-training performance of BERT
                versus the SMILES standardization noise (continued)}  \\ 
                \hline\hline
                \textbf{BERT} & \textbf{(Model,Data)} & \mr{2}{*}{\textbf{Epochs}} & \mr{2}{*}{$\tau$$^a$} & \mr{2}{*}{$\nu$$^a$} & \mr{2}{*}{\textbf{T-Loss} $\downarrow$$^d$} & \mr{2}{*}{\textbf{V-Loss} $\downarrow$$^d$} & \mr{2}{*}{\textbf{V-Acc} $\uparrow$$^d$} & \mr{2}{*}{\textbf{V-wF1} $\uparrow$$^d$} & \mr{2}{*}{\textbf{V-PPPL} $\downarrow$$^d$} \\
                \textbf{Variant} & \textbf{Seeds} & & & & & & & & \\
                \hline
                \endhead
            \textbf{Tiny} & (1234,1234) & 10 & 0 & 0.0 & 0.5721 & 0.5851 & 0.9090 & 0.9027 & 1.7952 \\
            \textbf{Tiny} & (1234,2345) & 10 & 0 & 0.0 & 0.5644 & 0.5718 & 0.9065 & 0.9023 & 1.7714 \\
            \textbf{Tiny} & (2345,1234) & 10 & 0 & 0.0 & 0.5527 & 0.5585 & 0.9337 & 0.9329 & 1.7480 \\
            \hline
            \mc{5}{c}{$\bar{x}$}    & 0.5631 & 0.5718 & 0.9164 & 0.9126 & 1.7715  \\
            \mc{5}{c}{$s_N$} & 0.0098 & 0.0133 & 0.0150 & 0.0175 & 0.0236  \\
            \hline
            \textbf{Tiny} & (1234,1234) & 10 & 0 & 0.5 & 0.5628 & 0.7640 & 0.9142 & 0.9091 & 2.1468 \\
            \textbf{Tiny} & (1234,2345) & 10 & 0 & 0.5 & 0.5673 & 0.7647 & 0.9071 & 0.9025 & 2.1483 \\
            \textbf{Tiny} & (2345,1234) & 10 & 0 & 0.5 & 0.5690 & 0.7698 & 0.9337 & 0.9324 & 2.1593 \\
            \hline
            \mc{5}{c}{$\bar{x}$}    & 0.5664 & 0.7662 & 0.9183 & 0.9147 & 2.1515  \\
            \mc{5}{c}{$s_N$} & 0.0032 & 0.0032 & 0.0138 & 0.0157 & 0.0068  \\
            \hline            
            \textbf{Tiny} & (1234,1234) & 10 & 0 & 1.0 & 0.5481 & 0.9467 & 0.8549 & 0.8419 & 2.5772 \\
            \textbf{Tiny} & (1234,2345) & 10 & 0 & 1.0 & 0.5587 & 0.9651 & 0.8555 & 0.8415 & 2.6251 \\
            \textbf{Tiny} & (2345,1234) & 10 & 0 & 1.0 & 0.5583 & 1.002  & 0.8644 & 0.8517 & 2.7224 \\
            \hline
            \mc{5}{c}{$\bar{x}$}    & 0.5550 & 0.9711 & 0.8582 & 0.8450 & 2.6416 \\
            \mc{5}{c}{$s_N$} & 0.0060 & 0.0279 & 0.0053 & 0.0058 & 0.0740 \\
            \hline            
            \textbf{Tiny} & (1234,1234) & 10 & 3 & 0.0 & 0.6109 & 0.5901 & 0.9097 & 0.9024 & 1.8041 \\
            \textbf{Tiny} & (1234,2345) & 10 & 3 & 0.0 & 0.6038 & 0.5839 & 0.9052 & 0.8990 & 1.7930 \\
            \textbf{Tiny} & (2345,1234) & 10 & 3 & 0.0 & 0.5934 & 0.5736 & 0.9337 & 0.9301 & 1.7747 \\
            \hline
            \mc{5}{c}{$\bar{x}$}    & 0.6027 & 0.5825 & 0.9162 & 0.9105 & 1.7906 \\
            \mc{5}{c}{$s_N$} & 0.0088 & 0.0083 & 0.0153 & 0.0170 & 0.0149 \\
            \hline            
            \textbf{Tiny} & (1234,1234) & 10 & 3 & 0.5 & 0.5974 & 0.6621 & 0.9071 & 0.9016 & 1.9388 \\
            \textbf{Tiny} & (1234,2345) & 10 & 3 & 0.5 & 0.6180 & 0.6796 & 0.9071 & 0.8977 & 1.9731 \\
            \textbf{Tiny} & (2345,1234) & 10 & 3 & 0.5 & 0.6186 & 0.6830 & 0.9362 & 0.9321 & 1.9799 \\
            \hline
            \mc{5}{c}{$\bar{x}$}    & 0.6114 & 0.6749 & 0.9168 & 0.9105 & 1.9639 \\
            \mc{5}{c}{$s_N$} & 0.0121 & 0.0112 & 0.0168 & 0.0188 & 0.0220 \\
            \hline            
            \textbf{Tiny} & (1234,1234) & 10 & 3 & 1.0 & 0.6007 & 0.7424 & 0.8908 & 0.8803 & 2.1009 \\
            \textbf{Tiny} & (1234,2345) & 10 & 3 & 1.0 & 0.6072 & 0.7566 & 0.8735 & 0.8637 & 2.1310 \\
            \textbf{Tiny} & (2345,1234) & 10 & 3 & 1.0 & 0.5908 & 0.7547 & 0.9004 & 0.8903 & 2.1270 \\
            \hline
            \mc{5}{c}{$\bar{x}$}    & 0.5996 & 0.7512 & 0.8882 & 0.8781 & 2.1196  \\
            \mc{5}{c}{$s_N$} & 0.0083 & 0.0077 & 0.0136 & 0.0134 & 0.0163  \\
            \hline            
            \alert{\textbf{Tiny}$^b$} & \alert{(1234,1234)} & \alert{6} & \alert{5} & \alert{0.0} & \alert{0.6166} & \alert{1.256} & \alert{0.7537} & \alert{0.7588} & \alert{3.5113} \\
            \alert{\textbf{Tiny}$^b$} & \alert{(1234,2345)} & \alert{4} & \alert{5} & \alert{0.0} & \alert{0.6773} & \alert{1.310} & \alert{0.7267} & \alert{0.7239} & \alert{3.7063} \\
            \alert{\textbf{Tiny}$^b$} & \alert{(2345,1234)} & \alert{5} & \alert{5} & \alert{0.0} & \alert{0.6450} & \alert{1.294} & \alert{0.7558} & \alert{0.7562} & \alert{3.6480} \\
            \hline
            \mc{5}{c}{\alert{$\bar{x}$}}    & \alert{0.6463} & \alert{1.2867} & \alert{0.7454} & \alert{0.7463} & \alert{3.6218} \\
            \mc{5}{c}{\alert{$s_N$}} & \alert{0.0304} & \alert{0.0278} & \alert{0.0163} & \alert{0.0195} & \alert{0.1001} \\
            \hline            
            \textbf{Tiny} & (1234,1234) & 10 & 5 & 0.5 & 0.6080 & 0.9488 & 0.7290 & 0.7335 & 2.5826 \\
            \textbf{Tiny} & (1234,2345) & 10 & 5 & 0.5 & 0.6004 & 0.9394 & 0.7572 & 0.7585 & 2.5584 \\
            \textbf{Tiny} & (2345,1234) & 10 & 5 & 0.5 & 0.6101 & 0.9657 & 0.7599 & 0.7678 & 2.6267 \\
            \hline
            \mc{5}{c}{$\bar{x}$}    & 0.6062 & 0.9513 & 0.7487 & 0.7533 & 2.5892 \\
            \mc{5}{c}{$s_N$} & 0.0051 & 0.0133 & 0.0171 & 0.0177 & 0.0346 \\
            \hline            
            \textbf{Tiny} & (1234,1234) & 10 & 5 & 1.0 & 0.5793 & 0.5853 & 0.9055 & 0.8996 & 1.7956 \\
            \textbf{Tiny} & (1234,2345) & 10 & 5 & 1.0 & 0.6076 & 0.6113 & 0.8981 & 0.8893 & 1.8427 \\
            \textbf{Tiny} & (2345,1234) & 10 & 5 & 1.0 & 0.5974 & 0.5985 & 0.9194 & 0.9114 & 1.8193 \\
            \hline
            \mc{5}{c}{$\bar{x}$}    & 0.5948 & 0.5983 & 0.9077 & 0.9001 & 1.8192 \\
            \mc{5}{c}{$s_N$} & 0.0143 & 0.0130 & 0.0108 & 0.0111 & 0.0236 \\
            \hline            
            \textbf{Small} & (1234,1234) & 10 & 0 & 0.0 & 0.2192 & 0.2596 & 0.9386 & 0.9384 & 1.2964 \\
            \textbf{Small} & (1234,2345) & 10 & 0 & 0.0 & 0.2198 & 0.2601 & 0.9399 & 0.9410 & 1.2971 \\
            \textbf{Small} & (2345,1234) & 10 & 0 & 0.0 & 0.2201 & 0.2607 & 0.9301 & 0.9276 & 1.2978 \\
            \hline
            \mc{5}{c}{$\bar{x}$}    & 0.2197 & 0.2601 & 0.9362 & 0.9357 & 1.2971 \\
            \mc{5}{c}{$s_N$} & 0.0004 & 0.0005 & 0.0054 & 0.0071 & 0.0007 \\
            \hline            
            \textbf{Small} & (1234,1234) & 10 & 0 & 0.5 & 0.2207 & 0.3573 & 0.9361 & 0.9373 & 1.4295 \\
            \textbf{Small} & (1234,2345) & 10 & 0 & 0.5 & 0.2196 & 0.3560 & 0.9361 & 0.9361 & 1.4277 \\
            \textbf{Small} & (2345,1234) & 10 & 0 & 0.5 & 0.2189 & 0.3544 & 0.9373 & 0.9354 & 1.4253 \\
            \hline
            \mc{5}{c}{$\bar{x}$}    & 0.2197 & 0.3559 & 0.9365 & 0.9363 & 1.4275 \\
            \mc{5}{c}{$s_N$} & 0.0009 & 0.0015 & 0.0007 & 0.0009 & 0.0021 \\
            \hline            
            \textbf{Small} & (1234,1234) & 10 & 0 & 1.0 & 0.2188 & 0.4468 & 0.9302 & 0.9347 & 1.5633 \\
            \textbf{Small} & (1234,2345) & 10 & 0 & 1.0 & 0.2206 & 0.4476 & 0.9315 & 0.9361 & 1.5646 \\
            \textbf{Small} & (2345,1234) & 10 & 0 & 1.0 & 0.2200 & 0.4468 & 0.9376 & 0.9406 & 1.5633 \\
            \hline
            \mc{5}{c}{$\bar{x}$}    & 0.2198 & 0.4471 & 0.9331 & 0.9371 & 1.5638 \\
            \mc{5}{c}{$s_N$} & 0.0009 & 0.0005 & 0.0039 & 0.0031 & 0.0008 \\
            \hline            
            \textbf{Small} & (1234,1234) & 10 & 3 & 0.0 & 0.2272 & 0.2623 & 0.9418 & 0.9416 & 1.2999 \\
            \textbf{Small} & (1234,2345) & 10 & 3 & 0.0 & 0.2268 & 0.2620 & 0.9374 & 0.9360 & 1.2996 \\
            \textbf{Small} & (2345,1234) & 10 & 3 & 0.0 & 0.2268 & 0.2626 & 0.9346 & 0.9329 & 1.3003 \\
            \hline
            \mc{5}{c}{$\bar{x}$}    & 0.2269 & 0.2623 & 0.9379 & 0.9368 & 1.2999 \\
            \mc{5}{c}{$s_N$} & 0.0002 & 0.0003 & 0.0036 & 0.0044 & 0.0003 \\
            \hline            
            \textbf{Small} & (1234,1234) & 10 & 3 & 0.5 & 0.2270 & 0.3004 & 0.9399 & 0.9409 & 1.3503 \\
            \textbf{Small} & (1234,2345) & 10 & 3 & 0.5 & 0.2256 & 0.2990 & 0.9393 & 0.9399 & 1.3485 \\
            \textbf{Small} & (2345,1234) & 10 & 3 & 0.5 & 0.2265 & 0.2988 & 0.9340 & 0.9315 & 1.3482 \\
            \hline
            \mc{5}{c}{$\bar{x}$}    & 0.2264 & 0.2994 & 0.9377 & 0.9374 & 1.3490 \\
            \mc{5}{c}{$s_N$} & 0.0007 & 0.0008 & 0.0033 & 0.0052 & 0.0011 \\
            \hline            
            \textbf{Small} & (1234,1234) & 10 & 3 & 1.0 & 0.2263 & 0.3311 & 0.9335 & 0.9332 & 1.3925 \\
            \textbf{Small} & (1234,2345) & 10 & 3 & 1.0 & 0.2277 & 0.3335 & 0.9256 & 0.9276 & 1.3958 \\
            \textbf{Small} & (2345,1234) & 10 & 3 & 1.0 & 0.2269 & 0.3336 & 0.9383 & 0.9392 & 1.3960 \\
            \hline
            \mc{5}{c}{$\bar{x}$}    & 0.2269 & 0.3327 & 0.9325 & 0.9333 & 1.3948 \\
            \mc{5}{c}{$s_N$} & 0.0007 & 0.0014 & 0.0064 & 0.0058 & 0.0020 \\
            \hline            
            \alert{\textbf{Small}$^b$} & \alert{(1234,1234)} & \alert{4} & \alert{5} & \alert{0.0} & \alert{0.2540} & \alert{1.209} & \alert{0.7893} & \alert{0.8040} & \alert{3.3516} \\
            \alert{\textbf{Small}$^b$} & \alert{(1234,2345)} & \alert{4} & \alert{5} & \alert{0.0} & \alert{0.2533} & \alert{1.199} & \alert{0.7900} & \alert{0.7924} & \alert{3.3178} \\
            \alert{\textbf{Small}$^b$} & \alert{(2345,1234)} & \alert{5} & \alert{5} & \alert{0.0} & \alert{0.2456} & \alert{1.240} & \alert{0.7667} & \alert{0.7849} & \alert{3.4556} \\
            \hline
            \mc{5}{c}{\alert{$\bar{x}$}}    & \alert{0.2510} & \alert{1.2163} & \alert{0.7820} & \alert{0.7937} & \alert{3.3750} \\
            \mc{5}{c}{\alert{$s_N$}} & \alert{0.0047} & \alert{0.0212} & \alert{0.0132} & \alert{0.0097} & \alert{0.0718} \\
            \hline            
            \alert{\textbf{Small}$^b$} & \alert{(1234,1234)} & \alert{6} & \alert{5} & \alert{0.5} & \alert{0.2380} & \alert{0.7735} & \alert{0.7825} & \alert{0.7974} & \alert{2.1674} \\
            \alert{\textbf{Small}$^b$} & \alert{(1234,2345)} & \alert{7} & \alert{5} & \alert{0.5} & \alert{0.2334} & \alert{0.7742} & \alert{0.7647} & \alert{0.7829} & \alert{2.1689} \\
            \alert{\textbf{Small}$^b$} & \alert{(2345,1234)} & \alert{7} & \alert{5} & \alert{0.5} & \alert{0.2339} & \alert{0.7710} & \alert{0.7811} & \alert{0.7945} & \alert{2.1620} \\
            \hline
            \mc{5}{c}{\alert{$\bar{x}$}}    & \alert{0.2351} & \alert{0.7729} & \alert{0.7761} & \alert{0.7916} & \alert{2.1661} \\
            \mc{5}{c}{\alert{$s_N$}} & \alert{0.0026} & \alert{0.0017} & \alert{0.0099} & \alert{0.0077} & \alert{0.0036} \\
            \hline            
            \textbf{Small} & (1234,1234) & 10 & 5 & 1.0 & 0.2229 & 0.2650 & 0.9302 & 0.9314 & 1.3034 \\
            \textbf{Small} & (1234,2345) & 10 & 5 & 1.0 & 0.2217 & 0.2645 & 0.9289 & 0.9301 & 1.3027 \\
            \textbf{Small} & (2345,1234) & 10 & 5 & 1.0 & 0.2230 & 0.2654 & 0.9369 & 0.9369 & 1.3040 \\
            \hline
            \mc{5}{c}{$\bar{x}$}    & 0.2225 & 0.2650 & 0.9320 & 0.9328 & 1.3034 \\
            \mc{5}{c}{$s_N$} & 0.0007 & 0.0005 & 0.0043 & 0.0036 & 0.0006 \\
            \hline            
            \textbf{Base} & (1234,1234) & 10 & 0 & 0.0 & 0.0139 & 0.2167 & 0.9437 & 0.9447 & 1.2420 \\
            \textbf{Base} & (1234,2345) & 10 & 0 & 0.0 & 0.1676 & 0.2160 & 0.9405 & 0.9406 & 1.2411 \\
            \textbf{Base} & (2345,1234) & 10 & 0 & 0.0 & 0.0138 & 0.2169 & 0.9386 & 0.9365 & 1.2422 \\
            \hline
            \mc{5}{c}{$\bar{x}$}    & 0.0651 & 0.2165 & 0.9409 & 0.9406 & 1.2417 \\
            \mc{5}{c}{$s_N$} & 0.0888 & 0.0005 & 0.0026 & 0.0041 & 0.0006 \\
            \hline            
            \alert{\textbf{Base}$^b$} & \alert{(1234,1234)} & \alert{4} & \alert{0} & \alert{0.5} & \alert{10.95} & \alert{5.335} & \alert{0.1466} & \alert{0.03750} & \alert{207.38} \\
            \textbf{Base} & (1234,2345) & 10 & 0 & 0.5 & 0.0275 & 0.2774 & 0.9374 & 0.9380 & 1.3197 \\
            \alert{\textbf{Base}$^b$} & \alert{(2345,1234)} & \alert{9} & \alert{0} & \alert{0.5} & \alert{4.814} & \alert{4.088} & \alert{0.1327} & \alert{0.03109} & \alert{59.598}  \\
            \hline
            \mc{5}{c}{\alert{$\bar{x}$}}    & \alert{5.2651} & \alert{3.2332} & \alert{0.4056} & \alert{0.3355} & \alert{89.4321} \\
            \mc{5}{c}{\alert{$s_N$}} & \alert{5.4772} & \alert{2.6346} & \alert{0.4606} & \alert{0.5218} & \alert{106.2198} \\
            \hline            
            \alert{\textbf{Base}$^{b,c}$} & \alert{(1234,1234)} & \alert{4} & \alert{0} & \alert{1.0} & \alert{23.94} & \alert{4.158} & \alert{0.1439} & \alert{0.03622} & \alert{63.969} \\
            \textbf{Base} & (1234,2345) & 10 & 0 & 1.0 & 0.02779 & 0.3313 & 0.9289 & 0.9297 & 1.3928 \\
            \textbf{Base} & (2345,1234) & 10 & 0 & 1.0 & 0.04259 & 0.3276 & 0.9417 & 0.9409 & 1.3876 \\
            \hline
            \mc{5}{c}{\alert{$\bar{x}$}}    & \alert{0.0352} & \alert{0.3294} & \alert{0.9353} & \alert{0.9353} & \alert{1.3902} \\
            \mc{5}{c}{\alert{$s_N$}} & \alert{0.0105} & \alert{0.0027} & \alert{0.0090} & \alert{0.0079} & \alert{0.0037} \\
            \hline            
            \textbf{Base} & (1234,1234) & 10 & 3 & 0.0 & 0.1720  & 0.2168 & 0.9418 & 0.9433 & 1.2420 \\
            \textbf{Base} & (1234,2345) & 10 & 3 & 0.0 & 0.02822 & 0.2177 & 0.9399 & 0.9407 & 1.2432 \\
            \alert{\textbf{Base}$^{b,c}$} & \alert{(2345,1234)} & \alert{7} & \alert{3} & \alert{0.0} & \alert{4.168} & \alert{4.125} & \alert{0.1434} & \alert{0.03597} & \alert{61.88} \\
            \hline
            \mc{5}{c}{\alert{$\bar{x}$}}    & \alert{0.1001} & \alert{0.2172} & \alert{0.9409} & \alert{0.9420} & \alert{1.2426} \\
            \mc{5}{c}{\alert{$s_N$}} & \alert{0.1017} & \alert{0.0006} & \alert{0.0013} & \alert{0.0018} & \alert{0.0008} \\
            \hline            
            \textbf{Base} & (1234,1234) & 10 & 3 & 0.5 & 0.0140 & 0.2460 & 0.9399 & 0.9410 & 1.2789 \\
            \textbf{Base} & (1234,2345) & 10 & 3 & 0.5 & 0.2256 & 0.2990 & 0.9393 & 0.9399 & 1.3485 \\
            \textbf{Base} & (2345,1234) & 10 & 3 & 0.5 & 0.02847 & 0.2457 & 0.9353 & 0.9326 & 1.2785 \\
            \hline
            \mc{5}{c}{$\bar{x}$}    & 0.0894 & 0.2636 & 0.9382 & 0.9379 & 1.3020 \\
            \mc{5}{c}{$s_N$} & 0.1182 & 0.0307 & 0.0025 & 0.0046 & 0.0403 \\
            \hline            
            \textbf{Base} & (1234,1234) & 10 & 3 & 1.0 & 0.1714 & 0.2675 & 0.9413 & 0.9411 & 1.3068 \\
            \textbf{Base} & (1234,2345) & 10 & 3 & 1.0 & 0.01406 & 0.2688 & 0.9328 & 0.9340 & 1.3084  \\
            \textbf{Base} & (2345,1234) & 10 & 3 & 1.0 & 0.1711 & 0.2682 & 0.9423 & 0.9432 & 1.3076 \\
            \hline
            \mc{5}{c}{$\bar{x}$}    & 0.1188 & 0.2682 & 0.9388 & 0.9394 & 1.3076 \\
            \mc{5}{c}{$s_N$} & 0.0907 & 0.0006 & 0.0052 & 0.0048 & 0.0008 \\
            \hline            
            \alert{\textbf{Base}$^b$} & \alert{(1234,1234)} & \alert{4} & \alert{5} & \alert{0.0} & \alert{0.2021} & \alert{1.244}  & \alert{0.7853} & \alert{0.7990} & \alert{3.4711} \\
            \alert{\textbf{Base}$^b$} & \alert{(1234,2345)} & \alert{4} & \alert{5} & \alert{0.0} & \alert{0.1360} & \alert{1.279}  & \alert{0.7629} & \alert{0.7722} & \alert{3.5929} \\
            \alert{\textbf{Base}$^b$} & \alert{(2345,1234)} & \alert{4} & \alert{5} & \alert{0.0} & \alert{0.2008} & \alert{1.230}  & \alert{0.7741} & \alert{0.7854} & \alert{3.4200} \\
            \hline
            \mc{5}{c}{\alert{$\bar{x}$}}    & \alert{0.1796} & \alert{1.2510} & \alert{0.7741} & \alert{0.7855} & \alert{3.4947} \\
            \mc{5}{c}{\alert{$s_N$}} & \alert{0.0378} & \alert{0.0253} & \alert{0.0112} & \alert{0.0134} & \alert{0.0888} \\
            \hline            
            \alert{\textbf{Base}$^b$} & \alert{(1234,1234)} & \alert{4} & \alert{5} & \alert{0.5} & \alert{0.2020} & \alert{0.7571} & \alert{0.7820} & \alert{0.7956} & \alert{2.1321} \\
            \alert{\textbf{Base}$^b$} & \alert{(1234,2345)} & \alert{5} & \alert{5} & \alert{0.5} & \alert{0.1942} & \alert{0.7442} & \alert{0.7639} & \alert{0.7817} & \alert{2.1048} \\
            \alert{\textbf{Base}$^b$} & \alert{(2345,1234)} & \alert{6} & \alert{5} & \alert{0.5} & \alert{0.1892} & \alert{0.7692} & \alert{0.7487} & \alert{0.7697} & \alert{2.1581} \\
            \hline
            \mc{5}{c}{\alert{$\bar{x}$}}    & \alert{0.1951} & \alert{0.7569} & \alert{0.7649} & \alert{0.7823} & \alert{2.1317} \\
            \mc{5}{c}{\alert{$s_N$}} & \alert{0.0065} & \alert{0.0125} & \alert{0.0167} & \alert{0.0130} & \alert{0.0267} \\
            \hline            
            \textbf{Base} & (1234,1234) & 10 & 5 & 1.0 & 0.1720 & 0.2234 & 0.9393 & 0.9391 & 1.2503 \\
            \textbf{Base} & (1234,2345) & 10 & 5 & 1.0 & 0.0745 & 0.2229 & 0.9361 & 0.9376 & 1.2497 \\
            \textbf{Base} & (2345,1234) & 10 & 5 & 1.0 & 0.0287 & 0.2234 & 0.9410 & 0.9404 & 1.2504 \\
            \hline
            \mc{5}{c}{$\bar{x}$}    & 0.0917 & 0.2232 & 0.9388 & 0.9390 & 1.2501 \\
            \mc{5}{c}{$s_N$} & 0.0732 & 0.0003 & 0.0025 & 0.0014 & 0.0004 \\
            \hline            
\hline\hline
\insertTableNotes
\end{xltabular}
}
\label{SITAB:PTSTANDARDIZATION}
\end{ThreePartTable}

\section{The Effect of Dataset and Model Sizes on Pre-training}
\label{SISEC:DATASIZEPRETRAIN}
In order to study the dataset and model size effects on pre-training performance
of \ac{BERT}, we create six dataset bins with different sizes, where the number
of training samples in each bin is defined as
\begin{equation}\label{EQ:DATASIZE}
    N_\text{train} = a \times 2^{k} + b,
\end{equation}
where $k=$ 0, 1, 2, 3, 4 and 5  denotes the bin index, $a =$ 2,979,620 is the
size of the first bin and $b$ is the correction factor which ensures that the
addition of the sixth bin ($k=5$) will cover $80\%$ of the PubChem dataset. As
such, $b = 4$ when $k = 5$ and zero otherwise.

\begin{ThreePartTable}
    \begin{TableNotes}[flushleft]
        \footnotesize
        \setlength{\itemsep}{-8pt}
        \item $^a$ The bin index, $k$, determines the number of samples in the
        training split.
        \item $^b$ The model diverges during training and/or stopped by the
        early stopping criterion.
        \item $^c$ The model is excluded from calculating the mean ($\bar{x}$) and
        standard deviation ($s_N$).
        \item $^d$ $\uparrow$: higher is better; $\downarrow$: lower is better.
    \end{TableNotes}
    \centering
    \setlength{\tabcolsep}{4pt} 
        {\renewcommand{\arraystretch}{0.64}
        \begin{xltabular}{\textwidth}{cccccccc}
            \caption{Variations of the pre-training performance of BERT versus
            dataset and model sizes} \addtocounter{table}{-1} \\
            \hline\hline
            \textbf{BERT} & \textbf{(Model,Data)} & \mr{2}{*}{$k$$^a$} & \mr{2}{*}{\textbf{T-Loss} $\downarrow$$^d$} & \mr{2}{*}{\textbf{V-Loss} $\downarrow$$^d$} & \mr{2}{*}{\textbf{V-Acc} $\uparrow$$^d$} & \mr{2}{*}{\textbf{V-wF1} $\uparrow$$^d$} & \mr{2}{*}{\textbf{V-PPPL} $\downarrow$$^d$} \\
            \textbf{Variant} & \textbf{Seed} & \\
            \hline
            \endfirsthead
            
            \caption{Variations of the pre-training performance of BERT versus
            dataset and model sizes (continued)}  \\
            \hline\hline
            \textbf{BERT} & \textbf{(Model,Data)} & \mr{2}{*}{$k$$^a$} & \mr{2}{*}{\textbf{T-Loss} $\downarrow$$^d$} & \mr{2}{*}{\textbf{V-Loss} $\downarrow$$^d$} & \mr{2}{*}{\textbf{V-Acc} $\uparrow$$^d$} & \mr{2}{*}{\textbf{V-wF1} $\uparrow$$^d$} & \mr{2}{*}{\textbf{V-PPPL} $\downarrow$$^d$} \\
            \textbf{Variant} & \textbf{Seed} & \\
            \hline
            \endhead
\textbf{Tiny} & (1234,1234) & 0 & 0.9602 & 1.3382 & 0.8618 & 0.8479 & 3.8123 \\
\textbf{Tiny} & (1234,2345) & 0 & 1.0184 & 1.3893 & 0.8606 & 0.8475 & 4.0122 \\
\textbf{Tiny} & (2345,1234) & 0 & 0.9708 & 1.3070 & 0.8650 & 0.8482 & 3.6950 \\
\hline
\mc{3}{c}{$\bar{x}$}    & 0.9831 & 1.3449 & 0.8625 & 0.8479 & 3.8398 \\
\mc{3}{c}{$s_N$}        & 0.0310 & 0.0416 & 0.0023 & 0.0004 & 0.1604 \\
\hline
\textbf{Tiny} & (1234,1234) & 1 & 0.7694 & 1.1407 & 0.8788 & 0.8681 & 3.1290 \\
\textbf{Tiny} & (1234,2345) & 1 & 0.7847 & 1.1735 & 0.8825 & 0.8694 & 3.2332 \\
\textbf{Tiny} & (2345,1234) & 1 & 0.7895 & 1.1738 & 0.8745 & 0.8607 & 3.2343 \\
\hline
\mc{3}{c}{$\bar{x}$}    & 0.7812 & 1.1627 & 0.8786 & 0.8660 & 3.1988 \\
\mc{3}{c}{$s_N$}        & 0.0105 & 0.0190 & 0.0040 & 0.0046 & 0.0605 \\
\hline
\textbf{Tiny} & (1234,1234) & 2 & 0.6798 & 0.8764 & 0.9026 & 0.8952 & 2.4023 \\
\textbf{Tiny} & (1234,2345) & 2 & 0.6691 & 0.8736 & 0.8926 & 0.8858 & 2.3956 \\
\textbf{Tiny} & (2345,1234) & 2 & 0.6715 & 0.8633 & 0.9037 & 0.8942 & 2.3710 \\
\hline
\mc{3}{c}{$\bar{x}$}    & 0.6734 & 0.8711 & 0.8996 & 0.8917 & 2.3897 \\
\mc{3}{c}{$s_N$}        & 0.0056 & 0.0069 & 0.0061 & 0.0052 & 0.0165 \\
\hline
\textbf{Tiny} & (1234,1234) & 3 & 0.5704 & 0.7628 & 0.9139 & 0.9080 & 2.1443 \\
\textbf{Tiny} & (1234,2345) & 3 & 0.5801 & 0.7654 & 0.9133 & 0.9072 & 2.1498 \\
\textbf{Tiny} & (2345,1234) & 3 & 0.5496 & 0.7156 & 0.9227 & 0.9189 & 2.0454 \\
\hline
\mc{3}{c}{$\bar{x}$}    & 0.5667 & 0.7479 & 0.9166 & 0.9114 & 2.1132 \\
\mc{3}{c}{$s_N$}        & 0.0156 & 0.0280 & 0.0052 & 0.0065 & 0.0587 \\
\hline
\textbf{Tiny} & (1234,1234) & 4 & 0.5688 & 0.6538 & 0.9165 & 0.9103 & 1.9229 \\
\textbf{Tiny} & (1234,2345) & 4 & 0.5612 & 0.6374 & 0.9190 & 0.9160 & 1.8915 \\
\textbf{Tiny} & (2345,1234) & 4 & 0.5771 & 0.6680 & 0.9246 & 0.9181 & 1.9503 \\
\hline
\mc{3}{c}{$\bar{x}$}    & 0.5690 & 0.6531 & 0.9200 & 0.9148 & 1.9216 \\
\mc{3}{c}{$s_N$}        & 0.0080 & 0.0153 & 0.0042 & 0.0040 & 0.0294 \\
\hline
\textbf{Tiny} & (1234,1234) & 5 & 0.5574 & 0.5760 & 0.9146 & 0.9100 & 1.7789 \\
\textbf{Tiny} & (1234,2345) & 5 & 0.5565 & 0.5820 & 0.9209 & 0.9155 & 1.7895 \\
\textbf{Tiny} & (2345,1234) & 5 & 0.5376 & 0.5589 & 0.9221 & 0.9168 & 1.7488 \\
\hline
\mc{3}{c}{$\bar{x}$}    & 0.5505 & 0.5723 & 0.9192 & 0.9141 & 1.7724 \\
\mc{3}{c}{$s_N$}        & 0.0111 & 0.0120 & 0.0040 & 0.0036 & 0.0211 \\
\hline
\textbf{Small} & (1234,1234) & 0 & 0.3026 & 0.7616 & 0.9209 & 0.9224 & 2.1418 \\
\textbf{Small} & (1234,2345) & 0 & 0.3007 & 0.7522 & 0.9247 & 0.9249 & 2.1217 \\
\textbf{Small} & (2345,1234) & 0 & 0.3009 & 0.7520 & 0.9210 & 0.9222 & 2.1211 \\
\hline
\mc{3}{c}{$\bar{x}$}    & 0.3014 & 0.7553 & 0.9222 & 0.9232 & 2.1282 \\
\mc{3}{c}{$s_N$}        & 0.0011 & 0.0055 & 0.0021 & 0.0015 & 0.0118 \\
\hline
\textbf{Small} & (1234,1234) & 1 & 0.2225 & 0.6476 & 0.9296 & 0.9293 & 1.9110 \\
\textbf{Small} & (1234,2345) & 1 & 0.2224 & 0.6332 & 0.9290 & 0.9282 & 1.8837 \\
\textbf{Small} & (2345,1234) & 1 & 0.2213 & 0.6311 & 0.9295 & 0.9314 & 1.8797 \\
\hline
\mc{3}{c}{$\bar{x}$}    & 0.2221 & 0.6373 & 0.9294 & 0.9297 & 1.8915 \\
\mc{3}{c}{$s_N$}        & 0.0007 & 0.0090 & 0.0003 & 0.0016 & 0.0171 \\
\hline
\textbf{Small} & (1234,1234) & 2 & 0.2083 & 0.4312 & 0.9328 & 0.9314 & 1.5391 \\
\textbf{Small} & (1234,2345) & 2 & 0.2055 & 0.4225 & 0.9352 & 0.9340 & 1.5258 \\
\textbf{Small} & (2345,1234) & 2 & 0.2069 & 0.4330 & 0.9360 & 0.9356 & 1.5419 \\
\hline
\mc{3}{c}{$\bar{x}$}    & 0.2069 & 0.4289 & 0.9347 & 0.9337 & 1.5356 \\
\mc{3}{c}{$s_N$}        & 0.0014 & 0.0056 & 0.0017 & 0.0021 & 0.0086 \\
\hline
\textbf{Small} & (1234,1234) & 3 & 0.1893 & 0.3779 & 0.9290 & 0.9306 & 1.4593 \\
\textbf{Small} & (1234,2345) & 3 & 0.1894 & 0.3766 & 0.9321 & 0.9325 & 1.4573 \\
\textbf{Small} & (2345,1234) & 3 & 0.1891 & 0.3722 & 0.9295 & 0.9299 & 1.4509 \\
\hline
\mc{3}{c}{$\bar{x}$}    & 0.1893 & 0.3755 & 0.9302 & 0.9310 & 1.4558 \\
\mc{3}{c}{$s_N$}        & 0.0002 & 0.0030 & 0.0017 & 0.0014 & 0.0044 \\
\hline
\textbf{Small} & (1234,1234) & 4 & 0.2085 & 0.3115 & 0.9402 & 0.9396 & 1.3654 \\
\textbf{Small} & (1234,2345) & 4 & 0.2078 & 0.3105 & 0.9346 & 0.9334 & 1.3641 \\
\textbf{Small} & (2345,1234) & 4 & 0.2077 & 0.3112 & 0.9341 & 0.9337 & 1.3651 \\
\hline
\mc{3}{c}{$\bar{x}$}    & 0.2080 & 0.3111 & 0.9363 & 0.9356 & 1.3649 \\
\mc{3}{c}{$s_N$}        & 0.0004 & 0.0005 & 0.0034 & 0.0035 & 0.0007 \\
\hline
\textbf{Small} & (1234,1234) & 5 & 0.2165 & 0.2611 & 0.9359 & 0.9358 & 1.2984 \\
\textbf{Small} & (1234,2345) & 5 & 0.2159 & 0.2601 & 0.9371 & 0.9369 & 1.2971 \\
\textbf{Small} & (2345,1234) & 5 & 0.2164 & 0.2609 & 0.9386 & 0.9392 & 1.2981 \\
\hline
\mc{3}{c}{$\bar{x}$}    & 0.2163 & 0.2607 & 0.9372 & 0.9373 & 1.2979 \\
\mc{3}{c}{$s_N$}        & 0.0003 & 0.0005 & 0.0014 & 0.0017 & 0.0007 \\
\hline
\alert{\textbf{Base}$^b$} & \alert{(1234,1234)} & \alert{0} & \alert{3.1122} & \alert{3.1122} & \alert{0.1504} & \alert{0.0393} & \alert{29.0734} \\
\textbf{Base} & (1234,2345) & 0 & 0.2275 & 0.6553 & 0.9284 & 0.9273 & 1.9257 \\
\alert{\textbf{Base}$^b$} & \alert{(2345,1234)} & \alert{0} & \alert{3.1113} & \alert{3.3697} & \alert{0.1600} & \alert{0.0441} & \alert{29.0711} \\
\hline
\mc{3}{c}{\alert{$\bar{x}$}}    & \alert{2.1503} & \alert{2.3791} & \alert{0.4129} & \alert{0.3369} & \alert{20.0234} \\
\mc{3}{c}{\alert{$s_N$}}        & \alert{1.6652} & \alert{1.4984} & \alert{0.4464} & \alert{0.5113} & \alert{15.6730} \\
\hline
\textbf{Base} & (1234,1234) & 1 & 0.1530 & 0.5360 & 0.9346 & 0.9331 & 1.7092 \\
\alert{\textbf{Base}$^b$} & \alert{(1234,2345)} & \alert{1} & \alert{3.1026} & \alert{3.3638} & \alert{0.1545} & \alert{0.0414} & \alert{28.8994} \\
\alert{\textbf{Base}$^b$} & \alert{(2345,1234)} & \alert{1} & \alert{3.0988} & \alert{3.3586} & \alert{0.1600} & \alert{0.0441} & \alert{28.7492} \\
\hline
\mc{3}{c}{\alert{$\bar{x}$}}    & \alert{2.1181} & \alert{2.4195} & \alert{0.4164} & \alert{0.3395} & \alert{19.7859} \\
\mc{3}{c}{\alert{$s_N$}}        & \alert{1.7019} & \alert{1.6311} & \alert{0.4488} & \alert{0.5141} & \alert{15.6551} \\
\hline
\textbf{Base} & (1234,1234) & 2 & 0.1461 & 0.3709 & 0.9371 & 0.9347 & 1.4490 \\
\textbf{Base} & (1234,2345) & 2 & 0.1476 & 0.3709 & 0.9408 & 0.9394 & 1.4490 \\
\textbf{Base} & (2345,1234) & 2 & 0.1457 & 0.3728 & 0.9386 & 0.9381 & 1.4518 \\
\hline
\mc{3}{c}{$\bar{x}$}    & 0.1465 & 0.3715 & 0.9389 & 0.9374 & 1.4499 \\
\mc{3}{c}{$s_N$}        & 0.0010 & 0.0011 & 0.0019 & 0.0024 & 0.0016 \\
\hline
\textbf{Base} & (1234,1234) & 3 & 0.1387 & 0.3295 & 0.9359 & 0.9354 & 1.3902 \\
\textbf{Base} & (1234,2345) & 3 & 0.1396 & 0.3310 & 0.9346 & 0.9347 & 1.3923 \\
\textbf{Base} & (2345,1234) & 3 & 0.1385 & 0.3299 & 0.9334 & 0.9334 & 1.3909 \\
\hline
\mc{3}{c}{$\bar{x}$}    & 0.1390 & 0.3301 & 0.9346 & 0.9345 & 1.3911 \\
\mc{3}{c}{$s_N$}        & 0.0006 & 0.0008 & 0.0012 & 0.0010 & 0.0011 \\
\hline
\textbf{Base} & (1234,1234) & 4 & 0.1580 & 0.2693 & 0.9446 & 0.9425 & 1.3091 \\
\textbf{Base} & (1234,2345) & 4 & 0.1584 & 0.2704 & 0.9384 & 0.9377 & 1.3105 \\
\alert{\textbf{Base}$^{b,c}$} & \alert{(2345,1234)} & \alert{4} & \alert{3.1273} & \alert{4.5999} & \alert{0.1390} & \alert{0.0339} & \alert{99.4766} \\
\hline
\mc{3}{c}{\alert{$\bar{x}$}}    & \alert{0.1582} & \alert{0.2699} & \alert{0.9415} & \alert{0.9401} & \alert{1.3098} \\
\mc{3}{c}{\alert{$s_N$}}        & \alert{0.0002} & \alert{0.0008} & \alert{0.0044} & \alert{0.0034} & \alert{0.0010} \\
\hline
\textbf{Base} & (1234,1234) & 5 & 0.1680 & 0.2168 & 0.9396 & 0.9395 & 1.2421 \\
\alert{\textbf{Base}$^{b,c}$} & \alert{(1234,2345)} & \alert{5} & \alert{0.8962} & \alert{3.5788} & \alert{0.1504} & \alert{0.0393} & \alert{35.8305} \\
\textbf{Base} & (2345,1234) & 5 & 0.1669 & 0.2174 & 0.9380 & 0.9379 & 1.2428 \\
\hline
\mc{3}{c}{\alert{$\bar{x}$}}    & \alert{0.1675} & \alert{0.2171} & \alert{0.9388} & \alert{0.9387} & \alert{1.2425} \\
\mc{3}{c}{\alert{$s_N$}}        & \alert{0.0008} & \alert{0.0004} & \alert{0.0011} & \alert{0.0011} & \alert{0.0005} \\
\hline\hline
\insertTableNotes
\end{xltabular}
}
\label{TAB:PTMODELDATASIZE}
\end{ThreePartTable}

\section{Ablation Studies and Hyperparameter Tuning}
\label{SISEC:ABLATION}
\subsection{Supervised fine-tuning on the Biogen ADME Dataset}
\label{SISUBSEC:SFTADME}
The \ac{ADME} public dataset involves six \textit{in vitro} \ac{ADME} endpoints:
\ac{HLM}, \ac{RLM}, \ac{HPPB}, \ac{RPPB}, \ac{MDR1} and
solubility.\cite{Fang:2023:3263} All \ac{SMILES} are standardized using
PubChem's standardization protocol and tokenized using the WordPiece tokenizer.
The dataset is randomly split into a training and testing subsets with a ratio
of 80:20\% within each property. The training set is further divided into 3
folds for cross-validation following hyperparameter optimization using grid
search for classical \ac{ML} models and Bayesian search for \ac{BERT} models.
The best performing model is selected based on the $R^2$ metric obtained from
the cross-validation procedure. In addition to $R^2$, we include other
regression performance metrics such as \ac{MSE}, \ac{MAE}, \ac{RMSE} and Pearson
$R$ for further analysis.

Tables~\ref{SITAB:ADMEHLM}-\ref{SITAB:ADMEMDR1ER} summarize the regression
performance statistics of the Tiny-, Small- and Base-\ac{BERT} as well as
classical \ac{ML} models, fine-tuned on Biogen's \ac{ADME} dataset.
Numerical entries without parentheses correspond to the cross-validation
results, summarized as $\bar{x} \pm s_N$. The test set performance
metrics are presented as numerical entries within the parentheses.

\begin{table}
\centering
\setlength{\tabcolsep}{4pt} 
\caption{Supervised fine-tuning performance metrics obtained via 3-fold
cross-validation hyperparameter optimization on Biogen's public ADME dataset
with experimental HLM data as labels$^{a,b}$}
\label{SITAB:ADMEHLM}
\resizebox{\textwidth}{!}{
    \begin{tabular}{lcccc}
        \hline\hline
        \textbf{Models} & \textbf{MAE} $\downarrow$$^c$ & \textbf{RMSE} $\downarrow$$^c$  & \textbf{Pearson R} $\uparrow$$^c$ & \textbf{R$^2$} $\uparrow$$^c$ \\
        \hline
        \textbf{Lasso Regression}       & 0.403 $\pm$ 0.006 (0.407) & 0.498 $\pm$ 0.007 (0.518) & 0.602 $\pm$ 0.010 (0.567) & 0.360 $\pm$ 0.012 (0.318) \\
        \textbf{Support Vector Machine} & 0.371 $\pm$ 0.004 (0.356) & 0.466 $\pm$ 0.004 (0.452) & 0.667 $\pm$ 0.010 (0.696) & 0.441 $\pm$ 0.013 (0.479) \\
        \textbf{Random Forest}          & 0.399 $\pm$ 0.006 (0.385) & 0.492 $\pm$ 0.006 (0.483) & 0.621 $\pm$ 0.011 (0.641) & 0.378 $\pm$ 0.010 (0.406) \\
        \textbf{XGBoost}                & 0.380 $\pm$ 0.009 (0.377) & 0.482 $\pm$ 0.005 (0.489) & 0.640 $\pm$ 0.013 (0.640) & 0.402 $\pm$ 0.018 (0.392) \\
        \textbf{LightGBM}               & 0.368 $\pm$ 0.008 (0.366) & 0.466 $\pm$ 0.004 (0.472) & 0.665 $\pm$ 0.024 (0.661) & 0.440 $\pm$ 0.031 (0.434) \\
        \textbf{BERT-Tiny}              & 0.441 $\pm$ 0.017 (0.450) & 0.545 $\pm$ 0.009 (0.556) & 0.495 $\pm$ 0.033 (0.470) & 0.224 $\pm$ 0.042 (0.198) \\
        \textbf{BERT-Small}             & 0.370 $\pm$ 0.013 (0.354) & 0.483 $\pm$ 0.016 (0.471) & 0.647 $\pm$ 0.034 (0.672) & 0.386 $\pm$ 0.063 (0.418) \\
        \textbf{BERT-Base}              & 0.354 $\pm$ 0.004 (\textbf{0.333}) & 0.454 $\pm$ 0.004 (\textbf{0.427}) & 0.681 $\pm$ 0.021 (\textbf{0.728}) & 0.453 $\pm$ 0.038 (\textbf{0.523}) \\
        \hline\hline
    \end{tabular}
    }
\begin{tablenotes}
    \footnotesize
    \item $^a$ The statistics are summarized as $\bar{x} \pm s_N$ where $\bar{x}$ and
    $s_N$ are the mean and standard deviation of the calculated metrics over
    the 3 cross-validation folds. The testing performance metrics (shown in
    parentheses) are obtained from refitting the best models on the entire
    training set and evaluating them on the test set.
    \item $^b$ The best test results are shown in boldface.
    \item $^c$ $\uparrow$: higher is better; $\downarrow$: lower is better.
\end{tablenotes}
\end{table}

\begin{table}
\centering
\setlength{\tabcolsep}{4pt} 
\caption{Supervised fine-tuning performance metrics obtained via 3-fold
cross-validation hyperparameter optimization on Biogen's public ADME dataset
with experimental RLM data as labels$^{a,b}$}
\label{SITAB:ADMERLM}
\resizebox{\textwidth}{!}{
    \begin{tabular}{lcccc}
        \hline\hline
        \textbf{Models} & \textbf{MAE} $\downarrow$$^c$ & \textbf{RMSE} $\downarrow$$^c$  & \textbf{Pearson R} $\uparrow$$^c$ & \textbf{R$^2$} $\uparrow$$^c$ \\
        \hline
        \textbf{Lasso Regression}       & 0.442 $\pm$ 0.010 (0.440) & 0.558 $\pm$ 0.011 (0.563) & 0.676 $\pm$ 0.017 (0.680) & 0.445 $\pm$ 0.021 (0.452) \\
        \textbf{Support Vector Machine} & 0.414 $\pm$ 0.004 (0.416) & 0.523 $\pm$ 0.008 (0.535) & 0.718 $\pm$ 0.011 (0.714) & 0.512 $\pm$ 0.016 (0.505) \\
        \textbf{Random Forest}          & 0.453 $\pm$ 0.008 (0.454) & 0.562 $\pm$ 0.008 (0.568) & 0.667 $\pm$ 0.020 (0.669) & 0.437 $\pm$ 0.024 (0.442) \\
        \textbf{XGBoost}                & 0.436 $\pm$ 0.005 (0.431) & 0.548 $\pm$ 0.011 (0.553) & 0.687 $\pm$ 0.016 (0.693) & 0.464 $\pm$ 0.025 (0.471) \\
        \textbf{LightGBM}               & 0.417 $\pm$ 0.006 (0.409) & 0.527 $\pm$ 0.009 (0.532) & 0.710 $\pm$ 0.015 (0.715) & 0.504 $\pm$ 0.022 (0.511) \\
        \textbf{BERT-Tiny}              & 0.498 $\pm$ 0.007 (0.499) & 0.626 $\pm$ 0.011 (0.634) & 0.560 $\pm$ 0.023 (0.590) & 0.272 $\pm$ 0.037 (0.298) \\
        \textbf{BERT-Small}             & 0.444 $\pm$ 0.014 (0.437) & 0.563 $\pm$ 0.011 (0.562) & 0.683 $\pm$ 0.020 (0.707) & 0.414 $\pm$ 0.032 (0.446) \\
        \textbf{BERT-Base}              & 0.411 $\pm$ 0.013 (\textbf{0.399}) & 0.528 $\pm$ 0.016 (\textbf{0.510}) & 0.716 $\pm$ 0.019 (\textbf{0.755}) & 0.496 $\pm$ 0.034 (\textbf{0.542}) \\
        \hline\hline
    \end{tabular}
    }
\begin{tablenotes}
    \footnotesize
    \item $^a$ The statistics are summarized as $\bar{x} \pm s_N$ where $\bar{x}$ and
    $s_N$ are the mean and standard deviation of the calculated metrics over
    the 3 cross-validation folds. The testing performance metrics (shown in
    parentheses) are obtained from refitting the best models on the entire
    training set and evaluating them on the test set.
    \item $^b$ The best test results are shown in boldface.
    \item $^c$ $\uparrow$: higher is better; $\downarrow$: lower is better.
\end{tablenotes}
\end{table}

\begin{table}
\centering
\setlength{\tabcolsep}{4pt} 
\caption{Supervised fine-tuning performance metrics obtained via 3-fold
cross-validation hyperparameter optimization on Biogen's public ADME dataset
with experimental hPPB data as labels$^{a,b}$}
\label{SITAB:ADMEHPPB}
\resizebox{\textwidth}{!}{
    \begin{tabular}{lcccc}
        \hline\hline
        \textbf{Models} & \textbf{MAE} $\downarrow$$^c$ & \textbf{RMSE} $\downarrow$$^c$  & \textbf{Pearson R} $\uparrow$$^c$ & \textbf{R$^2$} $\uparrow$$^c$ \\
        \hline
        \textbf{Lasso Regression}       & 0.387 $\pm$ 0.010 (0.410) & 0.502 $\pm$ 0.013 (0.542) & 0.721 $\pm$ 0.010 (0.686) & 0.519 $\pm$ 0.015 (0.469) \\
        \textbf{Support Vector Machine} & 0.363 $\pm$ 0.014 (0.388) & 0.479 $\pm$ 0.018 (0.536) & 0.752 $\pm$ 0.010 (0.701) & 0.562 $\pm$ 0.016 (0.481) \\
        \textbf{Random Forest}          & 0.408 $\pm$ 0.015 (0.422) & 0.523 $\pm$ 0.018 (0.548) & 0.696 $\pm$ 0.019 (0.679) & 0.477 $\pm$ 0.024 (0.458) \\
        \textbf{XGBoost}                & 0.390 $\pm$ 0.005 (0.421) & 0.509 $\pm$ 0.010 (0.547) & 0.713 $\pm$ 0.006 (0.681) & 0.505 $\pm$ 0.008 (0.459) \\
        \textbf{LightGBM}               & 0.374 $\pm$ 0.010 (\textbf{0.408}) & 0.491 $\pm$ 0.015 (\textbf{0.529}) & 0.735 $\pm$ 0.008 (\textbf{0.703}) & 0.539 $\pm$ 0.011 (\textbf{0.494}) \\
        \textbf{BERT-Tiny}              & 0.464 $\pm$ 0.012 (0.494) & 0.595 $\pm$ 0.017 (0.612) & 0.573 $\pm$ 0.021 (0.564) & 0.312 $\pm$ 0.015 (0.312) \\
        \textbf{BERT-Small}             & 0.393 $\pm$ 0.012 (0.452) & 0.507 $\pm$ 0.018 (0.601) & 0.716 $\pm$ 0.013 (0.638) & 0.495 $\pm$ 0.020 (0.344) \\
        \textbf{BERT-Base}              & 0.372 $\pm$ 0.009 (0.409) & 0.484 $\pm$ 0.007 (0.549) & 0.741 $\pm$ 0.003 (0.688) & 0.537 $\pm$ 0.008 (0.454) \\        
        \hline\hline
    \end{tabular}
    }
\begin{tablenotes}
    \footnotesize
    \item $^a$ The statistics are summarized as $\bar{x} \pm s_N$ where $\bar{x}$ and
    $s_N$ are the mean and standard deviation of the calculated metrics over
    the 3 cross-validation folds. The testing performance metrics (shown in
    parentheses) are obtained from refitting the best models on the entire
    training set and evaluating them on the test set.
    \item $^b$ The best test results are shown in boldface.
    \item $^c$ $\uparrow$: higher is better; $\downarrow$: lower is better.
\end{tablenotes}
\end{table}

\begin{table}
\centering
\setlength{\tabcolsep}{4pt} 
\caption{Supervised fine-tuning performance metrics obtained via 3-fold
cross-validation hyperparameter optimization on Biogen's public ADME dataset
with experimental rPPB data as labels$^{a,b}$}
\label{SITAB:ADMERPPB}
\resizebox{\textwidth}{!}{
    \begin{tabular}{lcccc}
        \hline\hline
        \textbf{Models} & \textbf{MAE} $\downarrow$$^c$ & \textbf{RMSE} $\downarrow$$^c$  & \textbf{Pearson R} $\uparrow$$^c$ & \textbf{R$^2$} $\uparrow$$^c$ \\
        \hline
        \textbf{Lasso Regression}       & 0.430 $\pm$ 0.021 (0.416) & 0.565 $\pm$ 0.032 (0.534) & 0.681 $\pm$ 0.033 (0.697) & 0.453 $\pm$ 0.044 (0.474) \\
        \textbf{Support Vector Machine} & 0.411 $\pm$ 0.014 (\textbf{0.381}) & 0.550 $\pm$ 0.031 (\textbf{0.512}) & 0.696 $\pm$ 0.029 (\textbf{0.720}) & 0.481 $\pm$ 0.042 (\textbf{0.516}) \\
        \textbf{Random Forest}          & 0.450 $\pm$ 0.017 (0.420) & 0.576 $\pm$ 0.030 (0.529) & 0.667 $\pm$ 0.042 (0.702) & 0.430 $\pm$ 0.044 (0.484) \\
        \textbf{XGBoost}                & 0.430 $\pm$ 0.014 (0.425) & 0.570 $\pm$ 0.028 (0.543) & 0.668 $\pm$ 0.026 (0.681) & 0.443 $\pm$ 0.036 (0.457) \\
        \textbf{LightGBM}               & 0.427 $\pm$ 0.019 (0.417) & 0.558 $\pm$ 0.026 (0.548) & 0.684 $\pm$ 0.024 (0.671) & 0.466 $\pm$ 0.033 (0.447) \\
        \textbf{BERT-Tiny}              & 0.525 $\pm$ 0.026 (0.514) & 0.649 $\pm$ 0.040 (0.620) & 0.531 $\pm$ 0.054 (0.560) & 0.259 $\pm$ 0.062 (0.287) \\
        \textbf{BERT-Small}             & 0.468 $\pm$ 0.043 (0.422) & 0.601 $\pm$ 0.048 (0.558) & 0.616 $\pm$ 0.047 (0.673) & 0.363 $\pm$ 0.056 (0.422) \\
        \textbf{BERT-Base}              & 0.435 $\pm$ 0.029 (0.401) & 0.567 $\pm$ 0.032 (0.526) & 0.670 $\pm$ 0.024 (0.700) & 0.442 $\pm$ 0.036 (0.487) \\
        \hline\hline
    \end{tabular}
    }
\begin{tablenotes}
    \footnotesize
    \item $^a$ The statistics are summarized as $\bar{x} \pm s_N$ where $\bar{x}$ and
    $s_N$ are the mean and standard deviation of the calculated metrics over
    the 3 cross-validation folds. The testing performance metrics (shown in
    parentheses) are obtained from refitting the best models on the entire
    training set and evaluating them on the test set.
    \item $^b$ The best test results are shown in boldface.
    \item $^c$ $\uparrow$: higher is better; $\downarrow$: lower is better.
\end{tablenotes}
\end{table}

\begin{table}
\centering
\setlength{\tabcolsep}{4pt} 
\caption{Supervised fine-tuning performance metrics obtained via 3-fold
cross-validation hyperparameter optimization on Biogen's public ADME dataset
with experimental solubility data as labels$^{a,b}$}
\label{SITAB:ADMESOL}
\resizebox{\textwidth}{!}{
    \begin{tabular}{lcccc}
        \hline\hline
        \textbf{Models} & \textbf{MAE} $\downarrow$$^c$ & \textbf{RMSE} $\downarrow$$^c$  & \textbf{Pearson R} $\uparrow$$^c$ & \textbf{R$^2$} $\uparrow$$^c$ \\
        \hline
        \textbf{Lasso Regression}       & 0.429 $\pm$ 0.015 (0.416) & 0.574 $\pm$ 0.019 (0.569) & 0.551 $\pm$ 0.002 (0.535) & 0.290 $\pm$ 0.008 (0.282) \\
        \textbf{Support Vector Machine} & 0.414 $\pm$ 0.020 (0.390) & 0.565 $\pm$ 0.025 (0.551) & 0.569 $\pm$ 0.010 (0.572) & 0.313 $\pm$ 0.014 (0.326) \\
        \textbf{Random Forest}          & 0.416 $\pm$ 0.010 (0.388) & 0.575 $\pm$ 0.021 (0.558) & 0.553 $\pm$ 0.011 (0.565) & 0.289 $\pm$ 0.006 (0.310) \\
        \textbf{XGBoost}                & 0.413 $\pm$ 0.005 (0.404) & 0.584 $\pm$ 0.010 (0.593) & 0.532 $\pm$ 0.020 (0.497) & 0.264 $\pm$ 0.033 (0.221) \\
        \textbf{LightGBM}               & 0.394 $\pm$ 0.008 (0.371) & 0.563 $\pm$ 0.014 (0.549) & 0.571 $\pm$ 0.026 (0.577) & 0.316 $\pm$ 0.031 (0.331) \\
        \textbf{BERT-Tiny}              & 0.424 $\pm$ 0.020 (0.403) & 0.592 $\pm$ 0.029 (0.594) & 0.490 $\pm$ 0.022 (0.467) & 0.231 $\pm$ 0.019 (0.209) \\
        \textbf{BERT-Small}             & 0.390 $\pm$ 0.040 (0.360) & 0.550 $\pm$ 0.036 (0.562) & 0.594 $\pm$ 0.039 (0.563) & 0.341 $\pm$ 0.039 (0.290) \\
        \textbf{BERT-Base}              & 0.390 $\pm$ 0.040 (\textbf{0.342}) & 0.559 $\pm$ 0.038 (\textbf{0.529}) & 0.578 $\pm$ 0.026 (\textbf{0.627}) & 0.322 $\pm$ 0.035 (\textbf{0.376}) \\
        \hline\hline
    \end{tabular}
    }
\begin{tablenotes}
    \footnotesize
    \item $^a$ The statistics are summarized as $\bar{x} \pm s_N$ where $\bar{x}$ and
    $s_N$ are the mean and standard deviation of the calculated metrics over
    the 3 cross-validation folds. The testing performance metrics (shown in
    parentheses) are obtained from refitting the best models on the entire
    training set and evaluating them on the test set.
    \item $^b$ The best test results are shown in boldface.
    \item $^c$ $\uparrow$: higher is better; $\downarrow$: lower is better.
\end{tablenotes}
\end{table}

\begin{table}
\centering
\setlength{\tabcolsep}{4pt} 
\caption{Supervised fine-tuning performance metrics obtained via 3-fold
cross-validation hyperparameter optimization on Biogen's public ADME dataset
with experimental MDR1-ER data as labels$^{a,b}$}
\label{SITAB:ADMEMDR1ER}
\resizebox{\textwidth}{!}{
    \begin{tabular}{lcccc}
        \hline\hline
        \textbf{Models} & \textbf{MAE} $\downarrow$$^c$ & \textbf{RMSE} $\downarrow$$^c$  & \textbf{Pearson R} $\uparrow$$^c$ & \textbf{R$^2$} $\uparrow$$^c$ \\
        \hline
        \textbf{Lasso Regression}       & 0.373 $\pm$ 0.007 (0.383) & 0.485 $\pm$ 0.008 (0.516) & 0.708 $\pm$ 0.010 (0.677) & 0.499 $\pm$ 0.014 (0.458) \\
        \textbf{Support Vector Machine} & 0.341 $\pm$ 0.008 (0.338) & 0.457 $\pm$ 0.011 (0.476) & 0.747 $\pm$ 0.013 (0.736) & 0.555 $\pm$ 0.018 (0.539) \\
        \textbf{Random Forest}          & 0.390 $\pm$ 0.007 (0.386) & 0.502 $\pm$ 0.004 (0.502) & 0.686 $\pm$ 0.008 (0.709) & 0.462 $\pm$ 0.011 (0.488) \\
        \textbf{XGBoost}                & 0.359 $\pm$ 0.005 (0.351) & 0.480 $\pm$ 0.006 (0.482) & 0.716 $\pm$ 0.010 (0.728) & 0.510 $\pm$ 0.013 (0.527) \\
        \textbf{LightGBM}               & 0.353 $\pm$ 0.005 (0.340) & 0.469 $\pm$ 0.006 (0.471) & 0.730 $\pm$ 0.007 (0.742) & 0.531 $\pm$ 0.009 (0.549) \\
        \textbf{BERT-Tiny}              & 0.408 $\pm$ 0.008 (0.402) & 0.539 $\pm$ 0.016 (0.541) & 0.622 $\pm$ 0.025 (0.640) & 0.376 $\pm$ 0.035 (0.393) \\
        \textbf{BERT-Small}             & 0.365 $\pm$ 0.007 (0.386) & 0.486 $\pm$ 0.012 (0.511) & 0.711 $\pm$ 0.013 (0.693) & 0.489 $\pm$ 0.034 (0.457) \\
        \textbf{BERT-Base}              & 0.341 $\pm$ 0.015 (\textbf{0.319}) & 0.457 $\pm$ 0.022 (\textbf{0.448}) & 0.744 $\pm$ 0.020 (\textbf{0.772}) & 0.535 $\pm$ 0.026 (\textbf{0.583}) \\
        \hline\hline
    \end{tabular}
    }
\begin{tablenotes}
    \footnotesize
    \item $^a$ The statistics are summarized as $\bar{x} \pm s_N$ where $\bar{x}$ and
    $s_N$ are the mean and standard deviation of the calculated metrics over
    the 3 cross-validation folds. The testing performance metrics (shown in
    parentheses) are obtained from refitting the best models on the entire
    training set and evaluating them on the test set.
    \item $^b$ The best test results are shown in boldface.
    \item $^c$ $\uparrow$: higher is better; $\downarrow$: lower is better.
\end{tablenotes}
\end{table}

\pagebreak

\subsection{Hyperparameter Search Spaces for Classical ML Models}
\label{SISUBSEC:SEARCHSPACEML}
The pre-defined search space for the hyperparameter optimization of the
classical \ac{ML} on Biogen's \ac{ADME} dataset is provided in Table
\ref{SITAB:CLASSICSEACHPARAMS} which closely follows that of
Ref.~\citenum{Fang:2023:3263}.

\begin{table}
\centering
\setlength{\tabcolsep}{4pt} 
\caption{Pre-defined search space for grid-search hyperparameter optimization
of classical machine learning models performed over ADME dataset$^a$}
\label{SITAB:CLASSICSEACHPARAMS}
\resizebox{\textwidth}{!}{
    \begin{tabular}{lll}
        \hline\hline
        \textbf{Model} & \textbf{Parameter} & \textbf{Allowed Values} \\
        \hline
        \mr{6}{*}{\textbf{BERT}} & \textbf{stack\_depth}        & $\lbrace$0, 1, 2, 3, 4, 5$\rbrace$ \\
                                 & \textbf{p}                   & $\lbrace$0.0, 0.1, 0.2, 0.3, 0.4, 0.5$\rbrace$ \\
                                 & \textbf{batch\_size}         & $\lbrace$8, 16, 32, 64, 128, 256, 512$\rbrace$ \\
                                 & \textbf{learning\_rate}      & $\lbrace$1E-06, 1E-05, 5E-05, 1E-04, 5E-04, 1E-03$\rbrace$ \\
                                 & \textbf{weight\_decay}       & $\lbrace$0.0, 1E-04, 5E-04, 1E-03, 5E-03, 1E-02, 5E-02, 1E-01$\rbrace$ \\
                                 & \textbf{freeze\_base\_model} & $\lbrace$True, False$\rbrace$ \\
        \hline
        \textbf{LASSO}           & \textbf{LASSO\_\_alpha}      & $\lbrace$0.001, 0.005, 0.01, 0.05, 0.1, 0.5, 1.0, 2.0, 5.0$\rbrace$ \\
        \hline
        \mr{3}{*}{\textbf{SVM}}  & \textbf{SVM\_\_C}            & $\lbrace$0.1, 1.0, 5.0, 10.0, 20.0, 50.0$\rbrace$ \\
                                 & \textbf{SVM\_\_epsilon}      & $\lbrace$0.01, 0.1, 0.3, 0.5$\rbrace$ \\
                                 & \textbf{SVM\_\_gamma}        & $\lbrace$``scale'', ``auto''$\rbrace$ \\
        \hline
        \mr{3}{*}{\textbf{RF}}   & \textbf{max\_depth}          & $\lbrace$15, 25, 40, None$\rbrace$ \\
                                 & \textbf{max\_features}       & $\lbrace$``sqrt'', 0.33, 0.67, None$\rbrace$ \\
                                 & \textbf{n\_estimators}       & $\lbrace$100, 250, 500, 750, 1000$\rbrace$ \\
        \hline
        \mr{8}{*}{\textbf{XGBoost}$^b$}  & \textbf{n\_estimators (Round 1)}       & $\lbrace$100, 250, 500, 750, 1000$\rbrace$ \\
                                         & \textbf{max\_depth (Round 2)}          & $\lbrace$3, 4, 5, 6, 7$\rbrace$ \\
                                         & \textbf{min\_child\_weight (Round 2)}  & $\lbrace$1, 2, 3$\rbrace$ \\
                                         & \textbf{gamma (Round 3)}               & $\lbrace$0.0, 0.05, 0.1$\rbrace$ \\
                                         & \textbf{subsample (Round 4)}           & $\lbrace$0.6, 0.7, 0.8, 0.9, 1.0$\rbrace$ \\
                                         & \textbf{colsample\_bytree (Round 4)}   & $\lbrace$0.6, 0.7, 0.8, 0.9, 1.0$\rbrace$ \\
                                         & \textbf{reg\_alpha (Round 5)}          & $\lbrace$0.0, 0.1, 0.2, 0.3, 0.4$\rbrace$ \\
                                         & \textbf{reg\_lambda (Round 5)}         & $\lbrace$1.0, 1.1, 1.2, 1.3, 1.4$\rbrace$ \\
        \hline
        \mr{8}{*}{\textbf{LightGBM}$^c$} & \textbf{n\_estimators (Round 1)}       & $\lbrace$100, 250, 500, 750, 1000$\rbrace$ \\
                                         & \textbf{num\_leaves (Round 2)}         & $\lbrace$15, 31, 45, 60, 75$\rbrace$ \\
                                         & \textbf{min\_child\_samples (Round 2)} & $\lbrace$10, 20, 30, 40$\rbrace$ \\
                                         & \textbf{subsample (Round 3)}           & $\lbrace$0.6, 0.7, 0.8, 0.9, 1.0$\rbrace$ \\
                                         & \textbf{colsample\_bytree (Round 3)}   & $\lbrace$0.6, 0.7, 0.8, 0.9, 1.0$\rbrace$ \\
                                         & \textbf{subsample\_freq (Round 3)}     & $\lbrace$1, 2, 3, 5$\rbrace$ \\
                                         & \textbf{reg\_alpha (Round 4)}          & $\lbrace$0.0, 0.2, 0.5, 0.8$\rbrace$ \\
                                         & \textbf{reg\_lambda (Round 4)}         & $\lbrace$0.0, 0.2, 0.5, 0.8$\rbrace$ \\
                                      
        \hline\hline
    \end{tabular}
}
\begin{tablenotes}
    \footnotesize
    \item $^a$ In order to reduce the computational cost of hyperparameter
    optimization, we followed Ref.\citenum{Fang:2023:3263} and performed a
    multi-step grid-search optimization for the XGBoost and LightGBM models. In
    each round of the grid-search, we optimized a subset of the total set of
    hyperparameters while keeping the rest of the hyperparameters fixed.
    \item $^b$ Initial parameters were set to: n\_estimators = 500; subsample =
    0.8; colsample\_bytree = 0.8; random\_state = 1234.
    \item $^c$ Initial parameters were set to: n\_estimators = 500; subsample =
    0.8; colsample\_bytree = 0.8; subsample\_freq = 1; random\_state = 1234.
\end{tablenotes}
\end{table}

\pagebreak
\subsection{Optimized Hyperparameters of the Fine-tuned Models}
\label{SISUBSEC:HYPERPARAMML}
Tables \ref{SITAB:LASSOPARAMS}--\ref{SITAB:LLMSEARCHPARAMS} provide a complete
list of optimized hyperparameters for all fine-tuned models considered in this
study.

\begin{table}
\centering
\setlength{\tabcolsep}{4pt} 
\caption{Optimized parameters for the LASSO model obtained from grid-search
    hyperparameter optimization with 3-fold cross-validation using ADME
    dataset}
\label{SITAB:LASSOPARAMS}
    \begin{tabular}{lcccccc}
        \hline\hline
        \mr{2}{*}{\textbf{Parameter}} & \mc{6}{c}{\textbf{Target Property}} \\
        \cline{2-7}
        & \textbf{HLM} & \textbf{RLM} & \textbf{hPPB} & \textbf{rPPB} & \textbf{SOL} & \textbf{MDR1-ER} \\
        \hline
        \textbf{LASSO\_\_alpha}  & 0.005 & 0.001 & 0.005 & 0.005 & 0.005 & 0.005 \\
        \hline\hline
    \end{tabular}
\end{table}

\begin{table}[!htpb]
\centering
\setlength{\tabcolsep}{4pt} 
\caption{Optimized hyperparameters for the support vector machine model
    obtained from grid-search hyperparameter optimization with 3-fold
    cross-validation using ADME dataset}
\label{SITAB:SVMPARAMS}
    \begin{tabular}{lcccccc}
        \hline\hline
        \mr{2}{*}{\textbf{Parameter}} & \mc{6}{c}{\textbf{Target Property}} \\
        \cline{2-7}
         & \textbf{HLM} & \textbf{RLM} & \textbf{hPPB} & \textbf{rPPB} & \textbf{SOL} & \textbf{MDR1-ER} \\
        \hline
        \textbf{SVM\_\_C}       & 5         & 5        & 10       & 5        & 5        & 5        \\
        \textbf{SVM\_\_epsilon} & 0.1       & 0.1      & 0.1      & 0.1      & 0.3      & 0.1      \\
        \textbf{SVM\_\_gamma}   & ``scale'' & ``auto'' & ``auto'' & ``auto'' & ``auto'' & ``auto'' \\
        \hline\hline
    \end{tabular}
\end{table}

\begin{table}
\centering
\setlength{\tabcolsep}{4pt} 
\caption{Optimized hyperparameters for the random forest model obtained from
    grid-search hyperparameter optimization with 3-fold cross-validation using
    ADME dataset}
\label{SITAB:RFPARAMS}
    \begin{tabular}{lcccccc}
        \hline\hline
        \mr{2}{*}{\textbf{Parameter}} & \mc{6}{c}{\textbf{Target Property}} \\
        \cline{2-7}
        & \textbf{HLM} & \textbf{RLM} & \textbf{hPPB} & \textbf{rPPB} & \textbf{SOL} & \textbf{MDR1-ER} \\
        \hline
        \textbf{max\_depth}      & 15   & 40   & 25   & 25   & None & None \\
        \textbf{max\_features}   & 0.67 & None & 0.67 & 0.67 & 0.33 & 0.67 \\
        \textbf{n\_estimators}   & 750  & 500  & 1000 & 1000 & 500  & 1000 \\
        \hline\hline
    \end{tabular}
\end{table}

\begin{table}
\centering
\setlength{\tabcolsep}{4pt} 
\caption{Optimized hyperparameters for the XGBoost model obtained from
grid-search hyperparameter optimization with 3-fold cross-validation using
ADME dataset$^{a,b}$}
\label{SITAB:XGBPARAMS}
    \begin{tabular}{lcccccc}
        \hline\hline
        \mr{2}{*}{\textbf{Parameter}} & \mc{6}{c}{\textbf{Target Property}} \\
        \cline{2-7}
        & \textbf{HLM} & \textbf{RLM} & \textbf{hPPB} & \textbf{rPPB} & \textbf{SOL} & \textbf{MDR1-ER} \\
        \hline
        \textbf{n\_estimators (Round 1)}        & 750 & 750 & 100 & 100 & 750 & 1000 \\
        \textbf{max\_depth (Round 2)}           & 3   & 3   & 3   & 3   & 5   & 4    \\
        \textbf{min\_child\_weight (Round 2)}   & 1   & 2   & 2   & 2   & 2   & 1    \\
        \textbf{gamma (Round 3)}                & 0.0 & 0.0 & 0.0 & 0.0 & 0.0 & 0.1  \\
        \textbf{colsample\_bytree (Round 4)}    & 0.8 & 0.9 & 0.7 & 0.9 & 0.6 & 0.6  \\
        \textbf{subsample (Round 4)}            & 0.9 & 0.9 & 0.9 & 1.0 & 1.0 & 1.0  \\
        \textbf{reg\_alpha (Round 5)}           & 0.3 & 0.1 & 0.0 & 0.1 & 0.1 & 0.1  \\
        \textbf{reg\_lambda (Round 5)}          & 1.4 & 1.0 & 1.0 & 1.0 & 1.2 & 1.3  \\
        \hline\hline
    \end{tabular}
\begin{tablenotes}
    \footnotesize
    \item $^a$ In order to reduce the computational cost of hyperparameter
    optimization, we followed Ref.\citenum{Fang:2023:3263} and performed a
    multi-step grid-search optimization for the XGBoost. In each round of the
    grid-search, we optimized a subset of the total set of hyperparameters
    while keeping the rest of the hyperparameters fixed to their initial
    values.
    \item $^b$ Initial parameters were set to: n\_estimators = 500; subsample =
    0.8; colsample\_bytree = 0.8; random\_state = 1234.
\end{tablenotes}
\end{table}

\begin{table}
\centering
\setlength{\tabcolsep}{4pt} 
\caption{Optimized hyperparameters for the LightGBM model obtained from
grid-search hyperparameter optimization with 3-fold cross-validation using
ADME dataset$^{a,b}$}
\label{SITAB:LGBMPARAMS}
    \begin{tabular}{lcccccc}
        \hline\hline
        \mr{2}{*}{\textbf{Parameter}} & \mc{6}{c}{\textbf{Target Property}} \\
        \cline{2-7}
         & \textbf{HLM} & \textbf{RLM} & \textbf{hPPB} & \textbf{rPPB} & \textbf{SOL} & \textbf{MDR1-ER} \\
        \hline
        \textbf{n\_estimators (Round 1)}         & 1000 & 750 & 1000 & 100 & 500 & 100 \\
        \textbf{min\_child\_samples (Round 2)}   & 40   & 10  & 20   & 10  & 10  & 20  \\
        \textbf{num\_leaves (Round 2)}           & 75   & 15  & 15   & 15  & 15  & 15  \\
        \textbf{colsample\_bytree (Round 3)}     & 0.9  & 1.0 & 0.6  & 0.8 & 1.0 & 0.8 \\
        \textbf{subsample (Round 3)}             & 0.8  & 0.9 & 0.9  & 0.8 & 0.9 & 1.0 \\
        \textbf{subsample\_freq (Round 3)}       & 3    & 1   & 1    & 5   & 1   & 3   \\
        \textbf{reg\_alpha (Round 4)}            & 0.2  & 0.0 & 0.0  & 0.2 & 0.0 & 0.0 \\
        \textbf{reg\_lambda (Round 4)}           & 0.5  & 0.0 & 0.0  & 0.5 & 0.0 & 0.0 \\
        \hline\hline
    \end{tabular}
\begin{tablenotes}
    \footnotesize
    \item $^a$ In order to reduce the computational cost of hyperparameter
    optimization, we followed Ref.\citenum{Fang:2023:3263} and performed a
    multi-step grid-search optimization for the LightGBM. In each round of the
    grid-search, we optimized a subset of the total set of hyperparameters
    while keeping the rest of the hyperparameters fixed to their initial
    values.
    \item $^b$ Initial parameters were set to: n\_estimators = 500; subsample =
    0.8; colsample\_bytree = 0.8; subsample\_freq = 1; random\_state = 1234.
\end{tablenotes}
\end{table}

\begin{table}
\centering
\setlength{\tabcolsep}{4pt} 
\caption{Optimized hyperparameters for different variants of BERT model obtained
from conducting Bayesian search hyperparameter optimization with 3-fold
cross-validation over ADME dataset}
\label{SITAB:LLMSEARCHPARAMS}
    \begin{tabular}{llcccccc}
        \hline\hline
        \mr{2}{*}{\textbf{Model}} & \mr{2}{*}{\textbf{Parameter}} & \mc{6}{c}{\textbf{Target Property}} \\
        \cline{3-8}
        & & \textbf{HLM} & \textbf{RLM} & \textbf{hPPB} & \textbf{rPPB} & \textbf{SOL} & \textbf{MDR1-ER} \\
        \hline
        \mr{6}{*}{\textbf{Tiny-BERT}}   & \textbf{stack\_depth}        & 5     & 1     & 5     & 5     & 1      & 2     \\
                                        & \textbf{p}                   & 0.3   & 0.4   & 0.0   & 0.4   & 0.4    & 0.5   \\
                                        & \textbf{batch\_size}         & 128   & 256   & 128   & 64    & 64     & 128   \\
                                        & \textbf{learning\_rate}      & 5E-04 & 1E-04 & 5E-05 & 1E-04 & 5E-05  & 5E-05 \\
                                        & \textbf{weight\_decay}       & 0.05  & 0.00  & 0.01  & 0.1   & 0.0001 & 0.01  \\
                                        & \textbf{freeze\_base\_model} & True  & False & False & False & False  & False \\
        \hline                                        
        \mr{6}{*}{\textbf{Small-BERT}}  & \textbf{stack\_depth}        & 0     & 1      & 3     & 5      & 3     & 3      \\
                                        & \textbf{p}                   & 0.5   & 0.3    & 0.2   & 0.3    & 0.4   & 0.4    \\
                                        & \textbf{batch\_size}         & 64    & 32     & 128   & 64     & 64    & 256    \\
                                        & \textbf{learning\_rate}      & 1E-04 & 5E-05  & 1E-04 & 1E-04  & 1E-03 & 1E-03  \\
                                        & \textbf{weight\_decay}       & 0.1   & 0.0005 & 0.01  & 0.0005 & 0.001 & 0.0005 \\
                                        & \textbf{freeze\_base\_model} & False & False  & False & False  & True  & True   \\
        \hline
        \mr{6}{*}{\textbf{Base-BERT}}   & \textbf{stack\_depth}        & 0      & 2     & 3      & 4     & 3     & 0      \\
                                        & \textbf{p}                   & 0.3    & 0.2   & 0.2    & 0.4   & 0.5   & 0.3    \\
                                        & \textbf{batch\_size}         & 128    & 64    & 256    & 128   & 512   & 32     \\
                                        & \textbf{learning\_rate}      & 5E-05  & 5E-05 & 1E-04  & 1E-04 & 1E-03 & 1E-04  \\
                                        & \textbf{weight\_decay}       & 0.0005 & 0.1   & 0.0005 & 0.1   & 0.1   & 0.0005 \\
                                        & \textbf{freeze\_base\_model} & False  & False & False  & False & True  & False  \\
        \hline\hline
    \end{tabular}
\end{table}

\pagebreak
\section{Budget and Cost Estimations}
\label{SISEC:BUDGET}
In this section, we offer an estimated list of costs associated with
pre-training and fine-tuning \ac{BERT}. Using Table \ref{SITAB:AWS}, one can
estimate the pre-training and evaluation cost of Base-\ac{BERT} on the PubChem
dataset for 10 epochs with a global batch size of 1024 on an AWS
\texttt{p5en.48xlarge} instance with 8 NVIDIA H200 GPUs to be around \$10,000
for roughly 6.5 days. Similarly, the total cost of pre-training and evaluation
of Small-BERT on two \texttt{g6e.12xlarge} nodes with a total of 8 NVIDIA L40S
GPUs is about \$2,700 for 5.3 days. Finally, pre-training and evaluation of
Tiny-BERT takes about 2.3 days on 4 nodes with a total of 16 NVIDIA A30 GPUs.
The corresponding cost on AWS with four \texttt{g6e.12xlarge} instances,
assuming that the total runtime remains the same, is around \$2,300.
On the other hand, the entire 3-fold cross-validation hyperparameter search on
the same dataset, using the \ac{SVM} model, can be completed in a few minutes on
a personal multi-core CPU computer. Pre-training 160 \ac{BERT} models and their
subsequent fine-tunings took us about 18 months on a university-wide shared
supercomputing cluster as a consistent and airtight implementation of the
experiments across partitions with carefully controlled environment was the only
viable option for us to enable a meaningful and systematic exploration in our
study. We hope releasing all our codes, scripts, data and model artifacts to the
public can facilitate transparency and reproducibility of our experiments and
enable further research in this area.

\begin{table}
\centering
\setlength{\tabcolsep}{4pt} 
\caption{Estimated costs required for pre-training BERT models on the entire
 PubChem (version 4-18-2025) dataset}
\label{SITAB:AWS}
\resizebox{\textwidth}{!}{
    \begin{tabular}{lcccccc}
        \hline\hline
        \textbf{Instance} & \textbf{vCPU}   & \textbf{GPU}   & \textbf{On-demand}        & \textbf{GPU}         & \textbf{Instance}    & \textbf{Instance}     \\
        \textbf{Name}     & \textbf{Count} & \textbf{Types} & \textbf{Price (\$/h)$^a$} & \textbf{Memory (GB)} & \textbf{Memory (GB)} & \textbf{Storage (GB)} \\
        \hline
        \textbf{p5en.48xlarge}  & 192 & 8 $\times$ H200 & 63.30  & 8 $\times$ 141 (HBM3e) & 2048 & 8 $\times$ 3840 NVMe SSD  \\        
        \textbf{p4de.24xlarge}  & 96  & 8 $\times$ A100 & 27.45  & 8 $\times$ 80 (HBM2e)  & 1152 & 8 $\times$ 1000 NVMe SSD  \\
        \textbf{g6e.12xlarge}   & 48  & 4 $\times$ L40S & 10.49  & 4 $\times$ 48          & 384  & 2 $\times$ 1900 NVMe SSD  \\
        \hline\hline
    \end{tabular}
}
\begin{tablenotes}
    \footnotesize
    \item $^a$ On-demand prices are based on per-instance-hour and are obtained
    from AWS website (\url{https://aws.amazon.com/ec2/pricing/on-demand}) on
    03/03/2026 and are intended for illustrative purposes only. Prices may vary
    based on the region, availability, and other factors.
\end{tablenotes}
\end{table}

\pagebreak
\section{BERT Architectural Details}
\label{SISEC:ARCHITECTURE}

Table \ref{SITAB:BERTARCH} provides the architectural details of Tiny-, Small-
and Base-BERT model variants.

\begin{table}
\centering
\setlength{\tabcolsep}{4pt} 
\caption{Architectural details of different BERT model variants}
\label{SITAB:BERTARCH}
    \begin{tabular}{lccc}
        \hline\hline
        \mr{2}{*}{\textbf{Model Details}}      & \mc{3}{c}{\textbf{BERT Variant}}     \\
        \cline{2-4}
                                               & \textbf{Tiny} & \textbf{Small} & \textbf{Base} \\
        \hline
        \textbf{Hidden Size}                   & 128 & 512 & 768 \\
        \textbf{Hidden Activation}             & GELU & GELU & GELU \\
        \textbf{Vocabulary Size}               & 30522 & 30522 & 30522 \\
        \textbf{Max Position Embeddings}       & 512 & 512 & 512 \\
        \textbf{Intermediate Size}             & 512 & 2048 & 3072 \\
        \textbf{Hidden Dropout Probability}    & 0.1 & 0.1 & 0.1 \\
        \textbf{Attention Dropout Probability} & 0.1 & 0.1 & 0.1 \\
        \textbf{\# Hidden Layers}              & 2 & 4 & 12 \\ 
        \textbf{\# Attention Heads}            & 2 & 8 & 12 \\  
        \textbf{\# Parameters (M)}             & 4.4 & 28.5 & 108 \\
        
        \hline\hline
    \end{tabular}
\end{table}

\begin{acronym}
    \acro{ADME}{absorption, distribution, metabolism, and excretion}
    \acro{ARC}{Advanced Research Computing}
    \acro{BERT}{Bidirectional Encoder Representations from Transformers}
    \acro{BPE}{byte-pair encoding}
    \acro{CI}{confidence interval}
    \acro{DDP}{distributed data parallel}
    \acro{DFT}{Density Functional Theory}
    \acro{GPU}{graphics processing unit}
    \acro{HLM}{human liver microsomal}
    \acro{HPC}{high performance computing}
    \acro{HPPB}[hPPB]{human plasma protein binding}
    \acro{INCHI}[InChI]{International Chemical Identifier}
    \acro{LLM}{large language model}
    \acro{MAP}{maximum a posteriori}
    \acro{MDR1}[MDR1-MDCK ER]{MDR1-MDCK efflux ratio}
    \acro{ML}{machine learning}
    \acro{MLM}{masked language modeling}
    \acro{MPP}{molecular property prediction}
    \acro{MSE}{mean squared error}
    \acro{MAE}{mean absolute error}
    \acro{NLP}{natural language processing}
    \acro{NN}{neural network}
    \acro{PPPL}{pseudo-perplexity}
    \acro{QSAR}{quantitative structure-activity relationship}
    \acro{RLM}{rat liver microsomal}
    \acro{RMSE}{root mean squared error}
    \acro{RPPB}[rPPB]{rat plasma protein binding}
    \acro{SMILES}{Simplified Molecular Input Line Entry System}
    \acro{SVM}{support vector machine}
\end{acronym}
\pagebreak
\bibliography{supplement}


\begin{sidewaystable}
\centering
\setlength{\tabcolsep}{4pt} 
\caption{Variations of the masked language modeling performance metrics with
using different model weight initialization and data sampling random seeds$^a$}
\label{SITAB:RANDOMNESS}
\resizebox{\textwidth}{!}{
\begin{tabular}{ccccccc}
\hline\hline
 \textbf{BERT} & \textbf{(Model,Data)}  & \mr{2}{*}{\textbf{T-Loss} $\downarrow$$^c$} & \mr{2}{*}{\textbf{V-Loss} $\downarrow$$^c$}  & \mr{2}{*}{\textbf{V-Acc} $\uparrow$$^c$}  & \mr{2}{*}{\textbf{V-wF1} $\uparrow$$^c$} & \mr{2}{*}{\textbf{V-PPPL} $\downarrow$$^c$} \\
 \textbf{Variant} & \textbf{Seed} & & & & & \\
\hline
                        \textbf{Tiny}  & (1234,1234) & 0.6483 & 0.4664 & 0.8793 & 0.8637 & 1.5943 \\
                        \textbf{Tiny}  & (1234,2345) & 0.6560 & 0.4727 & 0.8783 & 0.8658 & 1.6042 \\
                        \textbf{Tiny}  & (1234,3456) & 0.6449 & 0.4663 & 0.8930 & 0.8799 & 1.5941 \\
                        \textbf{Tiny}  & (2345,1234) & 0.6449 & 0.4669 & 0.8823 & 0.8703 & 1.5950 \\
                        \textbf{Tiny}  & (3456,1234) & 0.6613 & 0.4755 & 0.8786 & 0.8634 & 1.6088 \\
                        \hline
                        \mc{2}{c}{$\bar{x} \pm s_N$}  & 0.6511 $\pm$ 0.0073 & 0.4696 $\pm$ 0.0043 & 0.8823 $\pm$ 0.0062 & 0.8686 $\pm$ 0.0069 & 1.5993 $\pm$ 0.0068 \\            
                        \hline            
                        \textbf{Small} & (1234,1234) & 0.2352 & 0.1835 & 0.9461 & 0.9417 & 1.2014 \\
                        \textbf{Small} & (1234,2345) & 0.2359 & 0.1844 & 0.9338 & 0.9281 & 1.2025 \\
                        \textbf{Small} & (1234,3456) & 0.2353 & 0.1836 & 0.9502 & 0.9454 & 1.2015 \\
                        \textbf{Small} & (2345,1234) & 0.2359 & 0.1841 & 0.9402 & 0.9375 & 1.2021 \\
                        \textbf{Small} & (3456,1234) & 0.2342 & 0.1826 & 0.9565 & 0.9532 & 1.2004 \\
                        \hline
                        \mc{2}{c}{$\bar{x} \pm s_N$}  & 0.2353 $\pm$ 0.0007 & 0.1836 $\pm$ 0.0007 & 0.9453 $\pm$ 0.0088 & 0.9412 $\pm$ 0.0093 & 1.2016 $\pm$ 0.0008  \\            
                        \hline            
                        \alert{\textbf{Base}$^b$} & \alert{(1234,1234)} & \alert{1.3880} & \alert{4.7239} & \alert{0.1301} & \alert{0.0299} & \alert{112.61} \\
                        \textbf{Base}  & (1234,2345) & 0.1775 & 0.1357 & 0.9557 & 0.9528 & 1.1454 \\
                        \textbf{Base}  & (1234,3456) & 0.1777 & 0.1359 & 0.9617 & 0.9575 & 1.1456 \\
                        \textbf{Base}  & (2345,1234) & 0.1782 & 0.1350 & 0.9619 & 0.9587 & 1.1446 \\
                        \textbf{Base}  & (3456,1234) & 0.1781 & 0.1368 & 0.9588 & 0.9567 & 1.1465 \\
                        \hline
                        \mc{2}{c}{$\bar{x} \pm s_N$}  & 0.1779 $\pm$ 0.0003 & 0.1359 $\pm$ 0.0007 & 0.9595 $\pm$ 0.0029 & 0.9564 $\pm$ 0.0026 & 1.1455 $\pm$ 0.0008  \\            
\hline\hline
\end{tabular}
}
\begin{tablenotes}
    \footnotesize
    \item $^a$ The entire PubChem (ver.~04/18/2025) dataset (119,184,806
    entries) is randomly split into training (95,347,844 entries amounting to
    80\% of the data) and validation (23,836,962 data points which counts as
    20\% of the data) sets. 
    \item $^b$ The model diverges during training and the behavior is persistent
    over several runs. As such, this experiment is excluded from calculating the
    mean ($\bar{x}$) and standard deviation ($s_N$).
    \item $^c$ $\uparrow$: higher is better; $\downarrow$: lower is better.
\end{tablenotes}
\end{sidewaystable}

\begin{figure}[!htbp]
    \setlength{\abovecaptionskip}{0.14cm}
    \centering
    \subfigure[]{
        \includegraphics[width=0.36\textwidth]{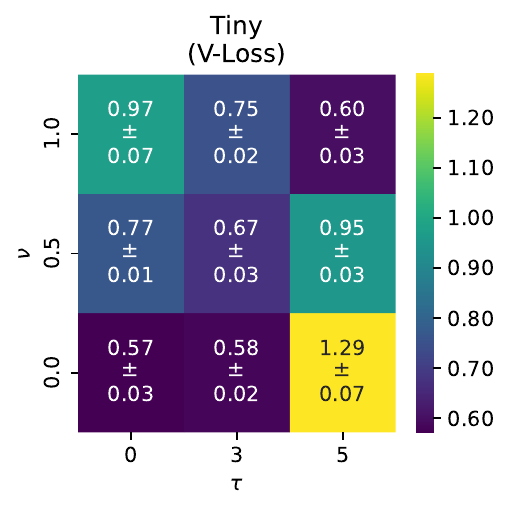}
    }
    \subfigure[]{
        \includegraphics[width=0.36\textwidth]{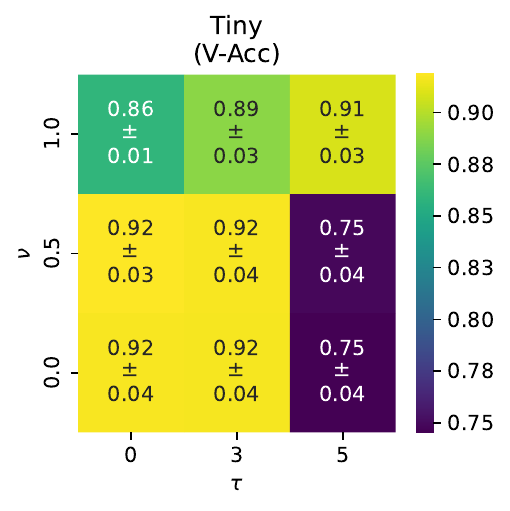}
    }
    \subfigure[]{
        \includegraphics[width=0.36\textwidth]{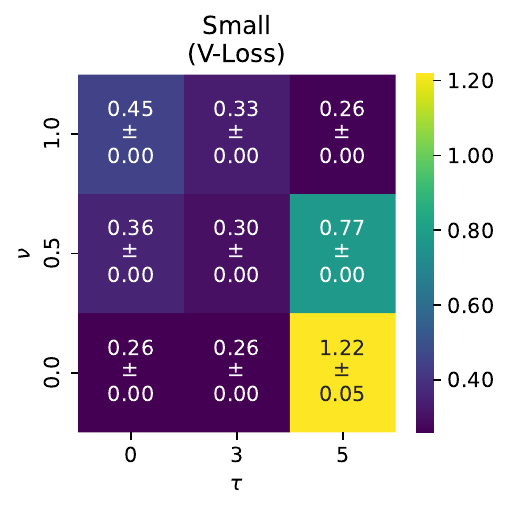}
    }    
    \subfigure[]{
        \includegraphics[width=0.36\textwidth]{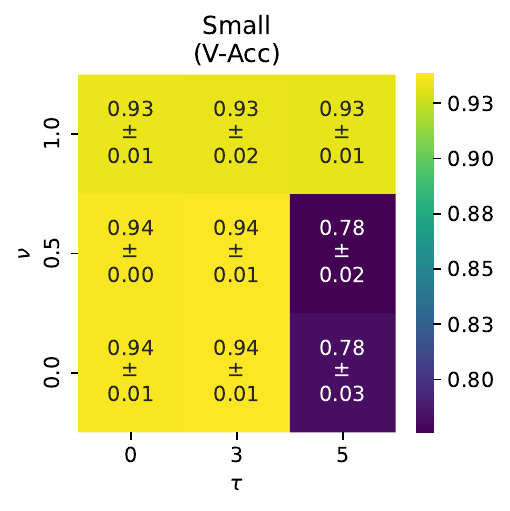}
    }
    \subfigure[]{
        \includegraphics[width=0.36\textwidth]{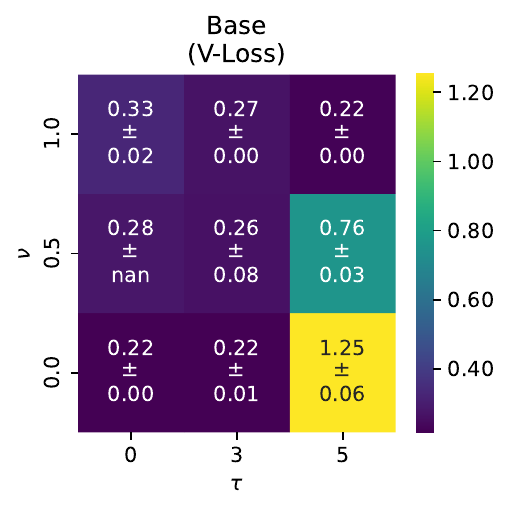}
    }
    \subfigure[]{
        \includegraphics[width=0.36\textwidth]{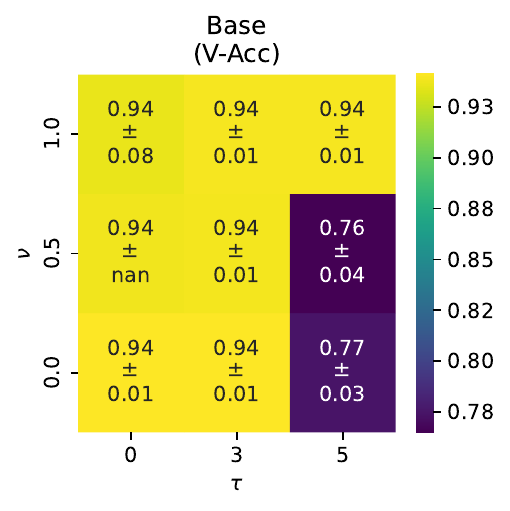}
    } \caption{The effect of standardization noise on the pre-training
    validation loss (a,c,e) and accuracy (b,d,f) of BERT for masked language
    modeling. Both metrics are averaged over three independent runs with
    different data sampling and model initialization seeds. The error bars are
    based on 95\% confidence interval which are not defined (NaN) for samples of
    $N=1$ runs, if the other two have diverged and removed from the calculating
    the statistical summaries. See \si\ for more details.}
    \label{EXFIG:PTSTDEFFECT}
\end{figure}

\begin{sidewaystable}
\centering
\setlength{\tabcolsep}{4pt} 
\caption{Variations of the masked language modeling performance metrics, for
pre-training multiple variants of BERT on canonical isomeric SMILES from
the PubChem (ver.~04/18/2025) dataset, tokenized with various tokenization
algorithms$^a$}
\label{SITAB:PTTOKENIZATION}
\resizebox{\textwidth}{!}{
\begin{tabular}{cccccccc}
\hline\hline
 \textbf{BERT} & \textbf{(Model,Data)} & \mr{2}{*}{\textbf{Tokenizer}} & \mr{2}{*}{\textbf{T-Loss} $\downarrow$$^c$} & \mr{2}{*}{\textbf{V-Loss} $\downarrow$$^c$}  & \mr{2}{*}{\textbf{V-Acc} $\uparrow$$^c$}  & \mr{2}{*}{\textbf{V-wF1} $\uparrow$$^c$} & \mr{2}{*}{\textbf{V-PPPL} $\downarrow$$^c$} \\
 \textbf{Variant} & \textbf{Seed} & & & & & \\
\hline
            \textbf{Tiny}  & (1234,1234) & WordPiece & 0.6483 & 0.4664 & 0.8793 & 0.8637 & 1.5943 \\
            \textbf{Tiny}  & (1234,2345) & WordPiece & 0.6560 & 0.4727 & 0.8783 & 0.8658 & 1.6042 \\
            \textbf{Tiny}  & (2345,1234) & WordPiece & 0.6449 & 0.4669 & 0.8823 & 0.8703 & 1.5950 \\
            \hline
            \mc{3}{c}{$\bar{x} \pm s_N$}  & 0.6497 $\pm$ 0.0057 & 0.4687 $\pm$ 0.0035 & 0.8800 $\pm$ 0.0021 & 0.8866 $\pm$ 0.0033 & 1.5978 $\pm$ 0.0056 \\            
            \hline            
            \textbf{Small} & (1234,1234) & WordPiece & 0.2352 & 0.1835 & 0.9461 & 0.9417 & 1.2014 \\
            \textbf{Small} & (1234,2345) & WordPiece & 0.2359 & 0.1844 & 0.9338 & 0.9281 & 1.2025 \\
            \textbf{Small} & (2345,1234) & WordPiece & 0.2359 & 0.1841 & 0.9402 & 0.9375 & 1.2021 \\
            \hline
            \mc{3}{c}{$\bar{x} \pm s_N$}  & 0.2357 $\pm$ 0.0004 & 0.1840 $\pm$ 0.0005 & 0.9400 $\pm$ 0.0062 & 0.9358 $\pm$ 0.0070 & 1.2020 $\pm$ 0.0006  \\            
            \hline            
            \alert{\textbf{Base}$^b$} & \alert{(1234,1234)} & \alert{WordPiece} & \alert{1.388} & \alert{4.724} & \alert{0.1301} & \alert{0.02994} & \alert{112.61} \\
            \textbf{Base}  & (1234,2345) & WordPiece & 0.1357 & 0.1775 & 0.9557 & 0.9528 & 1.1454 \\
            \textbf{Base}  & (2345,1234) & WordPiece & 0.1782 & 0.1350 & 0.9619 & 0.9587 & 1.1446 \\
            \hline
            \mc{3}{c}{$\bar{x} \pm s_N$}  & 0.1569 $\pm$ 0.0300 & 0.1562 $\pm$ 0.0300 & 0.9588 $\pm$ 0.0044 & 0.9557 $\pm$ 0.0042 & 1.1450 $\pm$ 0.0006  \\            
            \hline
            \textbf{Tiny}  & (1234,1234) & BPE       & 0.5982 & 0.4430 & 0.8652 & 0.8520 & 1.5574 \\
            \textbf{Tiny}  & (1234,2345) & BPE       & 0.5803 & 0.4258 & 0.8743 & 0.8611 & 1.5308 \\
            \textbf{Tiny}  & (2345,1234) & BPE       & 0.5867 & 0.4333 & 0.8762 & 0.8664 & 1.5423 \\
            \hline
            \mc{3}{c}{$\bar{x} \pm s_N$}  & 0.5884 $\pm$ 0.0090 & 0.4340 $\pm$ 0.0086 & 0.8719 $\pm$ 0.0058 & 0.8599 $\pm$ 0.0073 & 1.5435 $\pm$ 0.0133  \\            
            \hline            
            \textbf{Small} & (1234,1234) & BPE       & 0.2031 & 0.1597 & 0.9410 & 0.9374 & 1.1731 \\
            \textbf{Small} & (1234,2345) & BPE       & 0.2075 & 0.1626 & 0.9452 & 0.9422 & 1.1766 \\
            \textbf{Small} & (2345,1234) & BPE       & 0.2067 & 0.1620 & 0.9467 & 0.9441 & 1.1759 \\
            \hline
            \mc{3}{c}{$\bar{x} \pm s_N$}  & 0.2058 $\pm$ 0.0023 & 0.1614 $\pm$ 0.0015 & 0.9443 $\pm$ 0.0030 & 0.9412 $\pm$ 0.0034 & 1.1752 $\pm$ 0.0018  \\            
            \hline            
            \textbf{Base}  & (1234,1234) & BPE       & 0.1513 & 0.1171 & 0.9554 & 0.9537 & 1.1243 \\
            \alert{\textbf{Base}$^b$}  & \alert{(1234,2345)} & \alert{BPE} & \alert{1.457} & \alert{4.130} & \alert{0.1076} & \alert{0.02092} & \alert{62.180}  \\
            \textbf{Base}  & (2345,1234) & BPE       & 0.1499 & 0.1154 & 0.9573 & 0.9554 & 1.1223 \\
            \hline
            \mc{3}{c}{$\bar{x} \pm s_N$}  & 0.1506 $\pm$ 0.0010 & 0.1163 $\pm$ 0.0012 & 0.9563 $\pm$ 0.0013 & 0.9546 $\pm$ 0.0012 & 1.1233 $\pm$ 0.0014  \\            
\hline\hline
\end{tabular}
}
\begin{tablenotes}
    \footnotesize
    \item $^a$ The entire PubChem (ver.~04/18/2025) dataset (119,184,806
    entries) is randomly split into training (95,347,844 entries amounting to
    80\% of the data) and validation (23,836,962 data points which counts as
    20\% of the data) sets. 
    \item $^b$ The model diverges during training and the behavior is persistent
    over several runs. As such, this experiment is excluded from calculating the
    mean ($\bar{x}$) and standard deviation ($s_N$).
    \item $^c$ $\uparrow$: higher is better; $\downarrow$: lower is better.
\end{tablenotes}
\end{sidewaystable}

\begin{figure}[!tbph]
    \centering
    \includegraphics[]{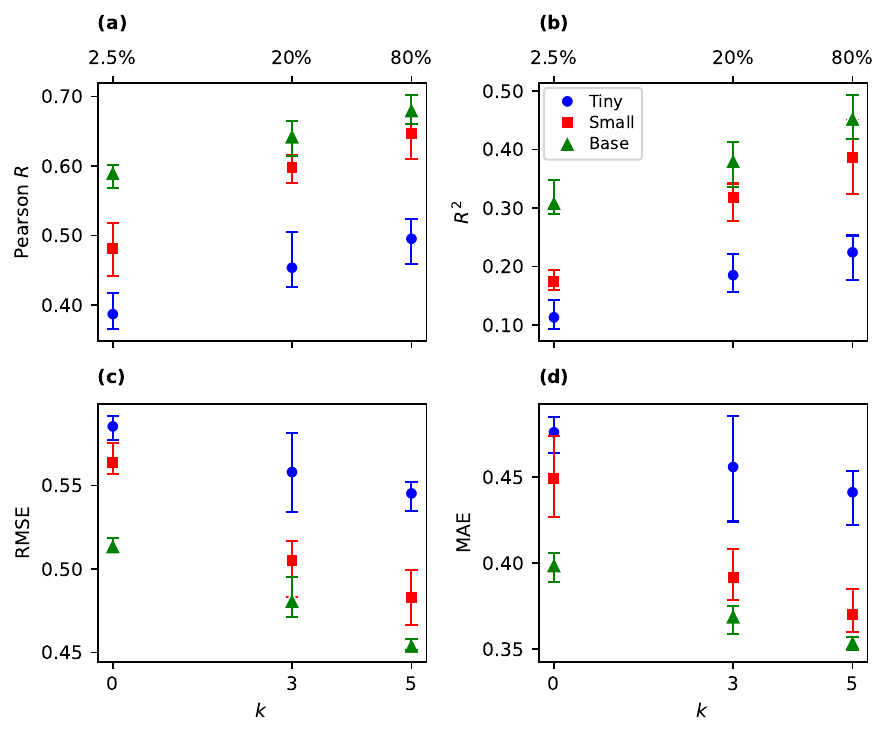}
    \caption{Variations of BERT's 3-fold cross-validation fine-tuning
    metrics, (a) Pearson $R$, (b) $R^2$, (c) RMSE, (d) MAE, versus the
    pre-training dataset size bin index, $k$, for the HLM endpoint (the
    corresponding data percentages are shown on the top axes). A hyperparameter
    search of 50 models is performed on each fold for each model variant and
    dataset size bin index and the best model is selected based on the
    validation $R^2$, following Ref.~\citenum{Fang:2023:3263}}
    \label{FIG:SFTHLMVAL}
\end{figure}

\begin{figure}[!tbph]
    \centering
    \includegraphics[]{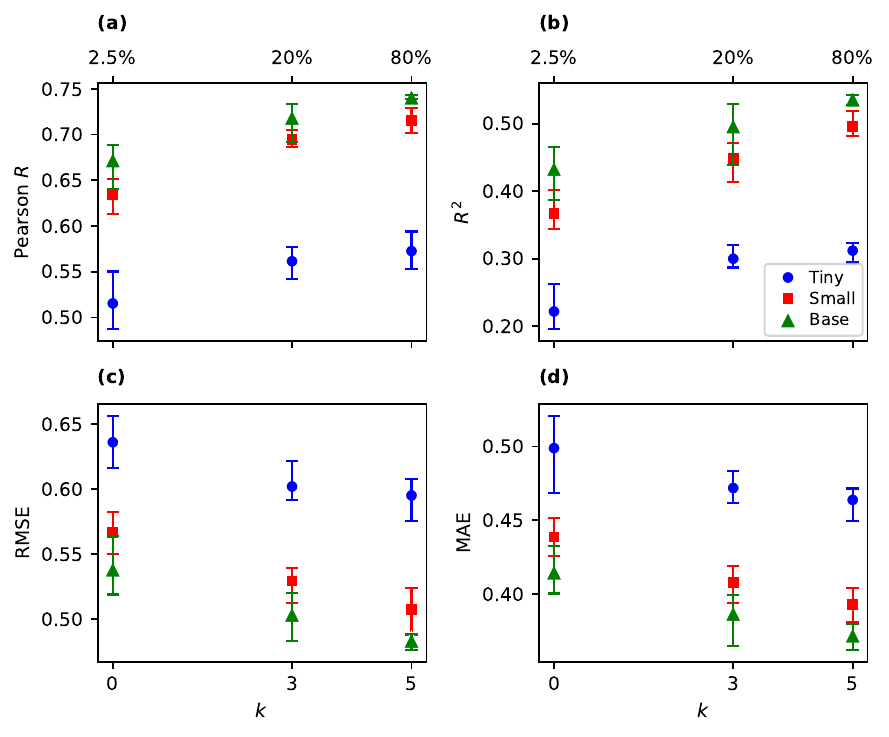}
    \caption{Variations of BERT's 3-fold cross-validation fine-tuning
    metrics, (a) Pearson $R$, (b) $R^2$, (c) RMSE, (d) MAE, versus the
    pre-training dataset size bin index, $k$, for the hPPB endpoint (the
    corresponding data percentages are shown on the top axes). A hyperparameter
    search of 50 models is performed on each fold for each model variant and
    dataset size bin index and the best model is selected based on the
    validation $R^2$, following Ref.~\citenum{Fang:2023:3263}}
    \label{FIG:SFTHPPBVAL}
\end{figure}

\begin{figure}[!tbph]
    \centering
    \includegraphics[]{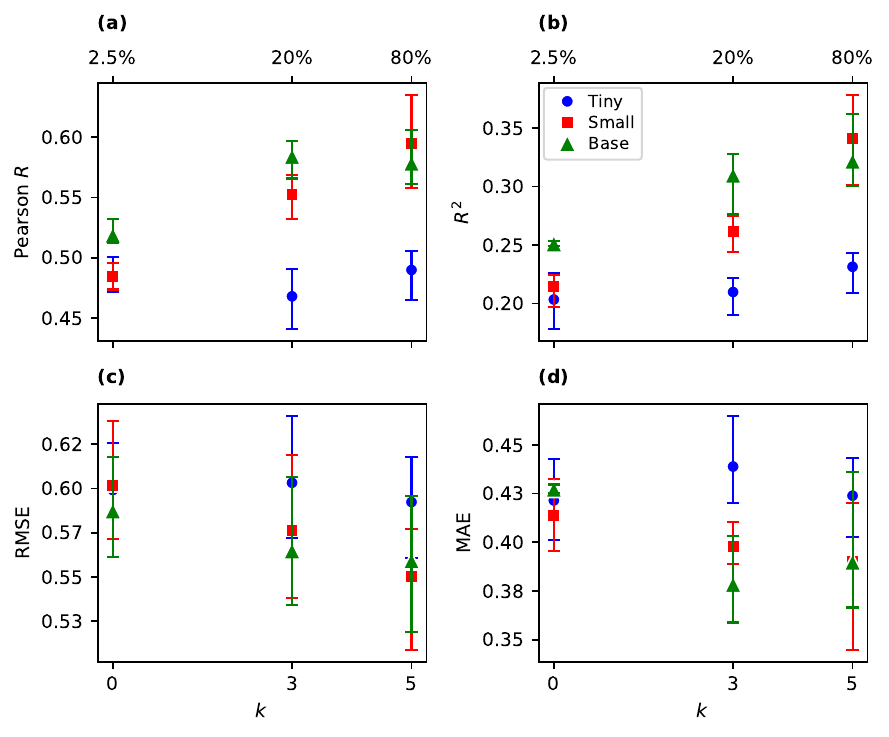}
    \caption{Variations of BERT's 3-fold cross-validation fine-tuning
    metrics, (a) Pearson $R$, (b) $R^2$, (c) RMSE, (d) MAE, versus the
    pre-training dataset size bin index, $k$, for the solubility endpoint (the
    corresponding data percentages are shown on the top axes). A hyperparameter
    search of 50 models is performed on each fold for each model variant and
    dataset size bin index and the best model is selected based on the
    validation $R^2$, following Ref.~\citenum{Fang:2023:3263}}
    \label{FIG:SFTHLMSOLVAL}
\end{figure}

\pagebreak
\bibliography{ms}